\pdfoutput=1
\documentclass{article}

\usepackage{iclr2020_conference, times}
\iclrfinalcopy

\date{}

\usepackage{authblk}
\usepackage[noend]{algpseudocode}
\usepackage{algorithm}
\usepackage{comment}
\usepackage{booktabs}
\usepackage{footnote}
\usepackage{amsmath}
\usepackage{booktabs}
\usepackage[T1]{fontenc}    %
\usepackage{amssymb}
\usepackage{url}
\usepackage{amsthm}
\usepackage{array}
\usepackage{tabularx}
\usepackage{subcaption}
\usepackage{caption}
\usepackage{multirow}
\usepackage{enumitem}
\usepackage{todonotes}
\usepackage{hyperref}

\usepackage{color}
\usepackage{graphicx}

\newcolumntype{C}{>{\centering}X}%

\usepackage{thmtools}
\usepackage{thm-restate}

\makeatletter
\def\blfootnote{\gdef\@thefnmark{*}\@footnotetext}
\makeatother

\title{A Closer Look at Deep Policy Gradients}

\author[1*]{Andrew Ilyas}
\author[1*]{Logan Engstrom}
\author[1]{Shibani Santurkar}

\author[1]{Dimitris Tsipras}
\author[2]{\\Firdaus Janoos}
\author[1,2]{Larry Rudolph}
\author[1]{Aleksander M\k{a}dry}

\affil[ ]{${}^{1}$MIT\ \ \ \ ${}^{2}$Two Sigma}
\affil[ ]{\texttt{\{ailyas,engstrom,shibani,tsipras,madry\}@mit.edu}}
\affil[ ]{\texttt{rudolph@csail.mit.edu, firdaus.janoos@twosigma.com}}

\begin{document}
\blfootnote{Equal contribution. Work done in part while interning at Two Sigma.}
\maketitle

\begin{abstract}
    We study how the behavior of deep policy gradient algorithms reflects the conceptual framework motivating their development. To this end, we propose a fine-grained analysis of state-of-the-art methods based on key elements of this framework: gradient estimation, value prediction, and optimization landscapes. Our results show that the behavior of deep policy gradient algorithms often deviates from what their motivating framework would predict: the surrogate objective does not match the true reward landscape, learned value estimators fail to fit the true value function, and gradient estimates poorly correlate with the ``true'' gradient. The mismatch between predicted and empirical behavior we uncover highlights our poor understanding of current methods, and indicates the need to move beyond current benchmark-centric evaluation methods.
\end{abstract}

\section{Introduction}
\label{sec:intro}
Deep reinforcement learning (RL) is behind some of the
most publicized achievements of modern machine
learning~\citep{silver2017mastering,dota,dayarathna2016data,openai2018learning}. In fact, to many, this
framework embodies the promise of the real-world impact of machine learning. 
However, the deep RL toolkit has not yet attained the same level of engineering
stability as, for example, the current deep (supervised) learning framework. Indeed,
recent studies demonstrate that state-of-the-art deep
RL algorithms suffer from oversensitivity to hyperparameter choices, lack of
consistency, and poor reproducibility \citep{henderson2017deep}.

This state of affairs suggests that it might be necessary to re-examine the conceptual underpinnings of deep RL methodology. More precisely, the overarching question that motivates this work is:
\begin{center}
{\em To what degree does current practice in deep RL reflect the principles informing its development?}
\end{center}
Our specific focus is on deep policy gradient methods,
a widely used class of deep RL algorithms. Our goal is to
explore the extent to which state-of-the-art implementations of
these methods succeed at realizing the key primitives of the general policy
gradient framework.

\paragraph{Our contributions.}
We take a broader look at policy gradient algorithms and
their relation to their underlying framework.  With this perspective in
mind, we perform a fine-grained examination of key RL primitives as they
manifest in practice. Concretely, we study:
\begin{itemize}
    \item \textbf{Gradient Estimation:} we find that even when agents improve
      in reward, their gradient estimates used in parameter updates
      poorly correlate with the ``true'' gradient. We additionally show that
      gradient estimate quality decays with training progress and task
      complexity. Finally, we demonstrate that varying the sample regime
      yields training dynamics that are unexplained by the motivating
      framework and run contrary to supervised learning intuition.
    \item \textbf{Value Prediction:} our experiments indicate that value
      networks successfully solve the supervised learning task they are trained
      on, but do {\em not} fit the true value function. Additionally, employing a
      value network as a baseline function only marginally decreases the
      variance of gradient estimates compared to using true value as a baseline
      (but still dramatically increases agent's performance compared to using no
      baseline at all).
    \item \textbf{Optimization Landscapes:} we show that the
	optimization landscape induced by modern policy gradient algorithms
	is often not reflective of the underlying true reward landscape,
	and that the latter is frequently poorly behaved in the relevant sample
	regime.
\end{itemize}

Overall, our results demonstrate that the motivating theoretical framework for
deep RL algorithms is often unpredictive of phenomena arising in practice. This
suggests that building reliable deep RL algorithms requires moving past
benchmark-centric evaluations to a multi-faceted understanding of their often
unintuitive behavior. We conclude (in Section~\ref{sec:takeaways}) by discussing
several areas where such understanding is most critically needed.

\section{Examining the Primitives of Deep Policy Gradient Algorithms}
In this section, we investigate the degree
to which our theoretical understanding of RL applies to modern methods.
We consider key primitives of policy gradient algorithms:
gradient estimation, value prediction and reward fitting.  In what follows,
we perform a fine-grained analysis of state-of-the-art policy gradient
algorithms (PPO and TRPO) through the lens of these primitives---detailed
preliminaries, background, and notation can be found in
Appendix~\ref{app:background}.

\subsection{Gradient estimate quality}
\label{sec:gradient}
A central premise of policy gradient methods is that stochastic gradient
ascent on a suitable objective function yields a good policy.
These algorithms use as a primitive the gradient of
that objective function:
\begin{align}
\label{eqn:grad_sr}
\hat{g} = \nabla_\theta \mathbb{E}_{(s_t, a_t) \sim
	\pi_0}\left[\frac{\pi_\theta(a_t|s_t)}{\pi_0(a_t|s_t)}\widehat{A}_{\pi_0}(s_t, a_t)\right]  = \mathbb{E}_{(s_t, a_t) \sim
     	\pi_0}\left[\frac{\nabla_\theta
    \pi_\theta(a_t|s_t)}{\pi_0(a_t|s_t)}\widehat{A}_{\pi_0}(s_t,
a_t)\right],
\end{align}
where in the above we use standard RL notation (see
Appendix~\ref{app:background} for more details).
An underlying assumption behind these methods is that we have access
to a reasonable estimate of this quantity. This assumption effectively translates into an
assumption that we can accurately estimate the expectation above using an
empirical mean of finite (typically $\sim 10^3$) samples. Evidently (since
the agent attains a high reward) these estimates are sufficient to
consistently improve reward---we are thus interested in the relative
quality of these gradient estimates in practice, and the effect of gradient
quality on optimization.

\begin{figure}[!htp]
	\includegraphics[width=1\textwidth]{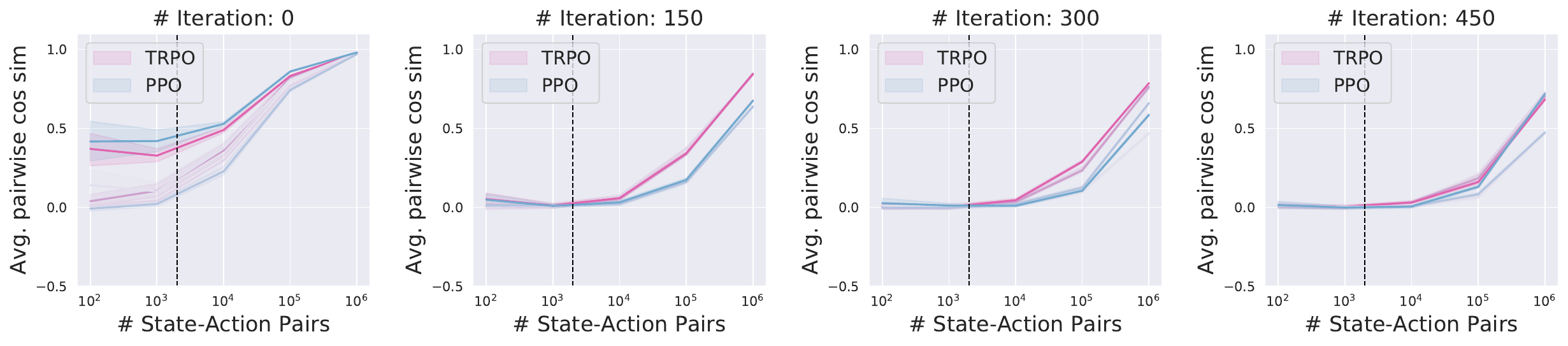}
	\caption{Empirical variance of the estimated gradient (c.f.
    \eqref{eqn:grad_sr}) as a function of the number of state-action pairs used
    in estimation in the MuJoCo Humanoid task. We measure the average pairwise
    cosine similarity between ten repeated gradient measurements taken from the
    same policy, with the $95\%$ confidence intervals (shaded). For each
    algorithm, we perform multiple trials with the same hyperparameter
    configurations but different random seeds, shown as repeated lines in the
    figure. The vertical line (at $x = 2$K) indicates the sample regime used for
    gradient estimation in standard implementations of policy gradient methods.
    In general, it seems that obtaining tightly concentrated gradient estimates
    would require significantly more samples than are used in practice,
    particularly after the first few timesteps. For other tasks -- such as
    Walker2d-v2 and Hopper-v2 -- the plots (seen in Appendix
    Figure~\ref{fig:gradvar_app}) have similar trends, except that gradient
    variance is slightly lower. Confidence intervals calculated with 500 sample
    bootstrapping.}
	\label{fig:gradvar}
\end{figure}

\begin{figure}[!htb]
	\includegraphics[width=1\textwidth]{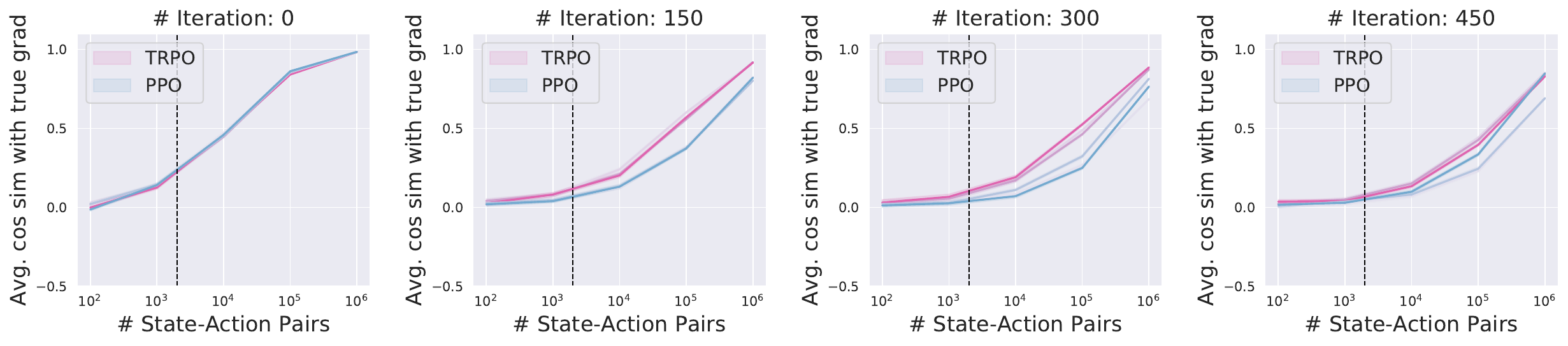}
	\caption{Convergence of gradient estimates (c.f. \eqref{eqn:grad_sr}) to the
    ``true'' expected gradient in the MuJoCo Humanoid task. We measure the mean
    cosine similarity between the ``true'' gradient approximated using ten
    million state-action pairs, and ten gradient estimates which use increasing
    numbers of state-action pairs (with 95\% confidence intervals). For each
    algorithm, we perform multiple trials with the same hyperparameter
    configurations but different random seeds. The vertical line (at $x = 2$K)
    indicates the sample regime used for gradient estimation in standard
    implementations of policy gradient methods. Observe that although it is
    possible to empirically estimate the true gradient, this requires
    several-fold more samples than are used commonly in practical applications
    of these algorithms. See additionally that the estimation task becomes more
    difficult further into training. For other tasks -- such as Walker2d-v2 and
    Hopper-v2 -- the plots (seen in Appendix Figure~\ref{fig:truegrad_app}) have
    similar trends, except that gradient estimation is slightly better.
    Confidence intervals calculated with 500 sample bootstrapping.}
	\label{fig:truegrad}
\end{figure}

\paragraph{How accurate are the gradient estimates we compute?} To answer
this question, we examine two of the most natural measures of estimate
quality: the empirical variance and the convergence to the ``true''
gradient. To evaluate the former, we measure the average pairwise cosine
similarity between estimates of the gradient computed from the same policy with
independent rollouts (Figure~\ref{fig:gradvar}). We evaluate the latter by
first forming an estimate
of the true gradient with a large number of state-action pairs. We then
examine the convergence of gradient estimates to this ``true''
gradient (which we once again measure
using cosine similarity) as we increase the number of samples
(Figure~\ref{fig:truegrad}).

We observe that {\em deep policy gradient methods operate with relatively
poor estimates of the gradient}, especially as task
complexity increases and as training progresses (contrast Humanoid-v2, a
``hard'' task, to other tasks and contrast successive checkpoints in
Figures~\ref{fig:gradvar} and~\ref{fig:truegrad}). This is in spite of the
fact that our agents continually improve throughout training, and attain
nowhere near the maximum reward possible on each task. In fact, we
sometimes observe a \emph{zero} or \emph{even negative} correlation in the relevant
sample regime\footnote{Deep policy gradient algorithms use gradients
    indirectly to compute steps---in Appendix~\ref{app:gradest} we show
that our results also hold true for these computed steps.}.

While these results might be reminiscent of the well-studied ``noisy
gradients'' problem in supervised
learning~\citep{convex1,convex2,dl1,dl2,dl3,dl4,dl5}, we have very little
understanding of how gradient quality affects optimization in
the substantially different reinforcement learning setting. For example:
\begin{itemize}
    \item The sample regime in which RL algorithms operate seems to have a
	profound impact on the {\em robustness and stability of agent
	training}---in particular, many of the sensitivity issues reported
	by~\citet{henderson2017deep} are claimed to
	disappear~\citep{ilyatalk} in higher-sample regimes.
  Understanding
	the implications of working in this sample regime, and more
	generally the impact of sample complexity on training stability
	remains to be precisely understood.
    \item Agent policy networks are trained concurrently with {\em value
      networks} (discussed more in the following section) meant to reduce the
      variance of gradient estimates. Under our conceptual framework, we might
      expect these networks to help gradient estimates more as training
      progresses, contrary to what we observe in Figure~\ref{fig:gradvar}. The
      value network also makes the now {\em two-player} optimization landscape
      and training dynamics even more difficult to grasp, as such interactions
      are poorly understood.
    \item The relevant measure of sample complexity for many settings
	(number of state-action pairs) can differ drastically from the
	number of \textit{independent} samples used at each training
	iteration (the number of complete trajectories). The latter
	quantity (a) tends to be much lower than the number of state-action
	pairs, and (b) decreases across iterations during training. 
\end{itemize}

All the above factors make it unclear to what degree our intuition from
classical settings transfer to the deep RL regime. And the policy gradient
framework, as of now, provides little predictive power regarding the
variance of gradient estimates and its impact on reward optimization.

Our results indicate that despite having a rigorous theoretical
framework for RL, we lack a precise understanding of the structure of the
reward landscape and optimization process.

\subsection{Value prediction}
\label{sec:val-est}
Our findings from the previous section motivate a deeper look into
gradient estimation. After all, 
the policy gradient in its original formulation~\citep{Sutton1999PolicyGM}
is known to be hard to estimate, and thus algorithms employ a variety
of variance reduction methods. The most popular
of these techniques is a baseline function. Concretely, an
equivalent form of the policy gradient is given by:
\begin{align}
\widehat{g}_\theta &= \mathbb{E}_{\tau \sim
    \pi_\theta}\left[\sum_{(s_t, a_t) \in \tau}\nabla_\theta\log
\pi_\theta(a_t|s_t) \cdot (Q_{\pi_\theta}(s_t, a_t) - b(s_t))\right]
\end{align}
where $b(s_t)$ is some fixed function of the state $s_t$. 
A canonical choice of baseline function is the value function
$V_\pi(s)$, the expected return from a given state (more details and
motivation in~\ref{app:background}):
\begin{align}
\label{eq:val}
V_{\pi_\theta}(s_t) &= \mathbb{E}_{\pi_\theta} [R_t|s_t] \;.
\end{align}
Indeed, fitting a value-estimating
function~\citep{Schulman2015HighDimensionalCC,sutton2018reinforcement} (a neural
network, in the deep RL setting) and using it as a baseline function is
precisely the approach taken by most deep policy gradient methods. Concretely,
one trains a value network $V_{\theta_t}^\pi$ such that:
\begin{align}
    \label{eq:val_targ}
    \theta_t = \min_\theta\ \mathbb{E}\left[\left(V^\pi_\theta(s_t) -
	    (V^\pi_{\theta_{t-1}}(s_t) + A_t)\right)^2\right]
\end{align}
where $V^\pi_{\theta_{t-1}}(s_t)$ are estimates given by the last value
function, and $A_t$ is the {\em advantage} of the policy, i.e. the returns
minus the estimated values.
(Typically, $A_t$ is estimated using generalized advantage estimation, as
described in~\citep{Schulman2015HighDimensionalCC}.) Our findings in the
previous section prompt us to take a closer look at the value network
and its impact on the variance of gradient estimates.

\begin{figure}[!h]
  \centering
	\begin{subfigure}[b]{1\textwidth}
		\centering
		\includegraphics[width=\textwidth]{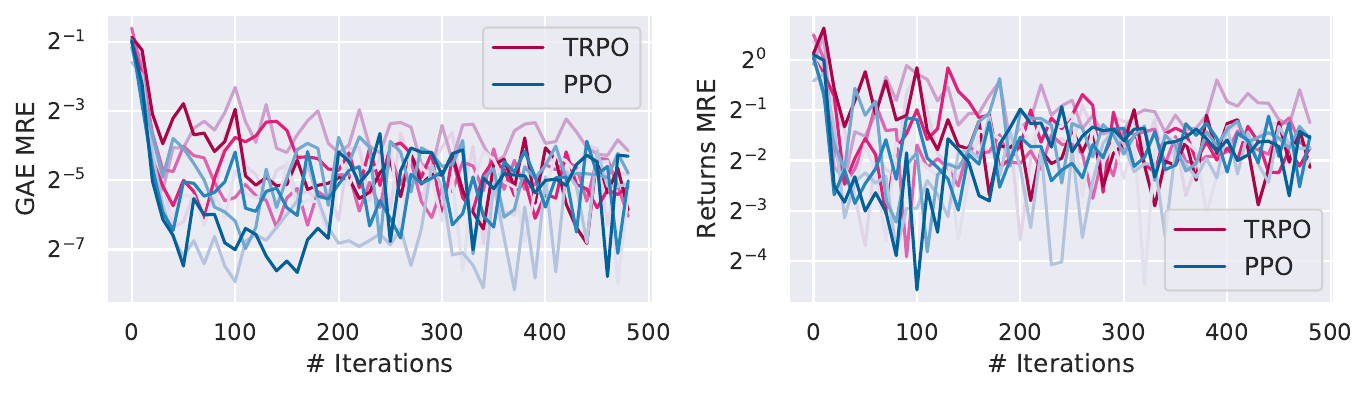}
		 
		\caption{}
		\label{fig:vala}
	\end{subfigure}
	\caption{Quality of value prediction in terms of mean relative error (MRE) on heldout state-action pairs for agents trained to solve the MuJoCo Walker2d-v2 task.
		We observe in (\emph{left}) that the agents do indeed succeed at solving
		the supervised learning task they are trained for---the
		MRE on the GAE-based value loss $(V_{old} + A_{GAE})^2$ 
		(c.f.~\eqref{eq:val_targ}) is small.
		On the other hand, in (\emph{right}) we see that the returns MRE is
		still quite high---the learned value function is off by
		about $50\%$ with respect to the underlying true value function.
		Similar plots for other MuJoCo tasks are in
		Appendix~\ref{app:value_pred}.
	}
	\label{fig:val}
\end{figure}

\begin{figure}[!h]
	\centering
	\begin{subfigure}[b]{\textwidth}
		\includegraphics[width=\textwidth]{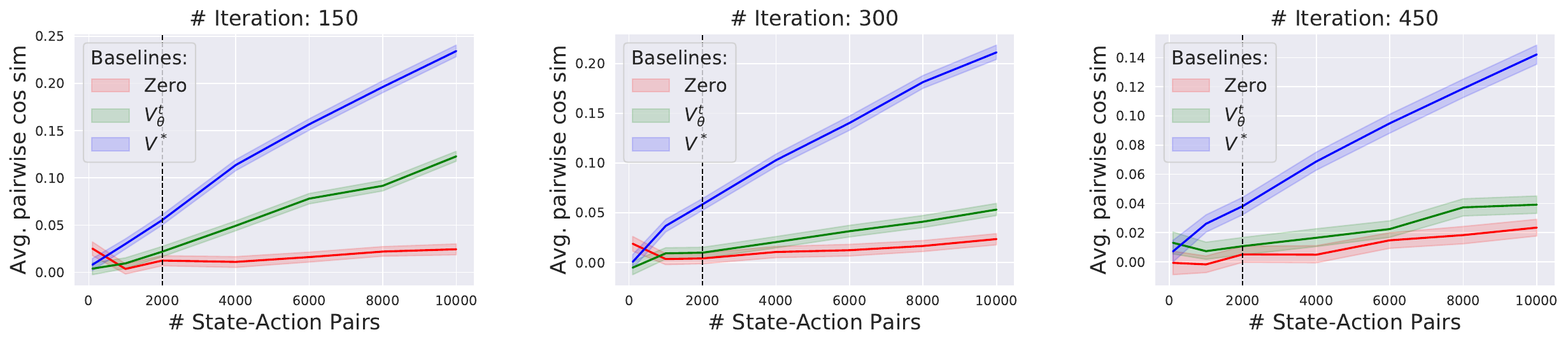}
		\caption{Walker2d-v2}
	\end{subfigure}
	\begin{subfigure}[b]{\textwidth}
		\includegraphics[width=\textwidth]{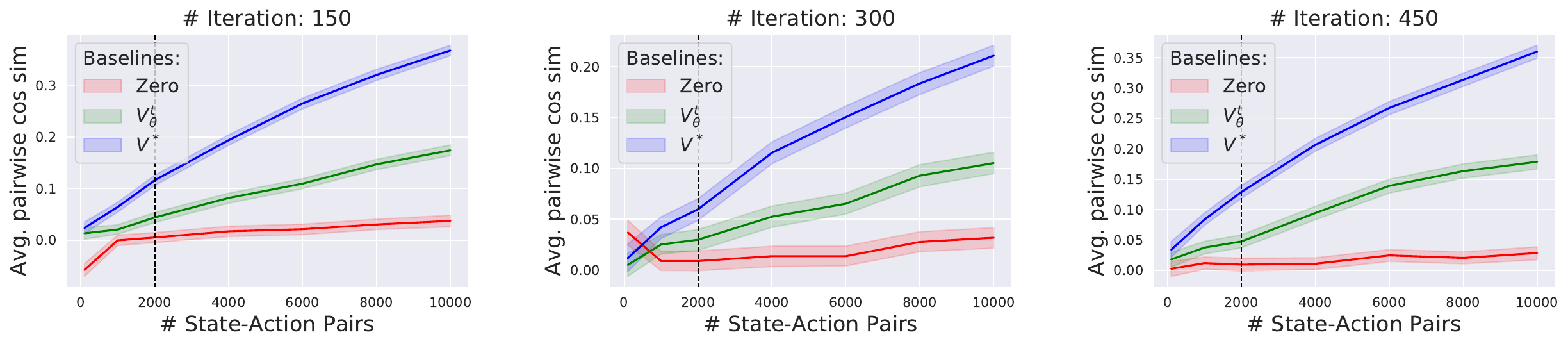}
		\caption{Hopper-v2}
	\end{subfigure}
	\caption{Efficacy of the value network as a variance reducing baseline for
    Walker2d-v2 (top) and Hopper-v2 (bottom) agents. We measure the empirical
    variance of the gradient (c.f. \eqref{eqn:grad_sr}) as a function of the
    number of state-action pairs used in estimation, for different choices of
    baseline functions: the value network (used by the agent in training), the
    ``true'' value function (fit to the returns using $5\cdot 10^6$ state-action
    pairs sampled from the \emph{current} policy) and the ``zero'' value
    function (i.e. replacing advantages with returns). We observe that using the
    true value function leads to a significantly lower-variance estimate of the
    gradient compared to the value network. In turn, employing the value network
    yields a noticeable variance reduction compared to the zero baseline
    function, even though this difference may appear rather small in the
    small-sample regime ($2$K). Confidence intervals calculated with 10 sample
    bootstrapping.
	}
	\label{fig:val-all}
\end{figure}

\paragraph{Value prediction as a supervised learning problem.} We first
analyze the value network through the lens of the supervised learning problem it
solves. After all,~\eqref{eq:val_targ} describes an empirical risk
minimization, where a loss is minimized over a set of
sampled %
$(s_t, a_t)$. 
So, how does $V^\pi_\theta$ perform as a solution
to~\eqref{eq:val_targ}? And in turn, how does~\eqref{eq:val_targ}
perform as a proxy for learning the true value function?

Our results (Figure~\ref{fig:vala}) show that the value
network \textit{does} succeed at both fitting the given loss function and
generalizing to unseen data, showing low and stable mean relative error (MRE). However,
the significant drop in performance as shown in Figure~\ref{fig:val}
indicates that the supervised learning problem induced
by~\eqref{eq:val_targ} \textit{does not} lead to $V^\pi_\theta$ learning
the underlying true value function.  

\paragraph{Does the value network lead to a reduction in variance?}
Though evaluating the $V^\pi_\theta$ baseline function as a value predictor
as we did above is informative, in the end the sole purpose of the value
function is to reduce variance.  So: how does using our value function
actually impact the variance of our gradient estimates? To answer this
question, we compare the variance reduction that results from employing our
value network against both a ``true'' value function and a trivial ``zero''
baseline function (i.e. simply replacing advantages with returns). Our
results, captured in Figure~\ref{fig:val-all}, show that the ``true'' value
function yields a much lower-variance estimate of the gradient. This is
especially true in the sample regime in which we operate. We note, however,
that despite not effectively predicting the true value function or inducing
the same degree of variance reduction, the value network \textit{does} help
to some degree (compared to the ``zero'' baseline). Additionally, the
seemingly marginal increase in gradient correlation provided by the value
network (compared to the ``true'' baseline function) turns out to result in
a significant improvement in agent performance. (Indeed, agents trained
without a baseline reach almost an order of magnitude worse reward.) 

Our findings suggest that we still need a better understanding
of the role of the value network in agent training, and raise several
questions that we discuss in Section~\ref{sec:takeaways}.

\subsection{Exploring the optimization landscape}
\label{sec:landscapes}
Another key assumption of policy gradient algorithms is that
first-order updates (w.r.t. policy parameters) actually
yield better policies. It is thus natural to examine how valid this
assumption is.

\begin{figure}[!b]
  \centering
	\begin{tabularx}{\textwidth}{CCC}
        {2,000 state-action pairs}
        & {20,000 state-action pairs}
        & {100,000 state-action pairs}
	\end{tabularx}
	\begin{center}
	\includegraphics[width=.95\textwidth]{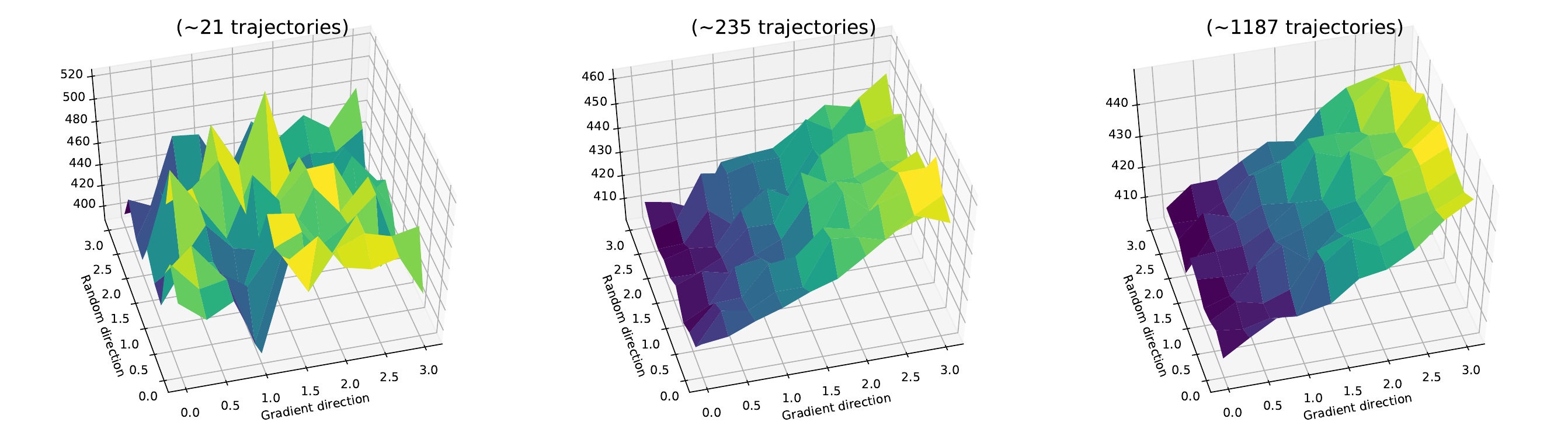}
    \end{center}
	\caption{True reward landscape concentration for TRPO on
	Humanoid-v2. We visualize the landscape at a
    training iteration 150 while varying the number of
    trajectories used in reward estimation (each subplot), both in the
direction of the step taken and a random direction.  Moving one unit along
the ``step direction'' axis corresponds to moving one full step
in parameter space. In the random direction one unit corresponds to
moving along a random norm $2$ Gaussian vector in the
parameter space. In practice, the norm of the step is typically
an order of magnitude lower than the random direction. While the landscape
is very noisy in the low-sample regime,
large numbers of samples reveal a well-behaved underlying
landscape. See
Figures~\ref{fig:ppo_landscape_concentration},~\ref{fig:trpo_landscape_concentration}
of the Appendix for additional plots.}
	\label{fig:trpo_landscape_concentration_main}
\end{figure}

\paragraph{The true rewards landscape.} We begin by
examining the landscape of agent reward with respect to the policy
parameters. Indeed, even if deep policy gradient methods do not optimize
for the true reward directly (e.g. if they use a surrogate objective), the
ultimate goal of any policy gradient algorithm is to navigate this
landscape. First, Figure~\ref{fig:trpo_landscape_concentration_main} shows
that 
while estimating the true reward landscape with a high number of samples
yields a relatively smooth reward landscape (perhaps suggesting viability
of direct reward optimization), 
estimating the true reward landscape in the typical, low sample regime
results in a landscape that appears jagged and poorly-behaved. The
low-sample regime thus gives rise to a certain kind of barrier to direct
reward optimization. Indeed, applying our algorithms in this regime makes
it impossible to distinguish between good and bad points in the landscape,
even though the true underlying landscape is fairly well-behaved.

\paragraph{The surrogate objective landscape.} The untamed nature
of the rewards landscape has led to the development of alternate approaches
to reward maximization. Recall that an important element of
many modern policy gradient methods is the maximization of a surrogate objective
function in place of the true rewards (the exact mechanism behind the surrogate objective is detailed in Appendix~\ref{app:background}, and particularly in \eqref{eqn:app_surrogate}). The surrogate objective, based
on relaxing the policy improvement theorem 
of Kakade and Langford~\citep{Kakade2002ApproximatelyOA}, can be viewed as a
simplification of the reward maximization objective.

As a purported approximation of the true returns, one would expect that the
surrogate objective landscape approximates the true reward landscape
fairly well.  That is, parameters corresponding to good surrogate
objective will also correspond to good true reward.

\begin{figure}[!t]
\centering
    \begin{tabular}{cc|c}
	    & Few state-action pairs (2,000) & Many state-action pairs ($10^6$) \\
	    & \begin{tabularx}{.4\textwidth}{CC} Surrogate & True reward \end{tabularx}
	    & \begin{tabularx}{.4\textwidth}{CC} Surrogate & True reward \end{tabularx} \\
	    \raisebox{1.5cm}{\rotatebox[origin=c]{90}{Step 0}} &
	    \includegraphics[width=.4\textwidth]{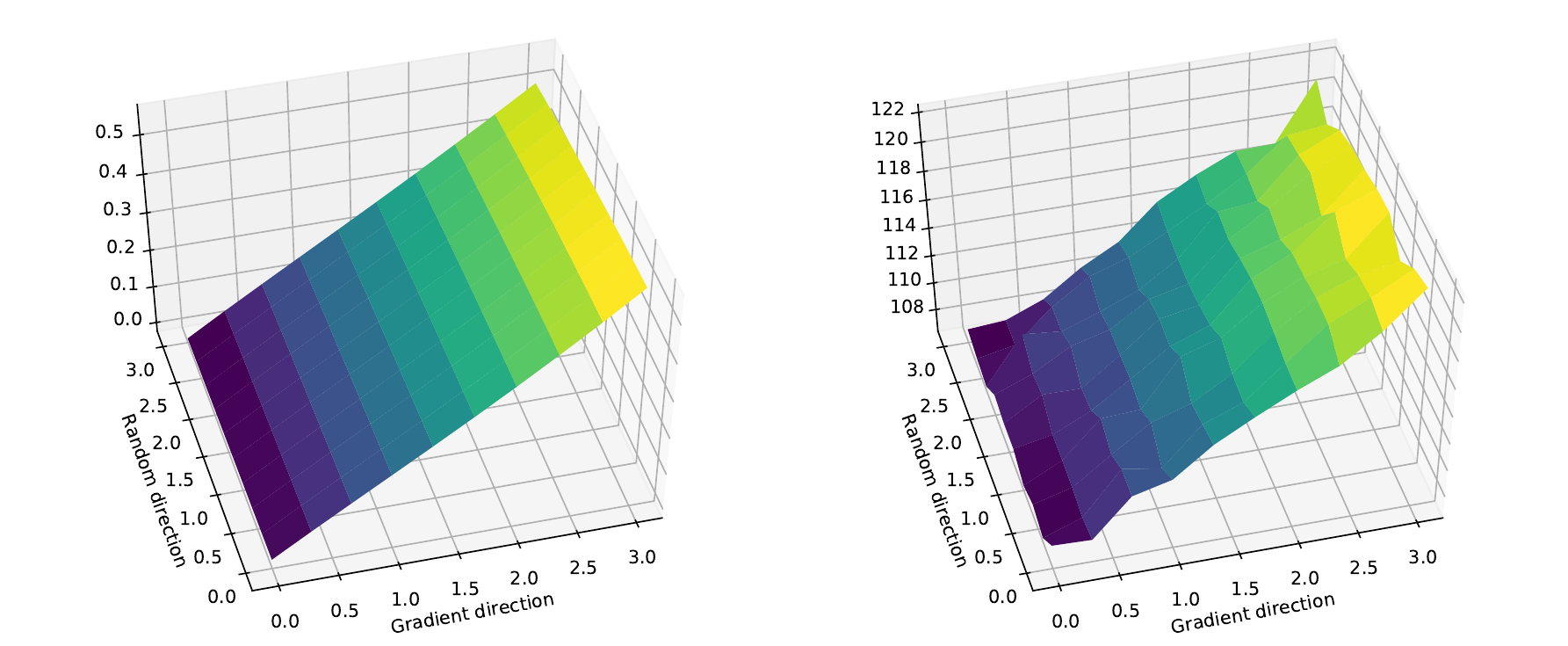} &
	    \includegraphics[width=.4\textwidth]{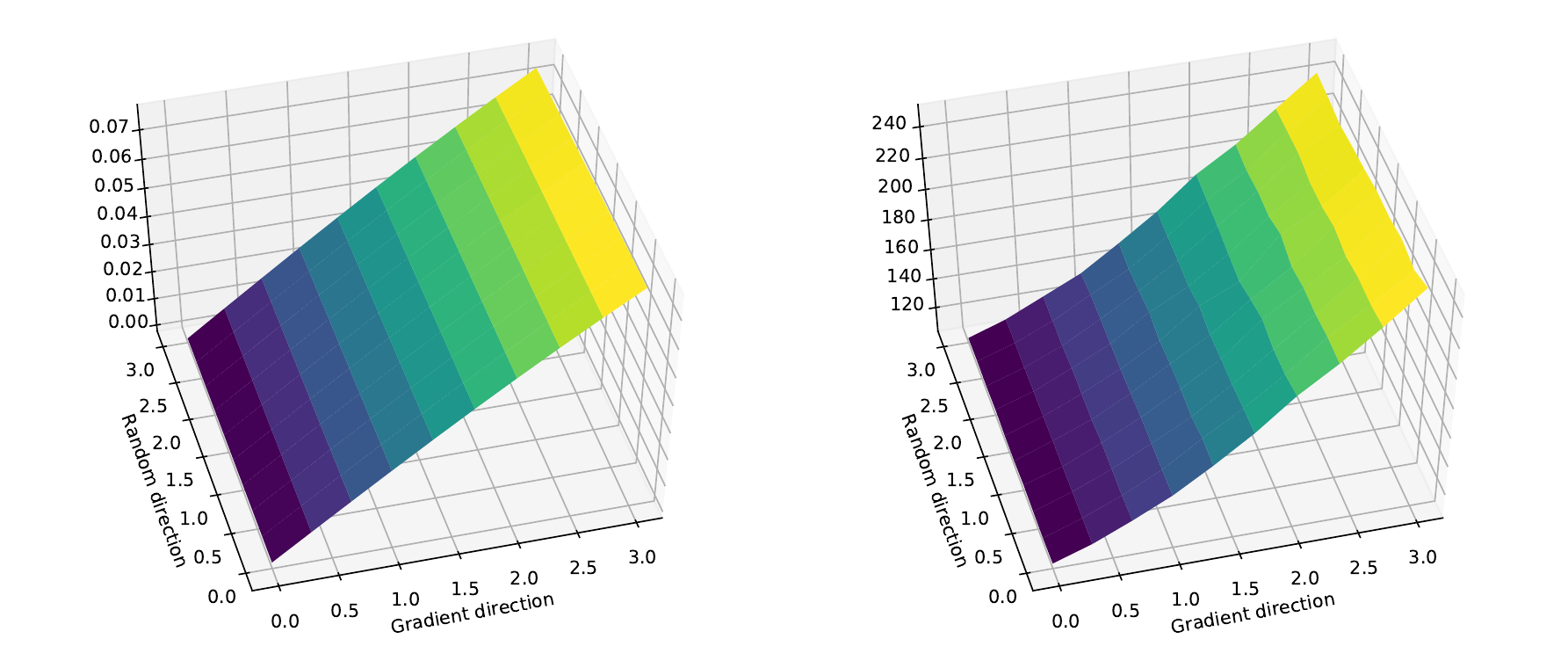} \\
	    \raisebox{1.5cm}{\rotatebox[origin=c]{90}{Step 300}} &
	    \includegraphics[width=.4\textwidth]{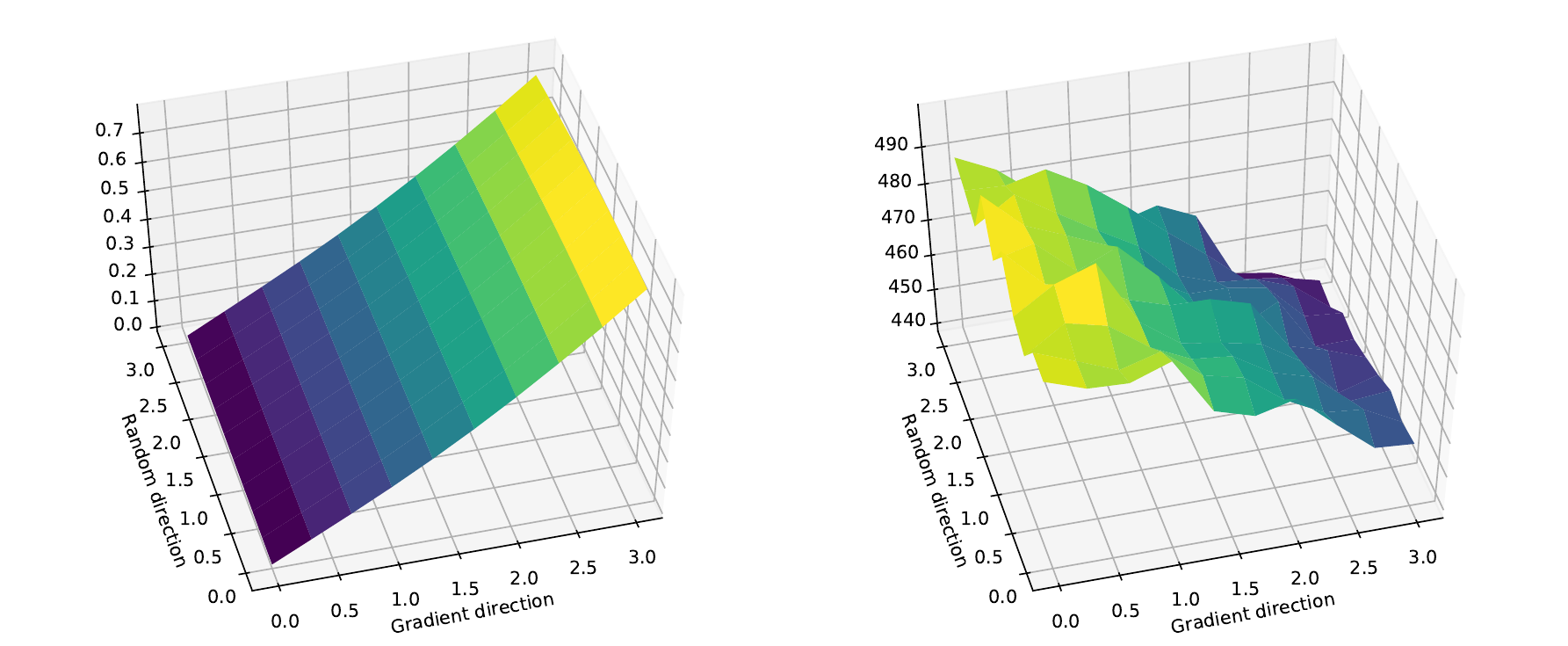} &
	    \includegraphics[width=.4\textwidth]{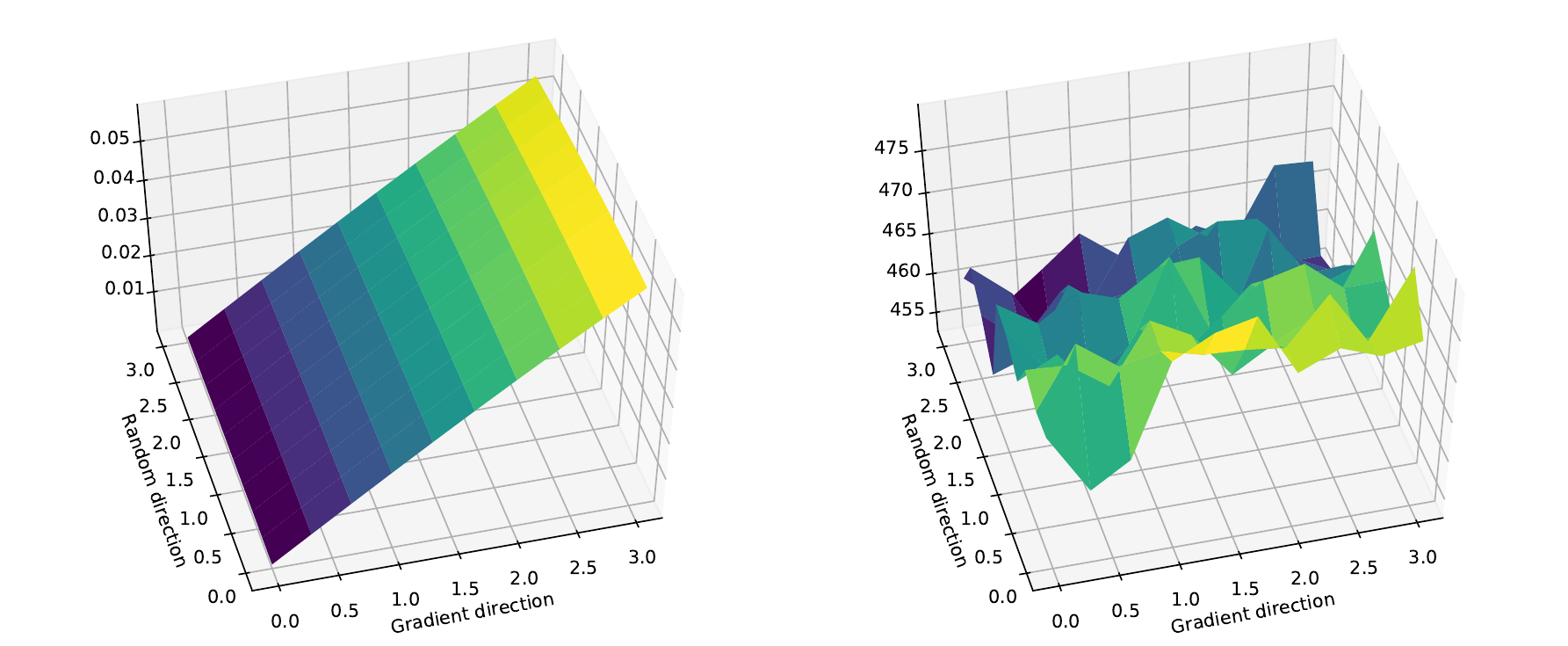}
    \end{tabular}
    \caption{True reward and surrogate objective landscapes for TRPO on the
    Humanoid-v2 MuJoCo task.  We visualize the landscapes
    in the direction of the update step and a random
direction (as in Figure~\ref{fig:trpo_landscape_concentration_main}).
The surrogate objective corresponds to the actual function optimized by the
algorithm at each step. We estimate true reward with $10^6$
state-action pairs per point. We compare the landscapes at different
points in training and with varying numbers of state-action pairs used
in the update step. Early in training the true and
surrogate landscapes align fairly well in both sample regimes, but later 
become misaligned in the low-sample regime.
More landscapes in
Appendix Figures~\ref{fig:humanoid_landscape_ppo}-\ref{fig:hopper_landscape_trpo}.}
    \label{fig:humanoid_trpo_landscape_main}
\end{figure}

\begin{figure}[!t]
\centering
\begin{tabular}{cc|c}
& Few state-action pairs (2,000) & Many state-action pairs ($10^6$) \\
& \begin{tabularx}{.4\textwidth}{CC} Surrogate & True reward \end{tabularx}
& \begin{tabularx}{.4\textwidth}{CC} Surrogate & True reward \end{tabularx} \\
\raisebox{1.5cm}{\rotatebox[origin=c]{90}{Step 0}} &
    \includegraphics[width=.375\textwidth]{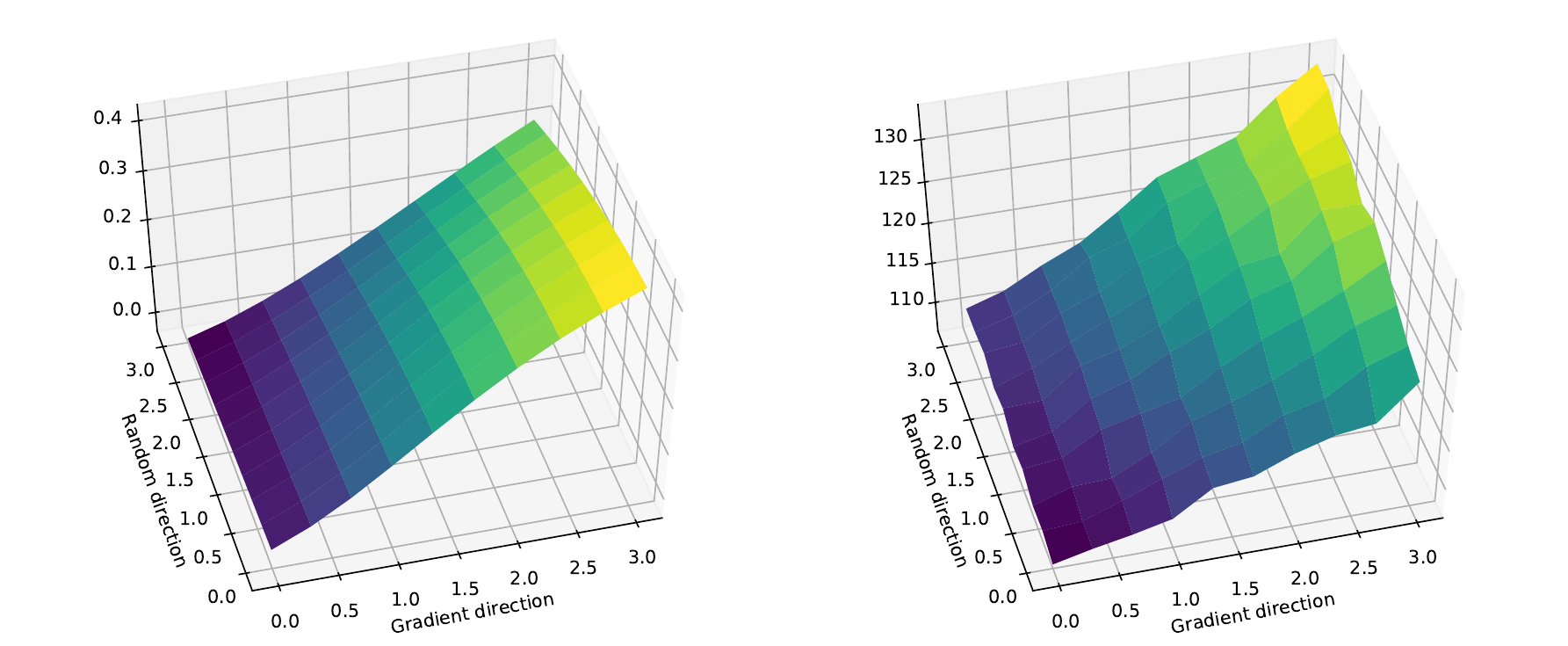} &
    \includegraphics[width=.375\textwidth]{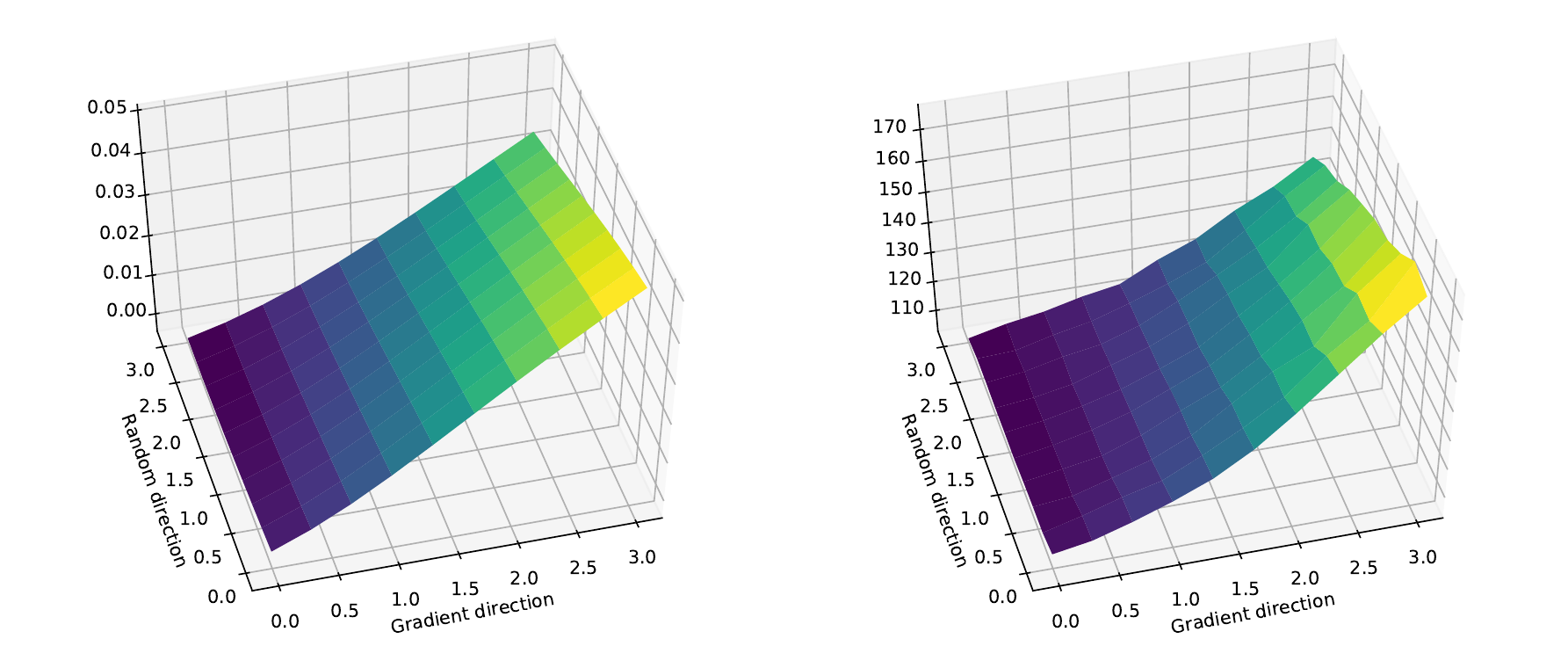} \\
\raisebox{1.5cm}{\rotatebox[origin=c]{90}{Step 300}}
    & \includegraphics[width=.375\textwidth]{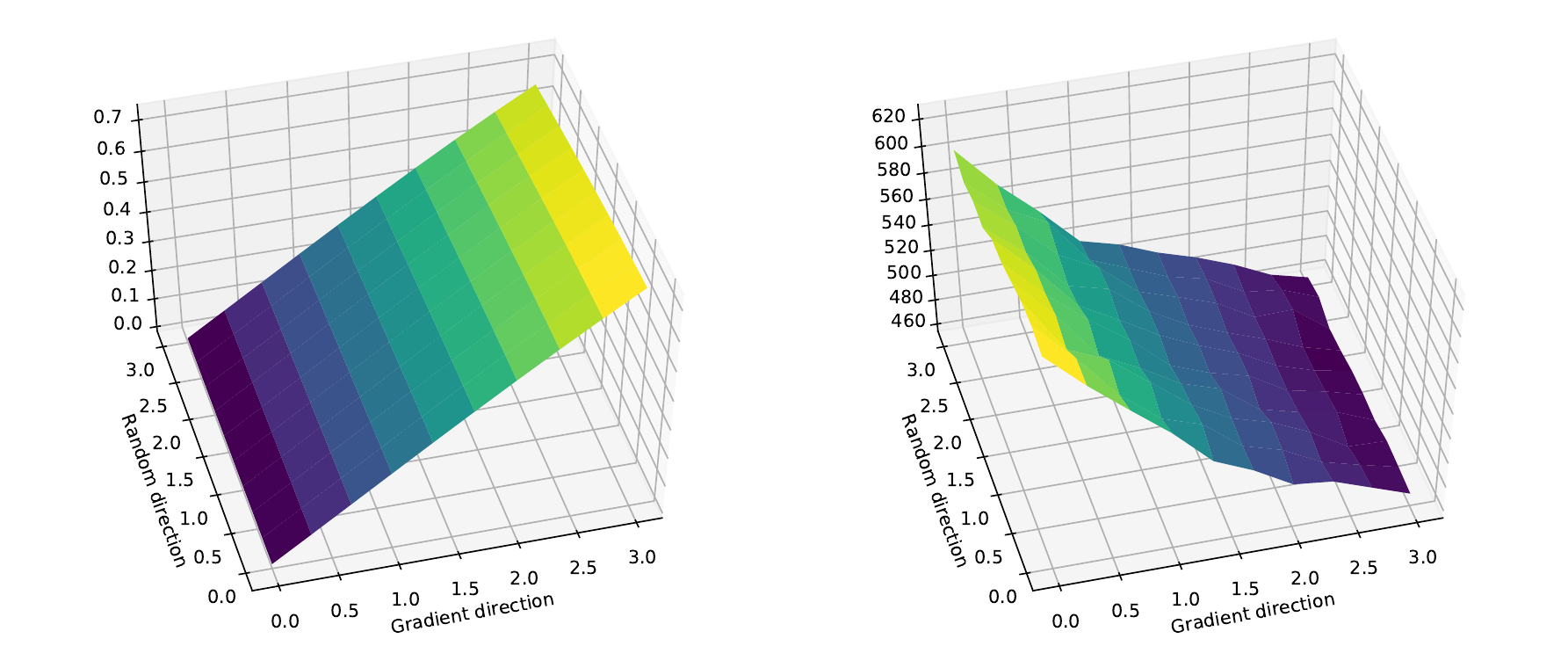} &
    \includegraphics[width=.375\textwidth]{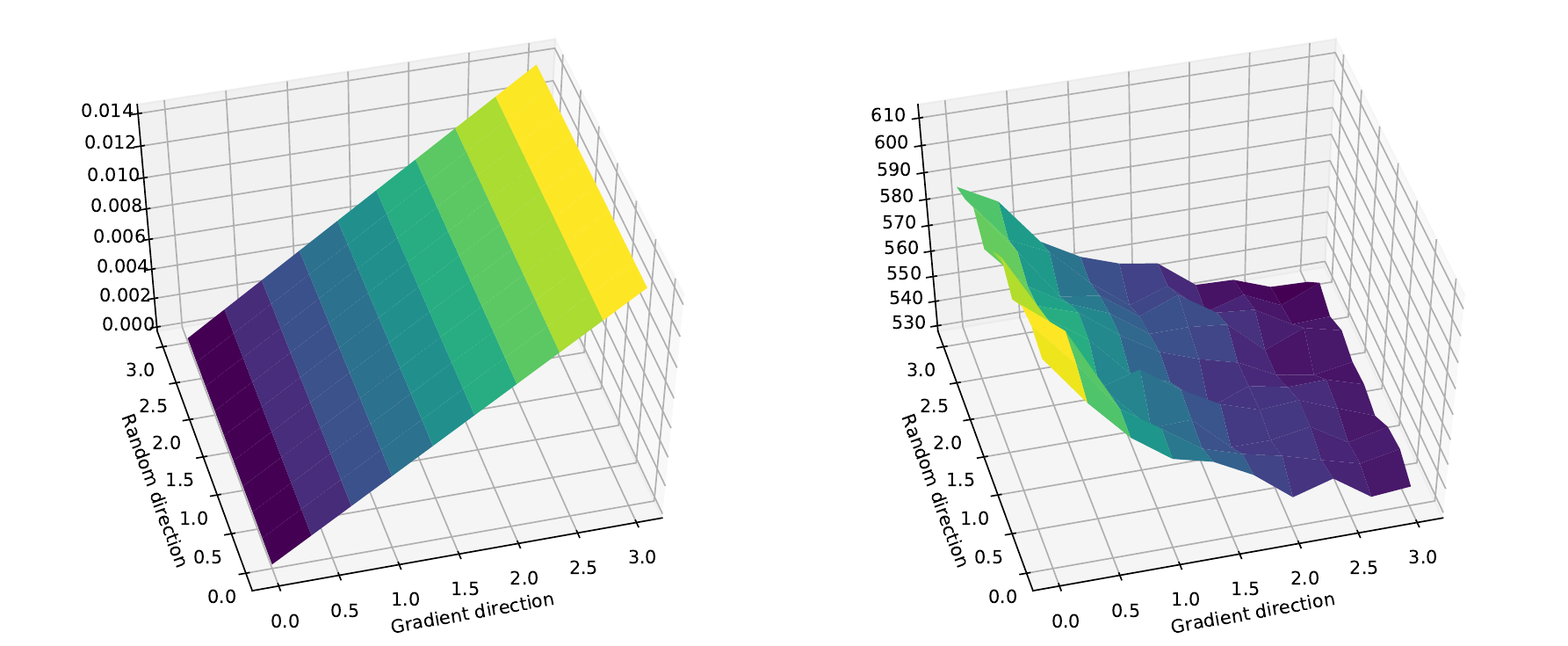}
\end{tabular}
\caption{True reward and surrogate objective landscapes for PPO on the Humanoid-v2 MuJoCo task.
See Figure~\ref{fig:humanoid_trpo_landscape_main} for a description.
We observe that early in training the true and surrogate landscapes align well.
However, later increasing the surrogate objective leads to lower true reward.}
\label{fig:humanoid_ppo_landscape_main}
\end{figure}

Figure~\ref{fig:humanoid_trpo_landscape_main} shows that in the early
stages of training, the optimization landscapes of the true reward and surrogate
objective are indeed approximately aligned. However, as training progresses, the
surrogate objective becomes much less predictive of the true reward in the
relevant sample regime.  
In particular, we often observe that directions that \emph{increase} the
surrogate objective lead to a \emph{decrease} of the true reward (see
Figures~\ref{fig:humanoid_trpo_landscape_main},~\ref{fig:humanoid_ppo_landscape_main}).
In a higher-sample regime (using several orders of magnitude more samples), we
find that PPO and TRPO turn out to behave rather differently. In the case
of TRPO, the update direction following the surrogate objective matches the
true reward much more closely. 
However, for PPO we consistently observe landscapes
where the step direction leads to lower true reward, even in the
high-sample regime.  This suggests that even when estimated accurately
enough, the surrogate objective might not be an accurate proxy for the true
reward. (Recall from Section~\ref{sec:gradient} that this is a
sample regime where we \textit{are} able to estimate the true gradient of
the reward fairly well.)

\section{Towards Stronger Foundations for Deep RL}
\label{sec:takeaways}
Deep reinforcement learning (RL) algorithms have shown great practical
promise, and are rooted in a well-grounded theoretical framework.
However, our results indicate that this framework often fails to
provide insight into the practical performance of these algorithms.
This disconnect impedes our understanding of why these algorithms succeed
(or fail), and is a major barrier to addressing key challenges
facing deep RL such as brittleness and poor reproducibility.

To close this gap, we need to either develop methods that adhere more
closely to theory, or build theory that can capture what makes existing
policy gradient methods successful. In both cases, the first step is to
precisely pinpoint where theory and practice diverge. To this end, we
analyze and consolidate our findings from the previous section.

\textbf{Gradient estimation.} Our analysis in Section~\ref{sec:gradient} shows
that the quality of gradient estimates that deep policy gradient algorithms use
is rather poor. Indeed, even when agents improve, such gradient estimates
often poorly correlate with the true gradient (c.f.
Figure~\ref{fig:truegrad}). We also note that gradient correlation decreases as
training progresses and task complexity increases. While this certainly does not
preclude the estimates from conveying useful signal, the exact underpinnings of
this phenomenon in deep RL still elude us. In particular, in
Section~\ref{sec:gradient} we outline a few keys ways in which the deep RL
setting is quite unique and difficult to understand from an optimization
perspective, both theoretically and in practice 
Overall, understanding the impact of gradient estimate quality on deep RL
algorithms is challenging and largely unexplored.

\textbf{Value prediction.} The findings presented in Section~\ref{sec:val-est}
identify two key issues. First, while the value network successfully solves the
supervised learning task it is trained on, it does not accurately model the
``true'' value function. Second, employing the value network as a baseline does
decrease the gradient variance (compared to the trivial (``zero'') baseline).
However, this decrease is rather marginal compared to the variance reduction
offered by the ``true'' value function.

It is natural to wonder whether this failure in modeling the value function
is inevitable. For example, how does the loss function used to train the value
network impact value prediction and variance reduction? More broadly, we lack an
understanding of the precise role of the value network in training. Can we
empirically quantify the relationship between variance reduction and
performance? And does the value network play a broader role than just
variance reduction?

\textbf{Optimization landscape.} We have also seen, in
Section~\ref{sec:landscapes}, that the optimization landscape induced by modern
policy gradient algorithms, the surrogate objective, is often not reflective of the
underlying true reward landscape. We thus need a deeper understanding of why
current methods succeed despite these issues, and, more broadly, how to better navigate
the true reward landscape.

\section{Related Work}
The idea of using gradient estimates to update neural network-based RL
agents dates back at least to the REINFORCE \citep{Williams1992SimpleSG}
algorithm. Later, Sutton \citep{Sutton1999PolicyGM} established a unifying
framework casting these algorithms as instances of the policy gradient class of
algorithms.
  
Our work focuses on proximal policy optimization
(PPO)~\citep{schulman2017proximal} and trust region policy optimization
(TRPO)~\citep{trpo}, which are two of the most prominent
policy gradient algorithms used in deep RL, drawing inspiration from works on
related algorithms, such as~\citep{Peters2010RelativeEP} and~\cite{Kakade2001ANP}. 

Many recent works document the brittleness of deep RL
algorithms~\citep{henderson2018did,henderson2017deep,islam2017reproducibility}.
\citep{Rajeswaran2017TowardsGA} and \citep{Mania2018SimpleRS}
demonstrate that on many benchmark tasks, state-of-the-art performance can
be attained by augmented randomized search
approaches. \cite{mcc2018empirical} investigates gradient noise in
large-batch settings, and \cite{ahmed2018understanding} investigates the
role of {\em entropy regularization} (which we do not study) on
optimization.

\section{Conclusion}
In this work, we analyze the degree to which key primitives of
deep policy gradient algorithms follow their conceptual underpinnings. Our
experiments show that these primitives often do not conform to the expected
behavior: gradient estimates poorly correlate with the true gradient, better
gradient estimates can require lower learning rates and can induce degenerate
agent behavior, value networks reduce gradient estimation variance to a
significantly smaller extent than the true value, and the underlying
optimization landscape can be misleading.

This demonstrates that there is a significant gap between the theory
inspiring current algorithms and the actual mechanisms driving their
performance. Overall, our findings suggest that developing a deep RL toolkit
that is truly robust and reliable will require moving beyond the current
benchmark-driven evaluation model to a more fine-grained understanding of deep
RL algorithms.

\section{Acknowledgements}
Work supported in part by the NSF grants CCF-1553428, CNS-1815221, the Google
PhD Fellowship, the Open Phil AI Fellowship, and the Microsoft Corporation.

\bibliography{../paper}
\bibliographystyle{iclr2020_conference}

\clearpage
\appendix
\section{Appendix}
\label{sec:appendix}

\subsection{Background}
\label{app:background}
In the reinforcement learning (RL) setting, an agent interacts with a stateful 
environment with the goal of maximizing cumulative reward. Formally, we model the environment as a
(possibly randomized) function mapping its current state 
$s$ and an action $a$ supplied by the agent to a new state $s'$
and a resulting reward $r$. The choice of actions of the agent is governed
by the its \textit{policy} $\pi$. This policy is a function
mapping
environment states to a distribution over the actions to take. The objective of an RL algorithm is to find a policy $\pi$ which maximizes the expected cumulative reward, where the expectation is taken over both environment randomness and the (randomized) action choices. 

\paragraph{Preliminaries and notation.} For a given policy $\pi$, we denote by $\pi(a|s)$ the
probability that this policy assigns to taking action $a$ when the environment is in the state
$s$. We use $r(s, a)$ to denote the reward that the agent
earns for playing action $a$ in response to the state $s$.  A {\em
trajectory} $\tau = \{(a_t, s_t): t \in \{1\ldots T\}\}$ is a sequence of
state-action pairs that constitutes a valid transcript of interactions of
the agent with the environment. (Here, $a_t$ (resp. $s_t$) corresponds to
the action taken by the agent (resp. state of the environment) in the
$t$-th round of interaction.) We then define $\pi(\tau)$ to be the
probability that the trajectory $\tau$ is executed if the agent follows
policy $\pi$ (provided the initial state of the environment is $s_1$).
Similarly, $r(\tau) = \sum_{t} r(s_t, a_t)$ denotes the cumulative reward
earned by the agent when following this trajectory, where $s_t$ (resp.
$a_t$) denote the $t$-th state (resp. action) in the trajectory $\tau$. In
the RL setting, however, we often choose to maximize the {\em discounted}
cumulative reward of a policy $R := R_1$, where $R_t$ is defined as
$$R_t(\tau) = \sum_{t'=t}^\infty \gamma^{(t'-t)} r_{t'}\;.$$
and $0<\gamma<1$ is a ``discount factor''. The discount factor ensures that the cumulative reward of a policy is well-defined even for an infinite time horizon, and it also incentivizes  
achieving reward earlier.

\paragraph{Policy gradient methods.} 
A widely used class of RL algorithms that will be the focus of our analysis
is the class of so-called \emph{policy gradient methods}. 
The central idea behind these algorithms is to first parameterize the
policy $\pi_\theta$ using a parameter vector $\theta$. (In the deep RL
context, $\pi_\theta$ is expressed by a neural network with weights
$\theta$.) Then, we perform stochastic gradient ascent on the cumulative reward
with respect to $\theta$. In other words, we want to apply the stochastic
ascent approach to our problem:
\begin{align}
\max_\theta \mathbb{E}_{\tau \sim\pi_\theta}[r(\tau)]\;,
\end{align}
where $\tau \sim \pi_\theta$ represents trajectories (rollouts) sampled from
the distribution induced by the policy $\pi_\theta$. This approach relies
on the key observation~\citep{Sutton1999PolicyGM} that
under mild
conditions, the gradient of our objective can be written as:
\begin{align}
\label{eq:pg}
\nabla_\theta \mathbb{E}_{\tau\sim \pi_\theta}[r(\tau)] &= \mathbb{E}_{\tau\sim
\pi_\theta}[\nabla_\theta\log(\pi_\theta(\tau))\ r(\tau)],
\end{align}
and the latter quantity can be estimated directly by sampling trajectories
according to the policy $\pi_\theta$. 

When we use the discounted variant of the cumulative reward and
note that the action of the policy at time $t$ cannot affect its
performance at earlier times, we can express our gradient estimate as:
\begin{align}
&\widehat{g_\theta} = \mathbb{E}_{\tau \sim \pi_\theta}
    \left[
	\sum_{(s_t, a_t) \in \tau}\nabla_\theta\log \pi_\theta(a_t|s_t)
	\cdot 
	Q_{\pi_\theta}(s_t, a_t)
    \right] \label{eqn:sr_grad} \;,
\end{align}
where $Q_{\pi_\theta}(s_t, a_t)$ represents the expected returns after taking action $a_t$ from state $s_t$:
\begin{align}
Q_{\pi_\theta}(s_t, a_t) = \mathbb{E}_{\pi_\theta}[R_t|a_t,s_t] \;.
\end{align}

\paragraph{Value estimation and advantage.} %
Unfortunately, the variance of the expectation in~\eqref{eqn:sr_grad} can
be (and often is) very large, which makes getting an accurate estimate of
this expectation quite challenging.
To alleviate this issue, a number of variance reduction techniques have been developed.
One of the most popular such techniques is the use of a so-called baseline
function, wherein a state-dependent value is subtracted from
$Q_{\pi_\theta}$.
Thus, instead of estimating~\eqref{eqn:sr_grad} directly, we use:
\begin{align}
    \widehat{g_\theta} &= \mathbb{E}_{\tau \sim
	\pi_\theta}\left[\sum_{(s_t, a_t) \in \tau} \nabla_\theta\log
\pi_\theta(a_t|s_t) \cdot (Q_{\pi_\theta}(s_t, a_t) - b(s_t))\right],
\end{align}
where $b(\cdot)$ is a baseline function of our choice.

A natural choice of the baseline function is the value function, i.e.
\begin{align}
V_{\pi_\theta}(s_t) &= \mathbb{E}_{\pi_\theta} [R_t|s_t] \;.
\end{align}
When we use the value function as our baseline, the resulting gradient estimation problem becomes:
\begin{align}
  \label{eq:vf_as_bl}
\widehat{g}_\theta &= \mathbb{E}_{\tau \sim \pi_\theta}\left[
    \sum_{(s_t, a_t) \in \tau}
    \nabla_\theta\log
\pi_\theta(a_t|s_t) \cdot A_{\pi_\theta}(s_t, a_t) \right], 
\end{align}
where
\begin{align}
  \label{eq:advantage}
A_{\pi_\theta}(s_t, a_t) & =  Q_{\pi_\theta}(s_t, a_t) - V_{\pi_\theta}(s_t)
\end{align}
is referred to as the \emph{advantage} of performing action $a_t$.
Different methods of estimating $V_{\pi_\theta}$  have been proposed, with
techniques ranging from moving averages to
the use of neural network predictors~\cite{schulman2015high}.

\paragraph{Surrogate Objective.} 
So far, our focus has been on 
extracting a good estimate of the gradient with respect to the policy
parameters $\theta$.
However, it turns out that directly optimizing the cumulative rewards can be challenging.
Thus, a modification used by modern
policy gradient algorithms is to optimize a ``surrogate objective'' instead.
We will focus on maximizing the following local approximation of the true reward~\cite{trpo}:
\begin{align}
&\max_\theta\ \mathbb{E}_{(s_t, a_t) \sim
\pi}\left[\frac{\pi_\theta(a_t|s_t)}{\pi(a_t|s_t)}A_{\pi}(s_t,
a_t)\right] \qquad \biggl(= \mathbb{E}_{\pi_\theta}\left[A_{\pi}\right]\biggr), \label{eqn:surr_rew}
\end{align}
or the normalized advantage variant proposed to reduce 
variance~\cite{schulman2017proximal}:
\begin{align}
\label{eqn:app_surrogate}
\max_\theta\ \mathbb{E}_{(s_t, a_t) \sim
	\pi}\left[\frac{\pi_\theta(a_t|s_t)}{\pi(a_t|s_t)}\widehat{A}_{\pi}(s_t, a_t)\right] 
\end{align}
where
\begin{align}
 \widehat{A}_{\pi} = \frac{A_{\pi} - \mu(A_{\pi})}{\sigma(A_{\pi})}
\end{align}
and $\pi$ is the current policy. 

\paragraph{Trust region methods.} 
The surrogate objective function, although easier to optimize, comes at a
cost: the gradient of the surrogate objective is only predictive of the policy
gradient locally (at the current policy). 
Thus, to ensure that our update steps we derive based on the surrogate objective
are predictive, they need to be
confined to a ``trust region'' around the current policy.
The resulting trust
region methods~\citep{Kakade2001ANP, trpo, schulman2017proximal} 
 try to constrain the local variation of the parameters in
policy-space by restricting the distributional distance between
successive policies.

A popular method in this class is trust region
policy optimization (TRPO)~\cite{trpo}, which
constrains the KL divergence between
successive policies on the optimization trajectory, leading to the
following problem:
\begin{align}
\max_\theta\quad &\mathbb{E}_{(s_t, a_t) \sim
\pi}\left[\frac{\pi_\theta(a_t|s_t)}{\pi(a_t|s_t)}\widehat{A}_{\pi}(s_t, a_t)\right] \nonumber \\
\text{s.t.}\quad &D_{KL}(\pi_\theta(\cdot \mid s)||\pi(\cdot\mid s)) \leq \delta,\quad \forall s\;. \label{eqn:trpo}
\end{align}
In practice, this objective is maximized using a second-order approximation
of the KL divergence and natural gradient descent, while replacing the
worst-case KL constraints over all possible states with an approximation of
the mean KL based on the states observed in the current trajectory.

\paragraph{Proximal policy optimization.}  
In practice, the TRPO algorithm can be
computationally costly---the step direction is
estimated with nonlinear conjugate gradients, which requires the
computation of multiple Hessian-vector products. To address this
issue, Schulman et al.~\cite{schulman2017proximal} propose proximal policy
optimization (PPO), which utilizes a different objective and does not
compute a projection. Concretely, PPO
proposes replacing the KL-constrained objective~\eqref{eqn:trpo} of TRPO
by clipping the objective function directly as:
\begin{align}
&\max_\theta\ \mathbb{E}_{(s_t, a_t) \sim
\pi}\left[\min\left(\text{clip}\left(\rho_t, 1-\varepsilon,
1+\varepsilon\right)\widehat{A}_{\pi}(s_t, a_t),\ 
\rho_t\widehat{A}_{\pi}(s_t, a_t)\right)\right] \label{eqn:ppo}
\end{align}
where
\begin{align}
\rho_t = \frac{\pi_\theta(a_t|s_t)}{\pi(a_t|s_t)}
\end{align}
In addition to being simpler, PPO is intended to be faster and
more sample-efficient than TRPO~\citep{schulman2017proximal}.

\clearpage

\subsection{Experimental Setup}
We use the following parameters for PPO and TRPO based on a
hyperparameter grid search:

\begin{table}[h]
\centering
\caption{Hyperparameters for PPO and TRPO algorithms.}
\label{app-tab:hyperparameters}
\begin{tabular}{@{}lllllll@{}}
	\toprule
	{} & \multicolumn{2}{c}{\bf Humanoid-v2} & \multicolumn{2}{c}{\bf
	Walker2d-v2} & \multicolumn{2}{c}{\bf Hopper-v2} \\ 
	{} &            PPO &      TRPO &            PPO &      TRPO &            PPO &      TRPO \\
	\midrule
	Timesteps per iteration          &           2048 &      2048 &           2048 &      2048 &           2048 &      2048 \\
	Discount factor ($\gamma$)         &           0.99 &      0.99 &           0.99 &      0.99 &           0.99 &      0.99 \\
	GAE discount ($\lambda$)           &           0.95 &      0.95 &           0.95 &      0.95 &           0.95 &      0.95 \\
	Value network LR                 &         0.0001 &    0.0003 &         0.0003 &    0.0003 &         0.0002 &    0.0002 \\
	Value net num. epochs        &             10 &        10 &             10 &        10 &             10 &        10 \\
	Policy net hidden layers     &       [64, 64] &  [64, 64] &       [64, 64] &  [64, 64] &       [64, 64] &  [64, 64] \\
	Value net hidden layers      &       [64, 64] &  [64, 64] &       [64,
	64] &  [64, 64] &       [64, 64] &  [64, 64] \\ \midrule
	KL constraint ($\delta$)          &            N/A &      0.07 &            N/A &      0.04 &            N/A &      0.13 \\
	Fisher est. fraction       &            N/A &       0.1 &            N/A &       0.1 &            N/A &       0.1 \\
	Conjugate grad. steps         &            N/A &        10 &            N/A &        10 &            N/A &        10 \\
	CG damping       &            N/A &       0.1 &            N/A &       0.1 &            N/A &       0.1 \\
	Backtracking steps               &            N/A &        10 &
	N/A &        10 &            N/A &        10 \\ \midrule
	Policy LR (Adam)                 &        0.00025 &       N/A &         0.0004 &       N/A &        0.00045 &       N/A \\
	Policy epochs                    &             10 &       N/A &             10 &       N/A &             10 &       N/A \\
	PPO Clipping $\varepsilon$        &            0.2 &       N/A &            0.2 &       N/A &            0.2 &       N/A \\
	Entropy coeff.                   &            0.0 &       0.0 &
	0.0 &       0.0 &            0.0 &       0.0 \\ 
	Reward clipping                  &  [-10, 10] &        -- &  [-10,
	10] &        -- &  [-10, 10] &        -- \\
	Reward normalization             &            On &       Off &
	On &      Off &            On &       Off \\ 
	State clipping                   &  [-10, 10] &        -- &  [-10, 10] &        -- &  [-10, 10] &        -- \\
	\bottomrule
	\end{tabular}
\end{table}

All error bars we plot are 95\% confidence intervals, obtained via
bootstrapped sampling.

\clearpage
\subsection{Standard Reward Plots}
\begin{figure}[!h]
    \centering
	\begin{subfigure}[b]{0.6\textwidth}
		\centering
		\includegraphics[width=1\textwidth]{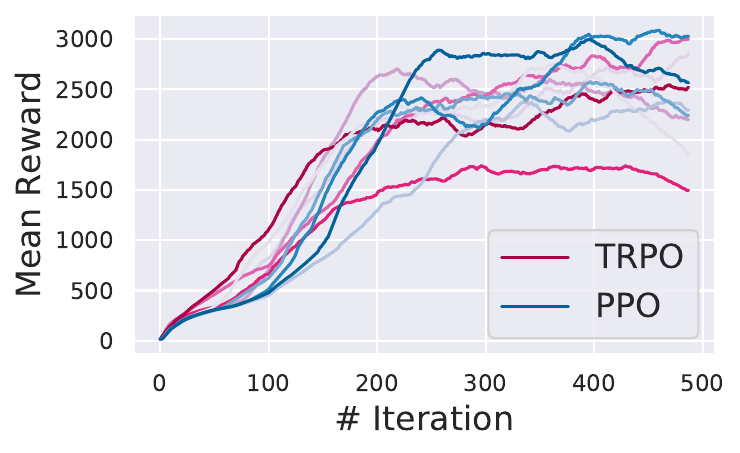}
		\caption{Hopper-v2}
	\end{subfigure}
	\begin{subfigure}[b]{0.6\textwidth}
		\centering
		\includegraphics[width=1\textwidth]{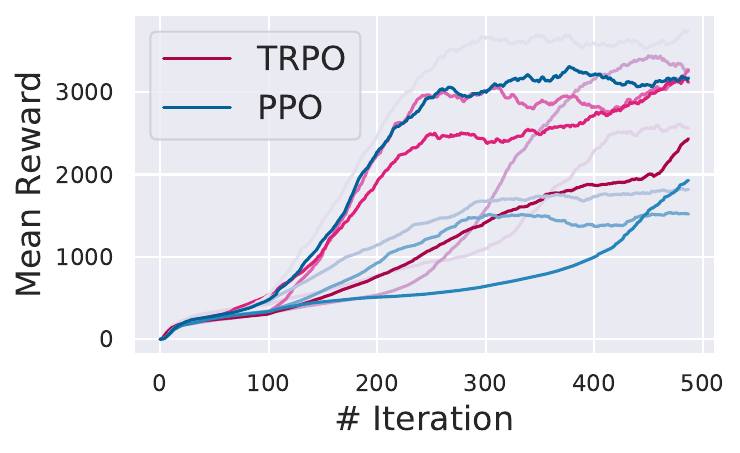} 
		\caption{Walker2d-v2}
	\end{subfigure}
	\begin{subfigure}[b]{0.6\textwidth}
		\centering
		\includegraphics[width=1\textwidth]{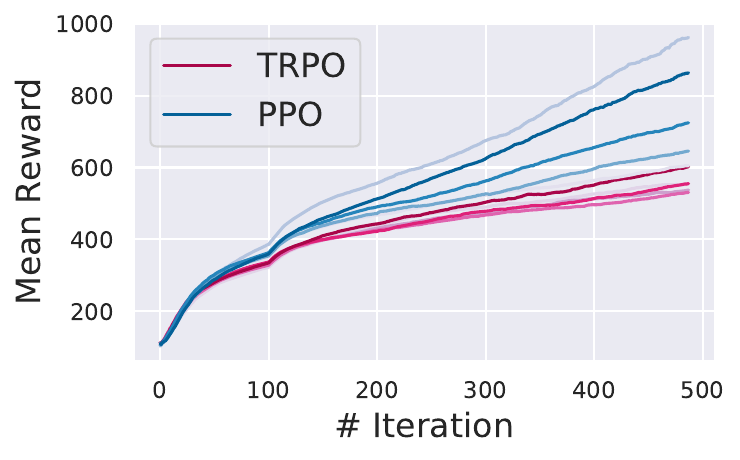} 
		\caption{Humanoid-v2}
	\end{subfigure}
	\caption{Mean reward for the studied policy gradient algorithms on
    standard MuJoCo benchmark tasks. For each algorithm, we perform $24$
    random trials using the best performing hyperparameter configuration,
    with $10$ of the random agents shown here.}
	\label{fig:reward}
\end{figure}

\clearpage
\subsection{Quality of Gradient Estimation}
\label{app:gradest}

\begin{figure}[!h]
	\begin{subfigure}[b]{1\textwidth}
		\includegraphics[width=1\textwidth]{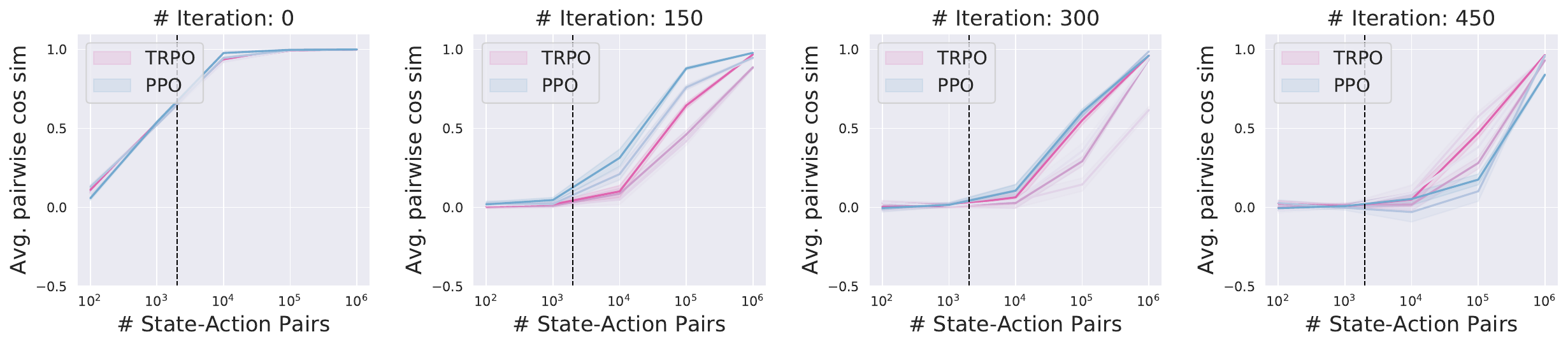} 
		\caption{Walker2d-v2}
	\end{subfigure}
	\begin{subfigure}[b]{1\textwidth}
		\includegraphics[width=1\textwidth]{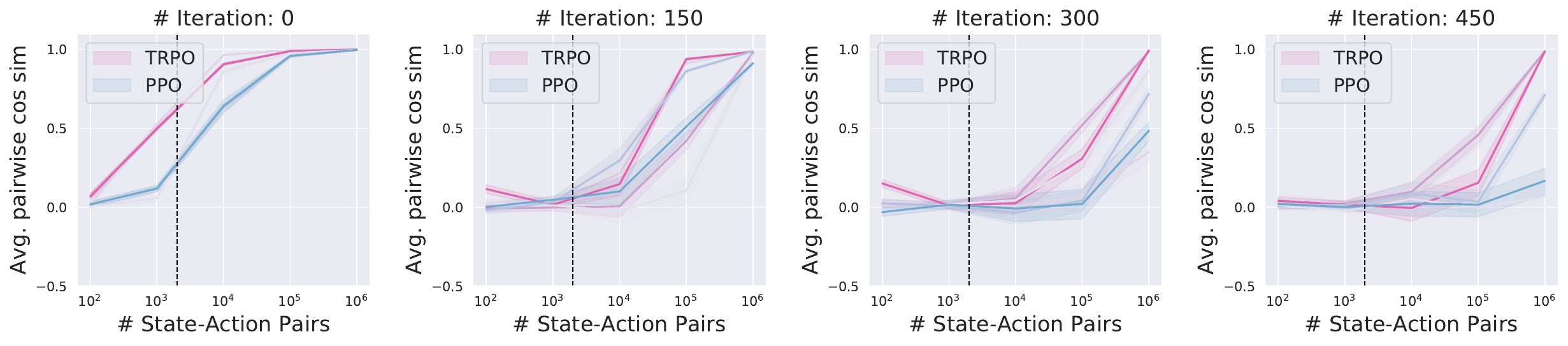} 
		\caption{Hopper-v2}
	\end{subfigure}
	\caption{Empirical variance of the gradient (c.f.~\eqref{eqn:grad_sr})
        as a function of the number of state-action pairs used in
        estimation for policy gradient methods.
		We obtain multiple  gradient estimates using a given number of
        state-action pairs from the policy at a particular iteration.
		We then measure the average pairwise cosine similarity between
        these repeated gradient measurements, along with the $95\%$
        confidence intervals (shaded).
		Each of the colored lines (for a specific algorithm) represents a
        particular trained agent (we perform multiple trials with the same
        hyperparameter configurations but different random seeds).  
		The dotted vertical black line (at $2$K) indicates the sample
        regime used for gradient estimation in standard practical
        implementations of policy gradient methods.}
	\label{fig:gradvar_app}
\end{figure}

\begin{figure}[!h]
	\begin{subfigure}[b]{1\textwidth}
		\includegraphics[width=1\textwidth]{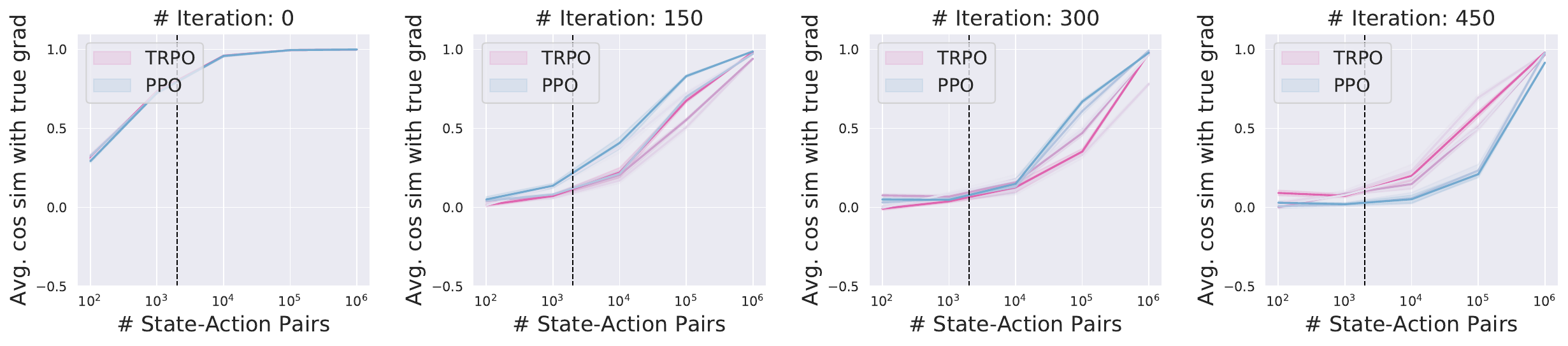} 
		\caption{Walker2d-v2}
	\end{subfigure}
	\begin{subfigure}[b]{1\textwidth}
		\includegraphics[width=1\textwidth]{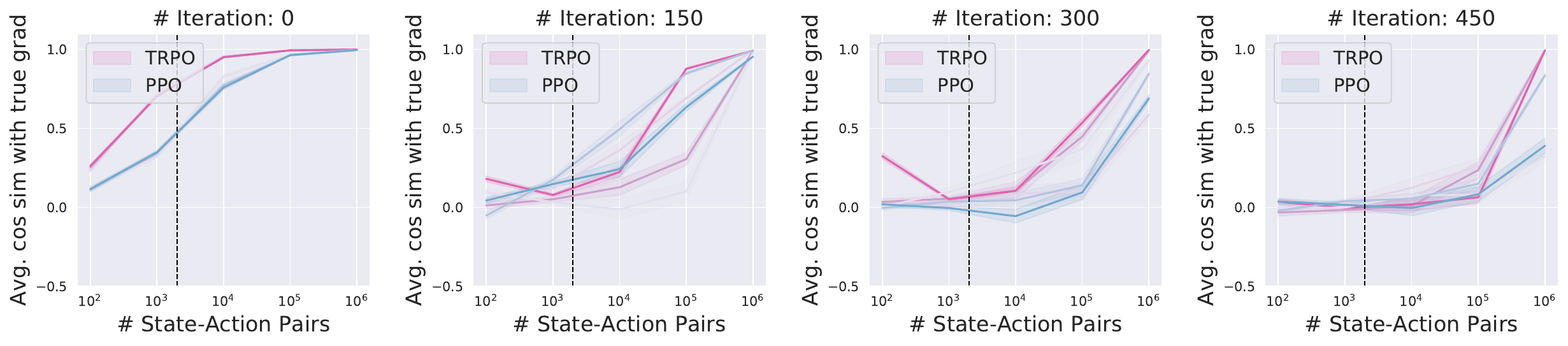} 
		\caption{Hopper-v2}
	\end{subfigure}
	\caption{Convergence of gradient estimates to the ``true'' expected gradient (c.f.~\eqref{eqn:grad_sr}).
		We measure the cosine similarity between the true gradient (approximated using around $1$M samples) and gradient estimates, as a function of number of state-action pairs used to obtain the later.
		For a particular policy and state-action pair count, we obtain multiple estimates of this cosine similarity and then report the average, along with the $95\%$ confidence intervals (shaded).
		Each of the colored lines (for a specific algorithm) represents a particular trained agent (we perform multiple trials with the same hyperparameter configurations but different random seeds).  
		The dotted vertical black line (at $2$K) indicates the sample regime used for gradient estimation in standard practical implementations of policy gradient methods.}
	\label{fig:truegrad_app}
\end{figure}

\clearpage

\subsection{Value Prediction}
\label{app:value_pred}

\begin{figure}[!htb]
	\begin{subfigure}[b]{1\textwidth}
		\centering
		\includegraphics[width=0.9\textwidth]{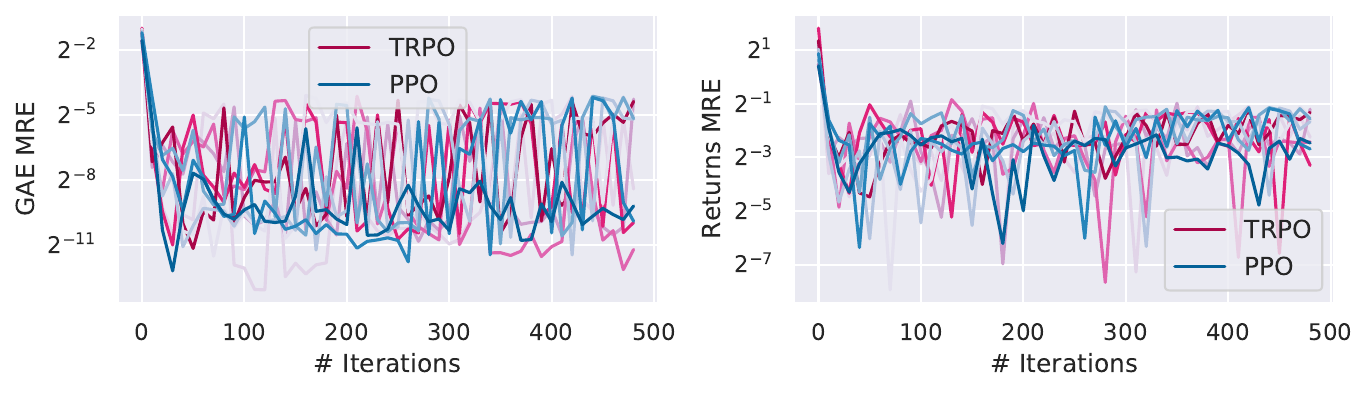} 
		\caption{Hopper-v2}
	\end{subfigure}
	\begin{subfigure}[b]{1\textwidth}
		\centering
	\includegraphics[width=0.9\textwidth]{Figures/value/Value_Hopper-v2_good_train} 
	\caption{Walker2d-v2}
	\end{subfigure}
	\begin{subfigure}[b]{1\textwidth}
		\centering
		\includegraphics[width=0.9\textwidth]{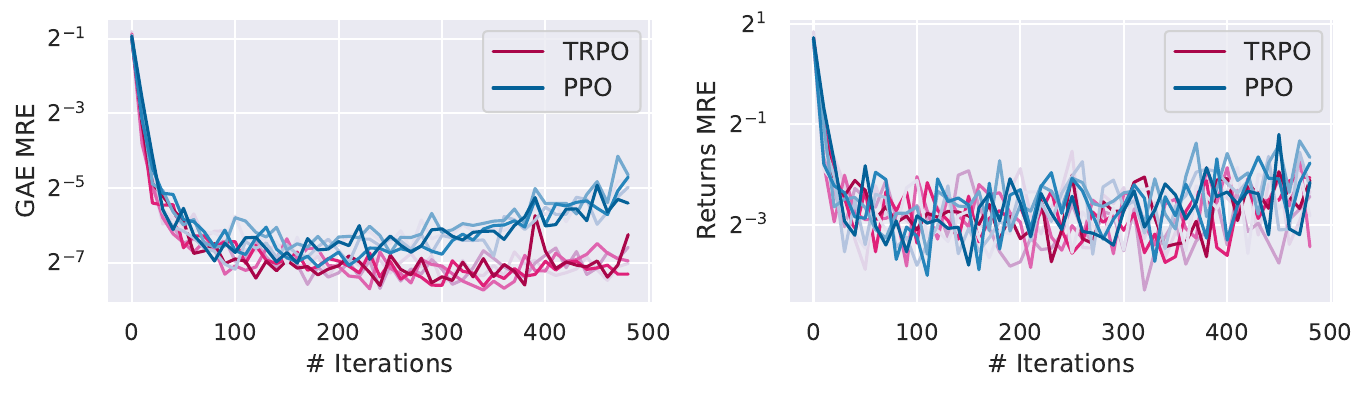} 
		\caption{Humanoid-v2}
	\end{subfigure}
	\caption{Quality of value prediction in terms of mean relative error (MRE) on train state-action pairs for agents trained to solve the MuJoCo tasks.
	We see in that the agents do indeed succeed at solving
	the supervised learning task they are trained for -- the
	train MRE on the
	GAE-based value loss $(V_{old} + A_{GAE})^2$ 
	(c.f.~\eqref{eq:val_targ}) is small (left column).
	We observe that the returns MRE is  quite small as well (right column).
	}
	\label{fig:val_app_train}
\end{figure}

\begin{figure}[tb]
	\begin{subfigure}[b]{1\textwidth}
		\includegraphics[width=1\textwidth]{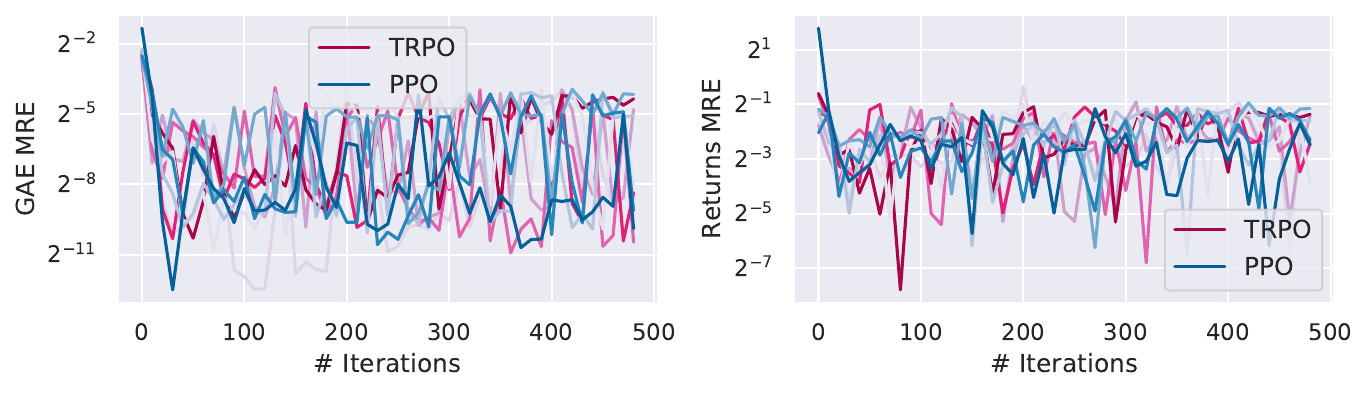} 
		\caption{Hopper-v2}
	\end{subfigure}
\begin{subfigure}[b]{1\textwidth}
	\includegraphics[width=1\textwidth]{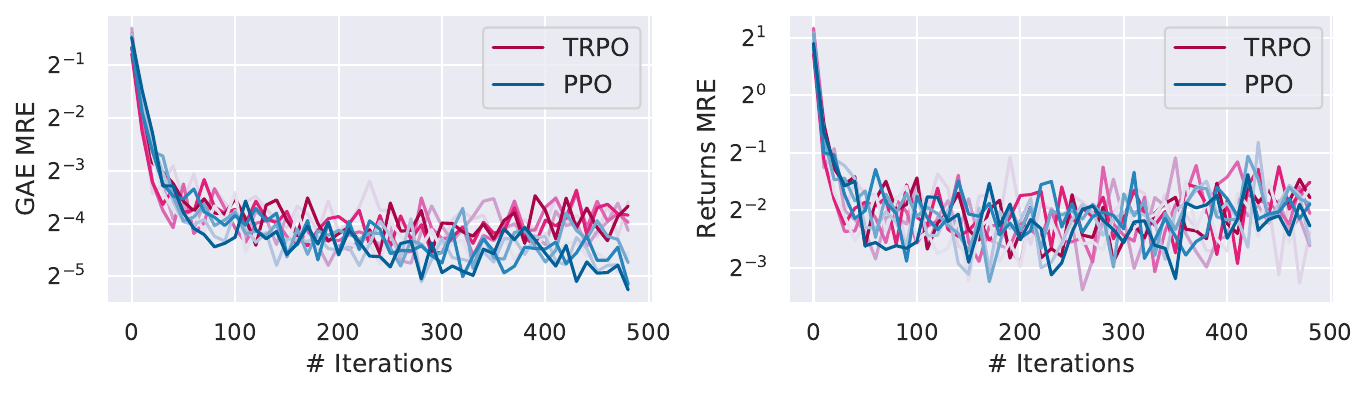} 
	\caption{Humanoid-v2}
\end{subfigure}
	\caption{Quality of value prediction in terms of mean relative error (MRE) on heldout state-action pairs for agents trained to solve MuJoCo tasks.
		We see in that the agents do indeed succeed at solving
		the supervised learning task they are trained for -- the
		validation MRE on the
		GAE-based value loss $(V_{old} + A_{GAE})^2$ 
		(c.f.~\eqref{eq:val_targ}) is small (left column).
		On the other hand, we see that the returns MRE is
		still quite high -- the learned value function is off by
		about $50\%$ with respect to the underlying true value
		function (right column).
	}
	\label{fig:val_app_heldout}
\end{figure}

\clearpage
\subsection{Optimization Landscape}
\begin{figure}[htp]
\begin{tabular}{cc|c}
& Few state-action pairs (2,000) & Many state-action pairs ($10^6$) \\
& \begin{tabularx}{.45\textwidth}{CC} Surrogate & True reward \end{tabularx}
& \begin{tabularx}{.45\textwidth}{CC} Surrogate & True reward \end{tabularx} \\
\raisebox{1.5cm}{\rotatebox[origin=c]{90}{Step 0}} &
    \includegraphics[width=.45\textwidth]{Figures/landscapes/humanoid_bad_0_ppo-all-hacks} &
    \includegraphics[width=.45\textwidth]{Figures/landscapes/humanoid_good_0_ppo-all-hacks} \\
\raisebox{1.5cm}{\rotatebox[origin=c]{90}{Step 150}} &
    \includegraphics[width=.45\textwidth]{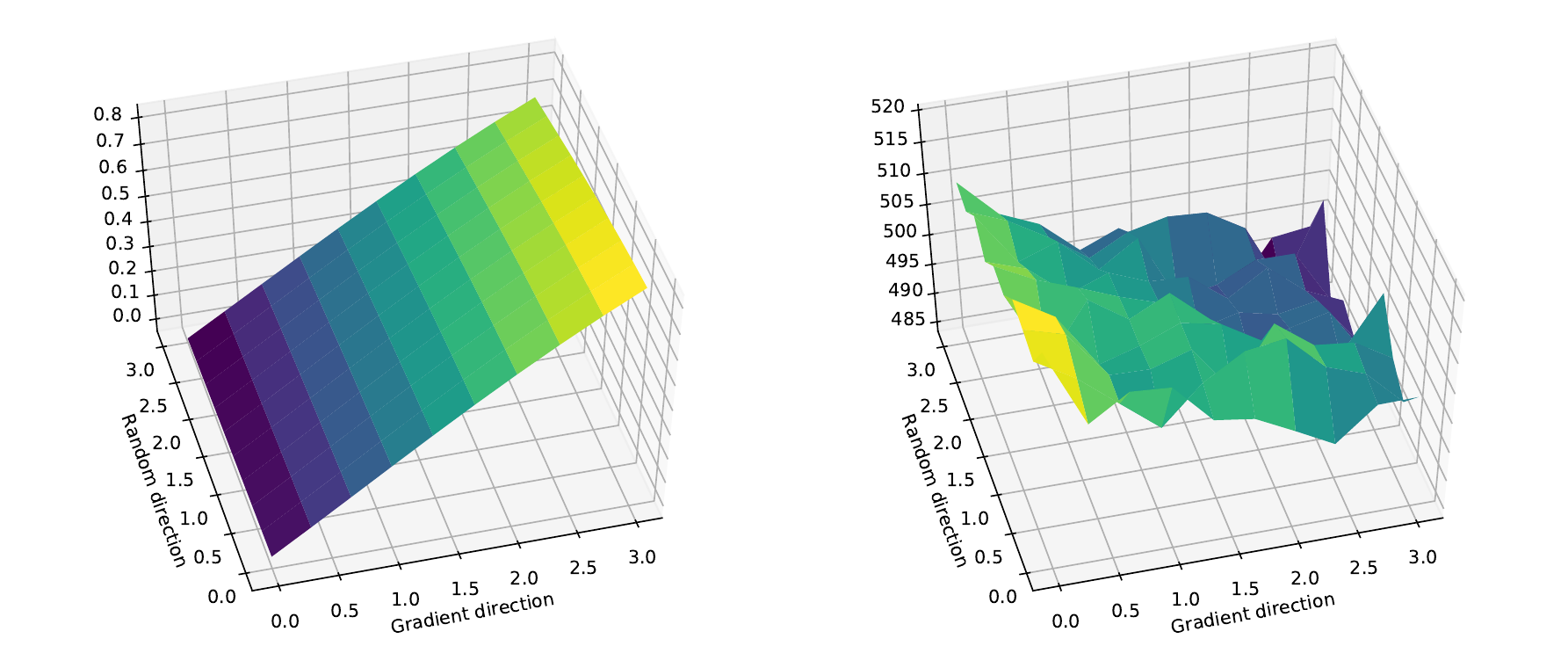} &
    \includegraphics[width=.45\textwidth]{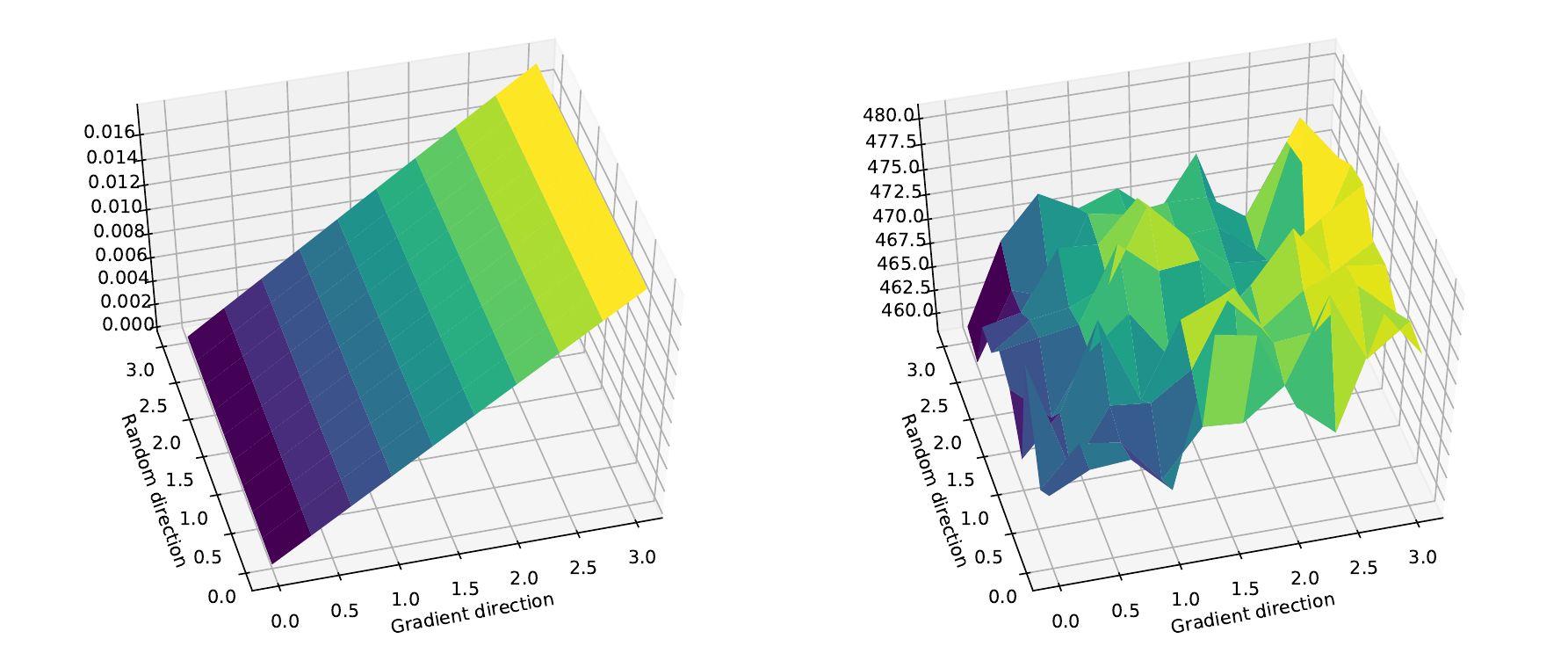} \\
\raisebox{1.5cm}{\rotatebox[origin=c]{90}{Step 300}} &
    \includegraphics[width=.45\textwidth]{Figures/landscapes/humanoid_bad_300_ppo-all-hacks} &
    \includegraphics[width=.45\textwidth]{Figures/landscapes/humanoid_good_300_ppo-all-hacks} \\
\raisebox{1.5cm}{\rotatebox[origin=c]{90}{Step 450}} &
    \includegraphics[width=.45\textwidth]{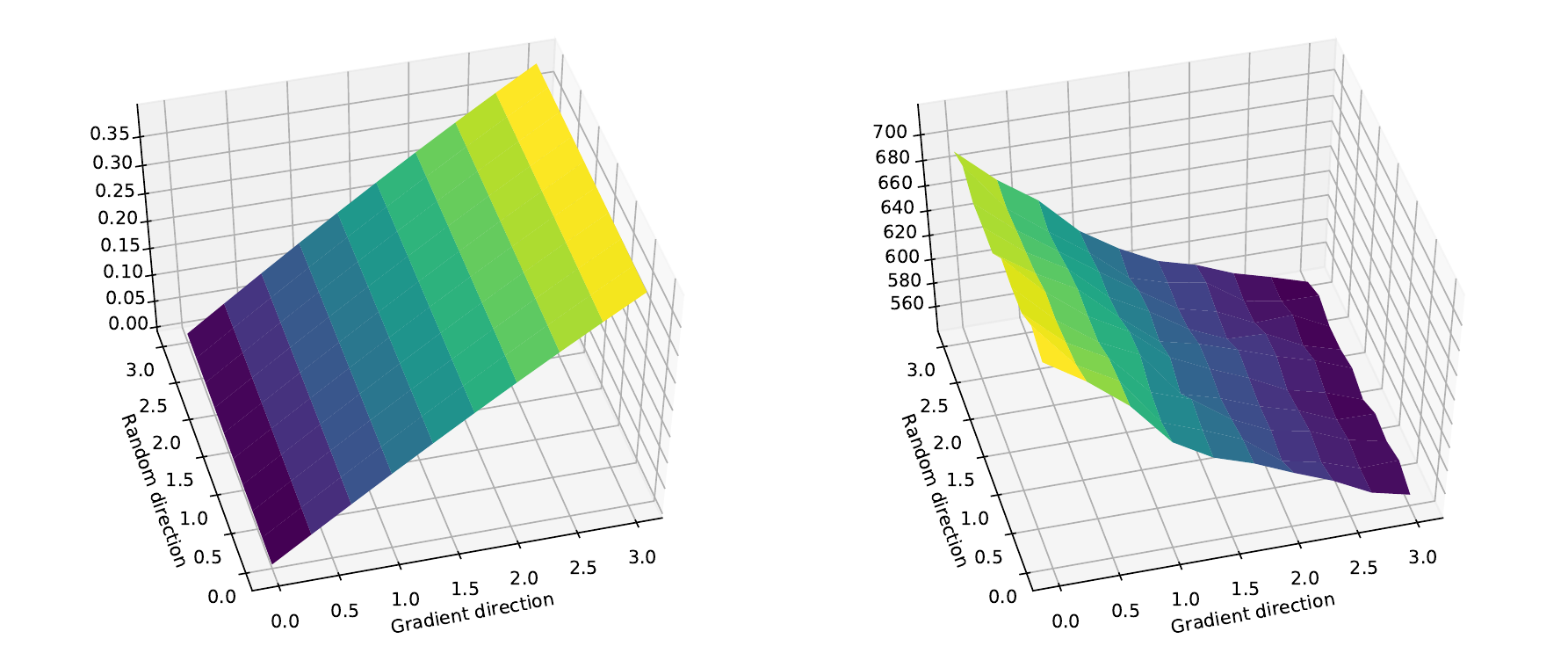} &
    \includegraphics[width=.45\textwidth]{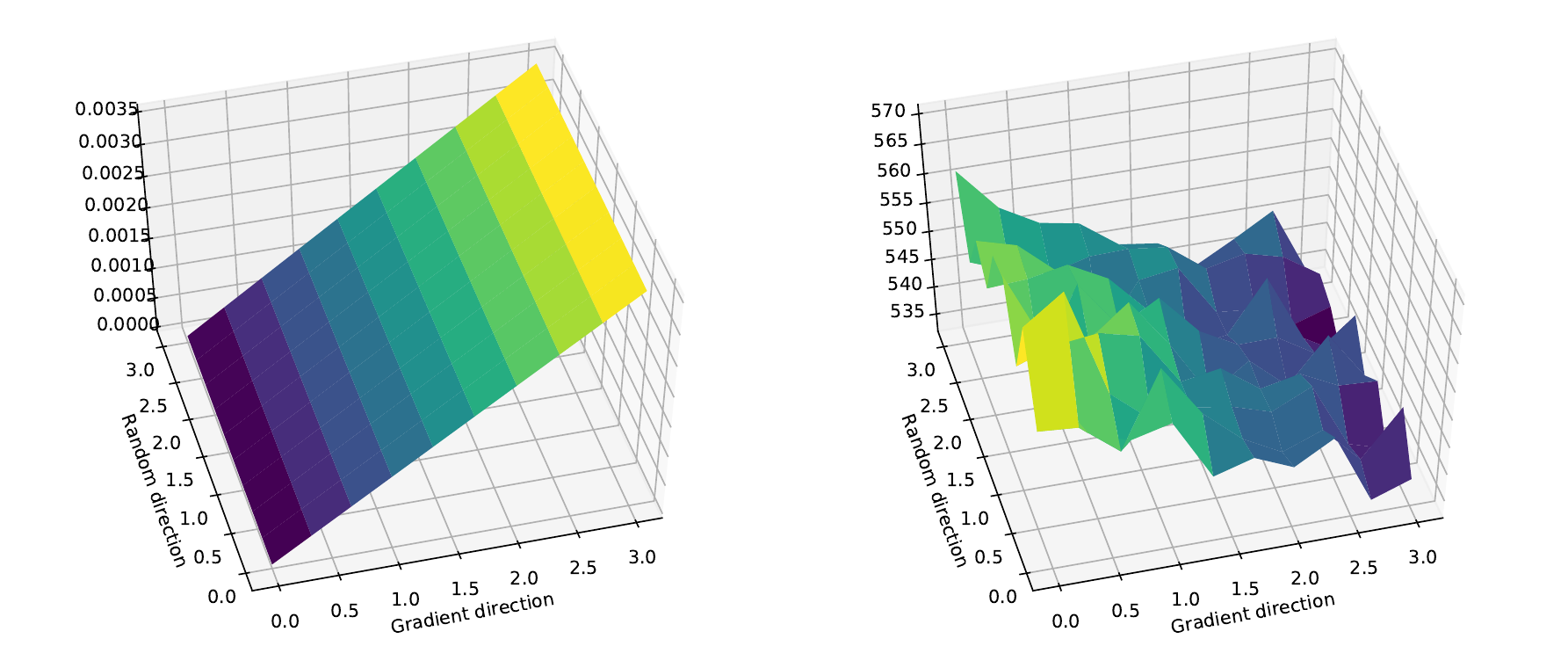}
\end{tabular}
\caption{Humanoid-v2 -- PPO reward landscapes.}
\label{fig:humanoid_landscape_ppo}
\end{figure}

\begin{figure}[htp]
\begin{tabular}{cc|c}
& Few state-action pairs (2,000) & Many state-action pairs ($10^6$) \\
& \begin{tabularx}{.45\textwidth}{CC} Surrogate & True reward \end{tabularx}
& \begin{tabularx}{.45\textwidth}{CC} Surrogate & True reward \end{tabularx} \\
\raisebox{1.5cm}{\rotatebox[origin=c]{90}{Step 0}} &
    \includegraphics[width=.45\textwidth]{Figures/landscapes/humanoid_bad_0_trpo-no-hacks} &
    \includegraphics[width=.45\textwidth]{Figures/landscapes/humanoid_good_0_trpo-no-hacks} \\
\raisebox{1.5cm}{\rotatebox[origin=c]{90}{Step 150}} &
    \includegraphics[width=.45\textwidth]{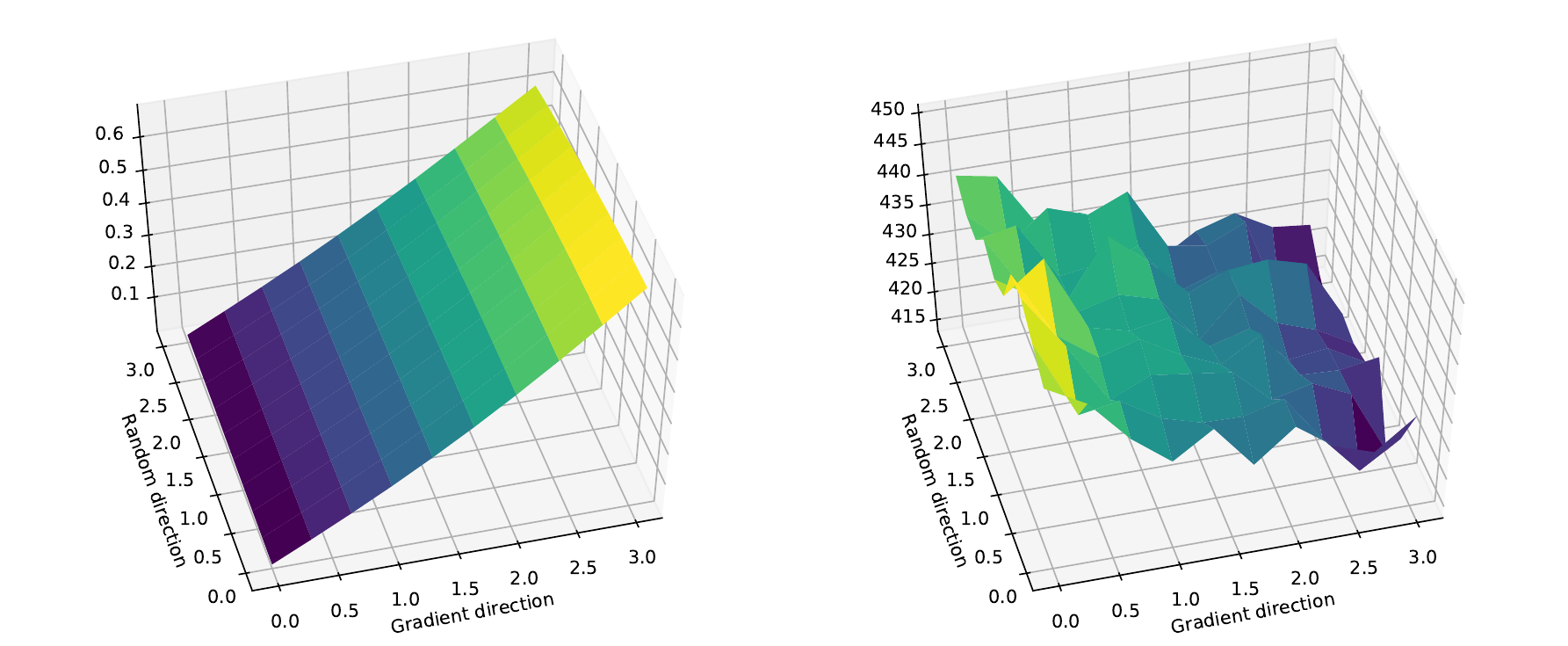} &
    \includegraphics[width=.45\textwidth]{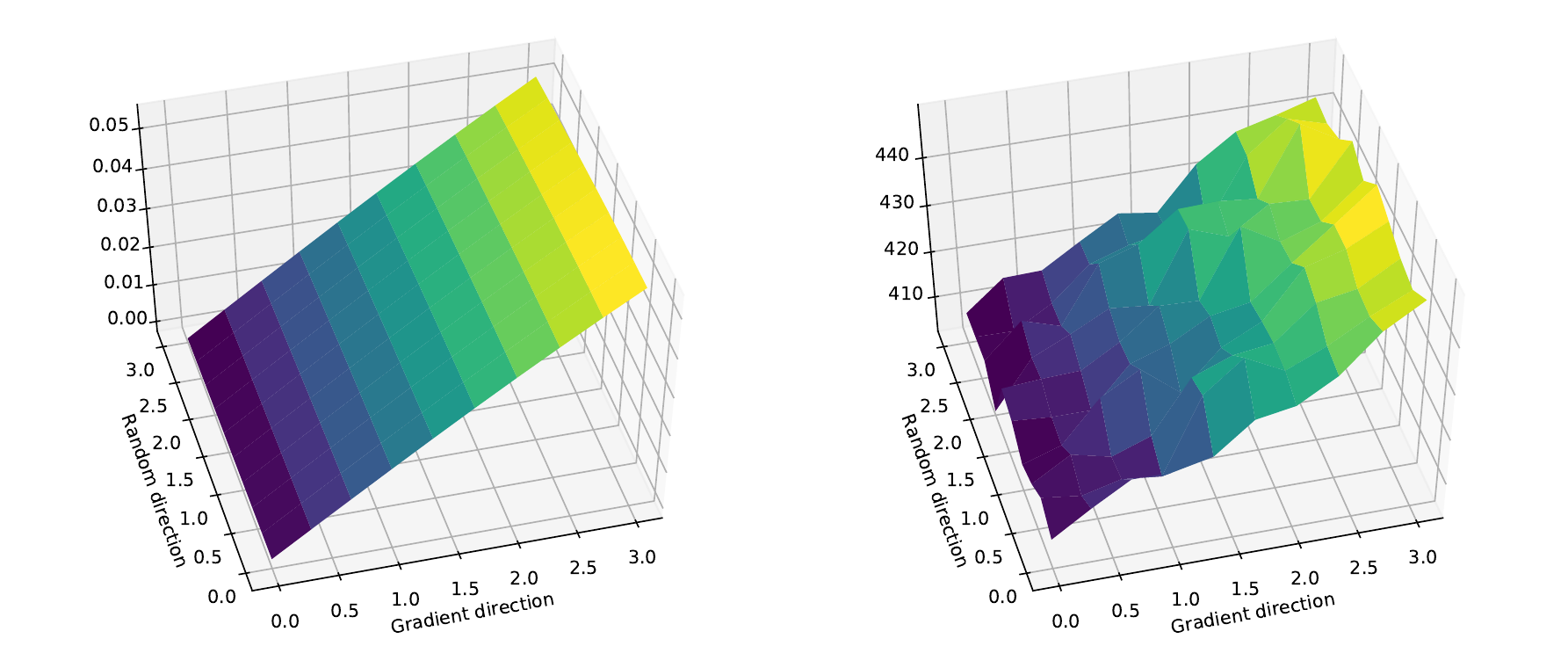} \\
\raisebox{1.5cm}{\rotatebox[origin=c]{90}{Step 300}} &
    \includegraphics[width=.45\textwidth]{Figures/landscapes/humanoid_bad_300_trpo-no-hacks} &
    \includegraphics[width=.45\textwidth]{Figures/landscapes/humanoid_good_300_trpo-no-hacks} \\
\raisebox{1.5cm}{\rotatebox[origin=c]{90}{Step 450}} &
    \includegraphics[width=.45\textwidth]{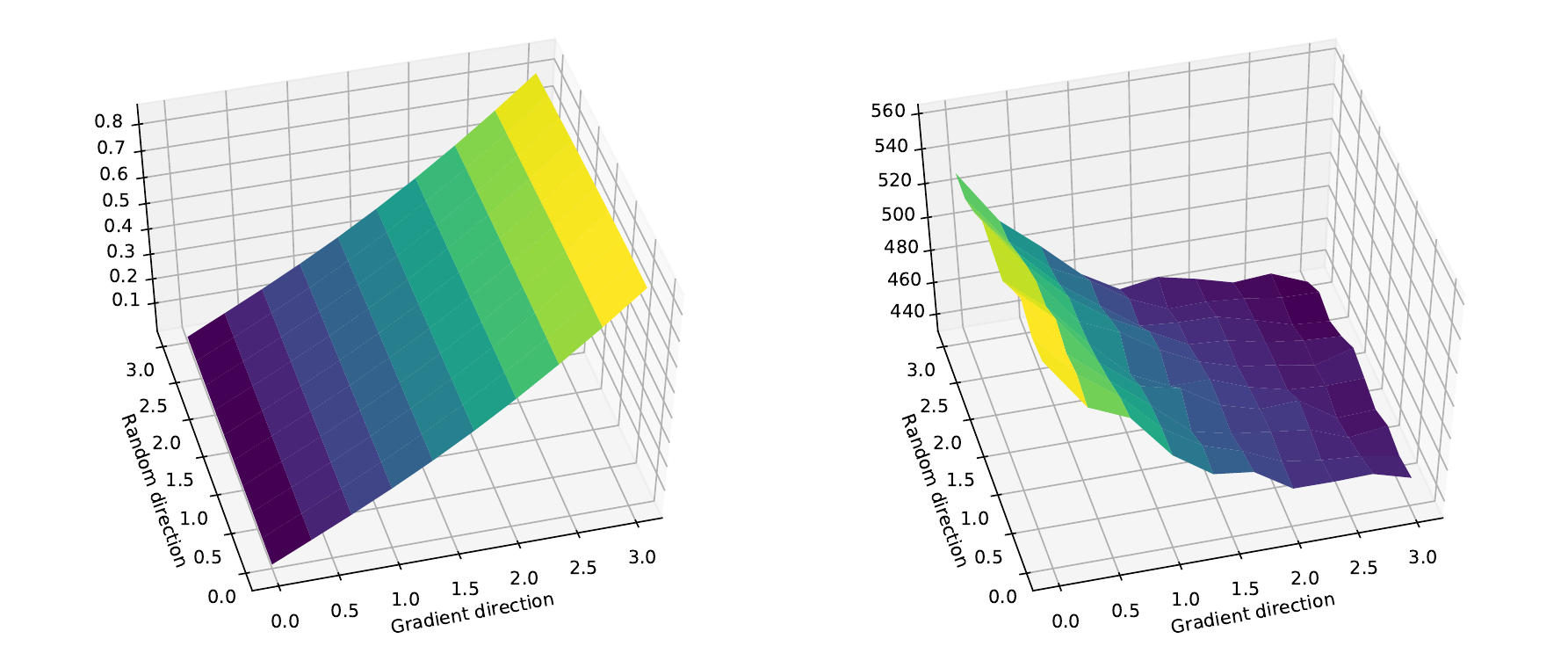} &
    \includegraphics[width=.45\textwidth]{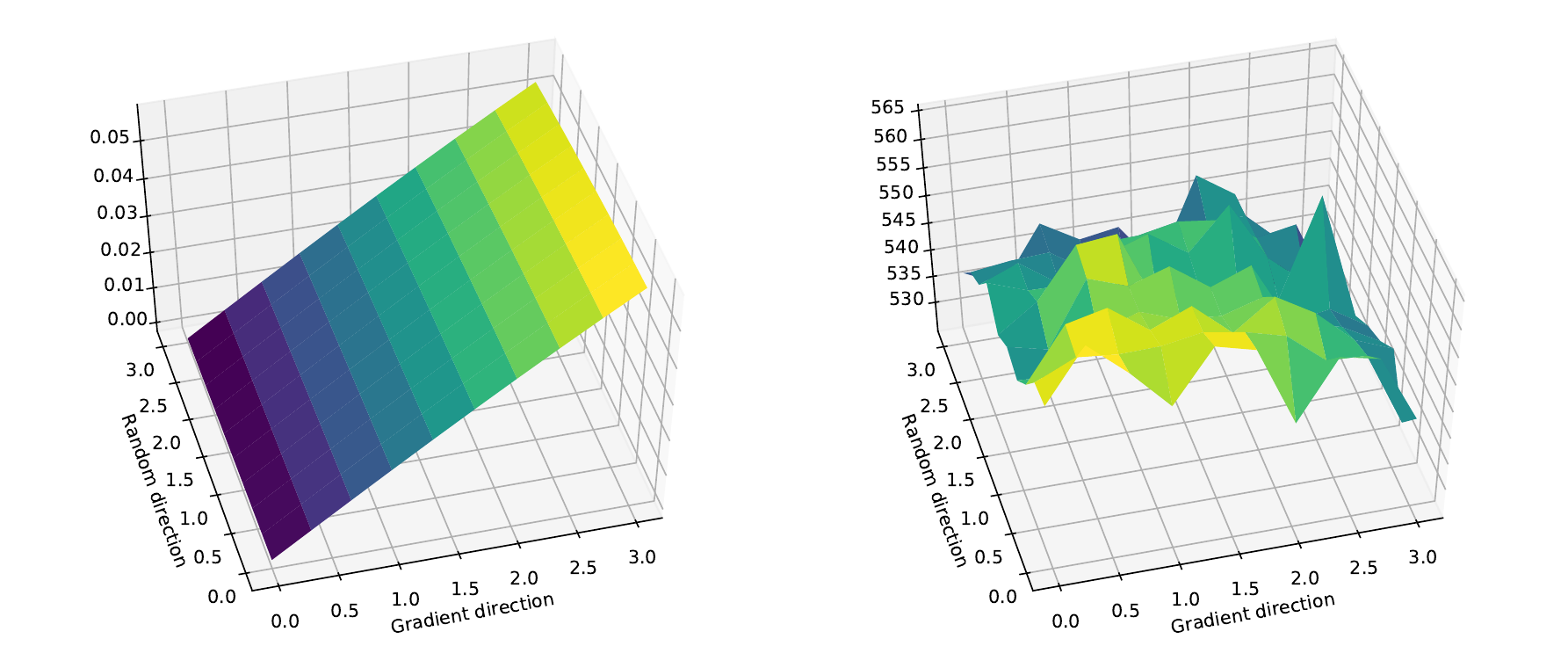}
\end{tabular}
\caption{Humanoid-v2 -- TRPO reward landscapes.}
\label{fig:humanoid_landscape_trpo}
\end{figure}

\begin{figure}[htp]
\begin{tabular}{cc|c}
& Few state-action pairs (2,000) & Many state-action pairs ($10^6$) \\
& \begin{tabularx}{.45\textwidth}{CC} Surrogate & True reward \end{tabularx}
& \begin{tabularx}{.45\textwidth}{CC} Surrogate & True reward \end{tabularx} \\
\raisebox{1.5cm}{\rotatebox[origin=c]{90}{Step 0}} &
    \includegraphics[width=.45\textwidth]{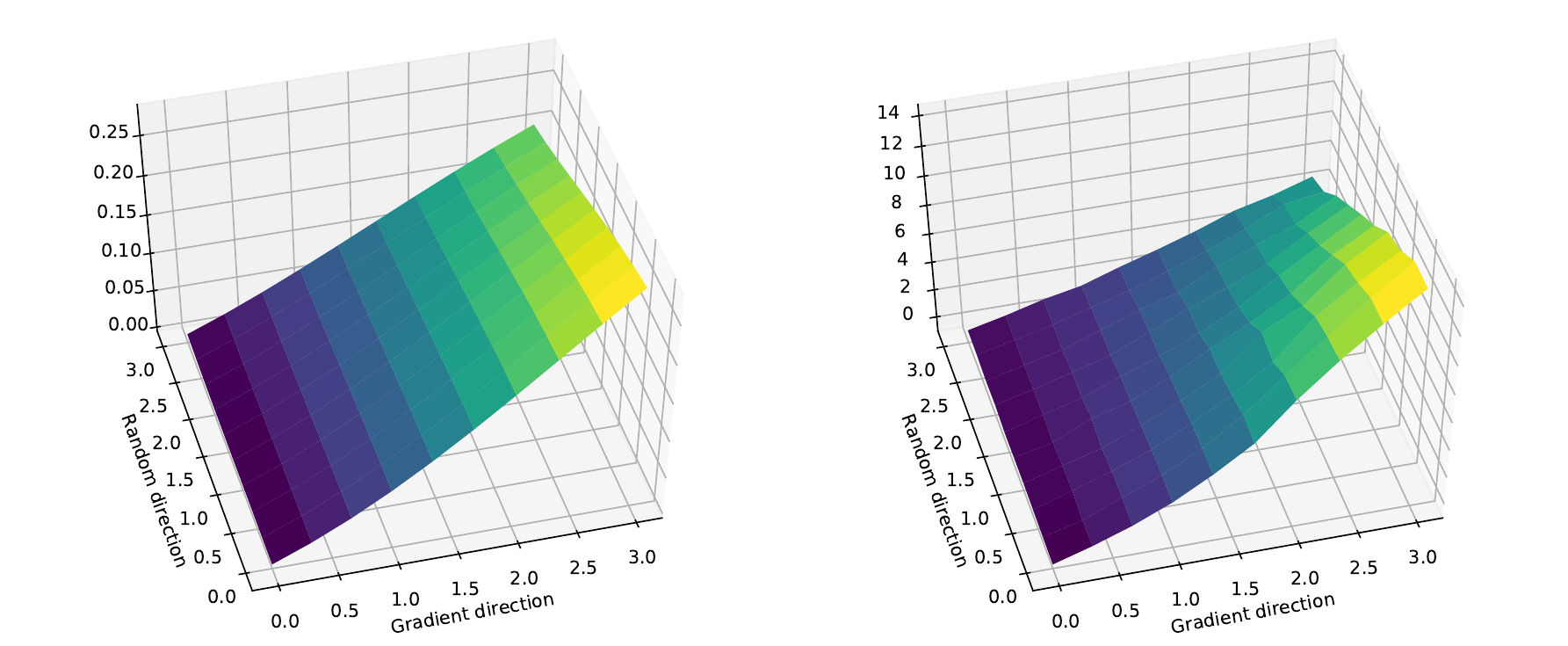} &
    \includegraphics[width=.45\textwidth]{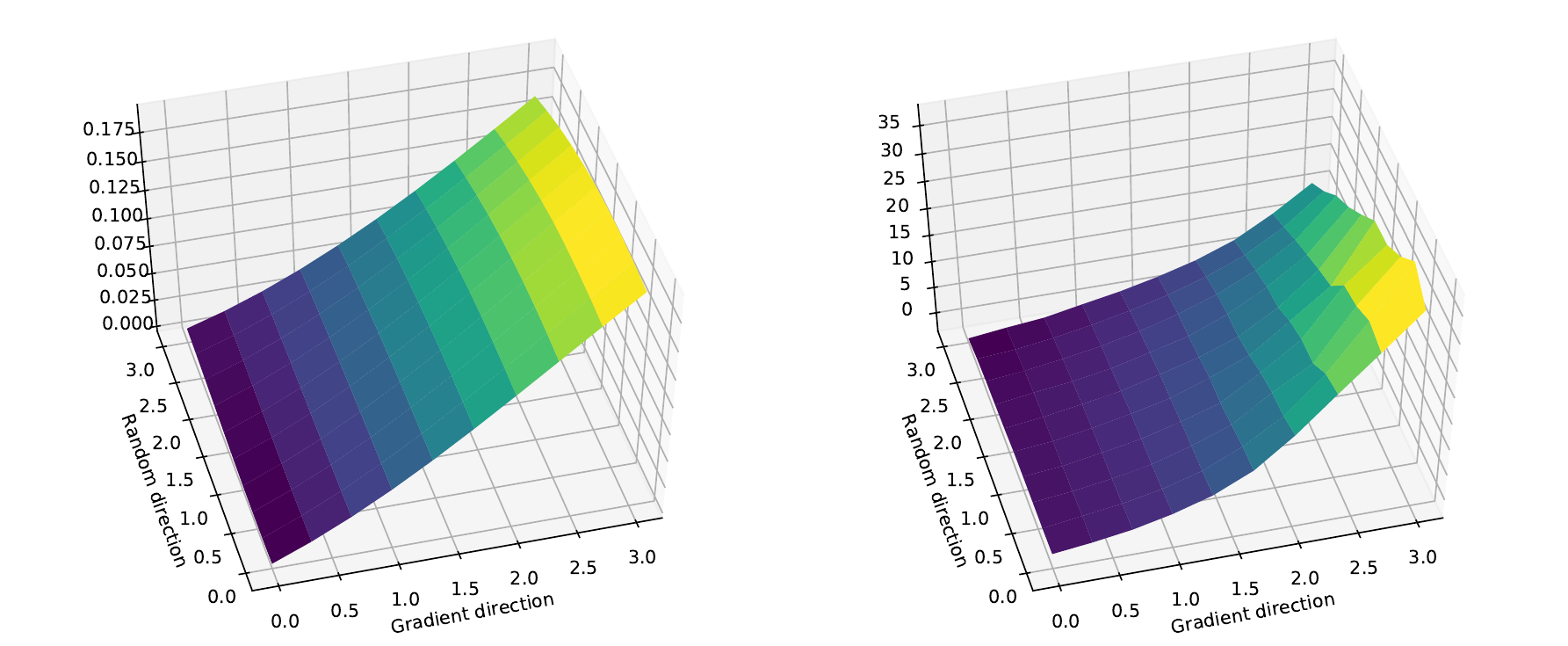} \\
\raisebox{1.5cm}{\rotatebox[origin=c]{90}{Step 150}} &
    \includegraphics[width=.45\textwidth]{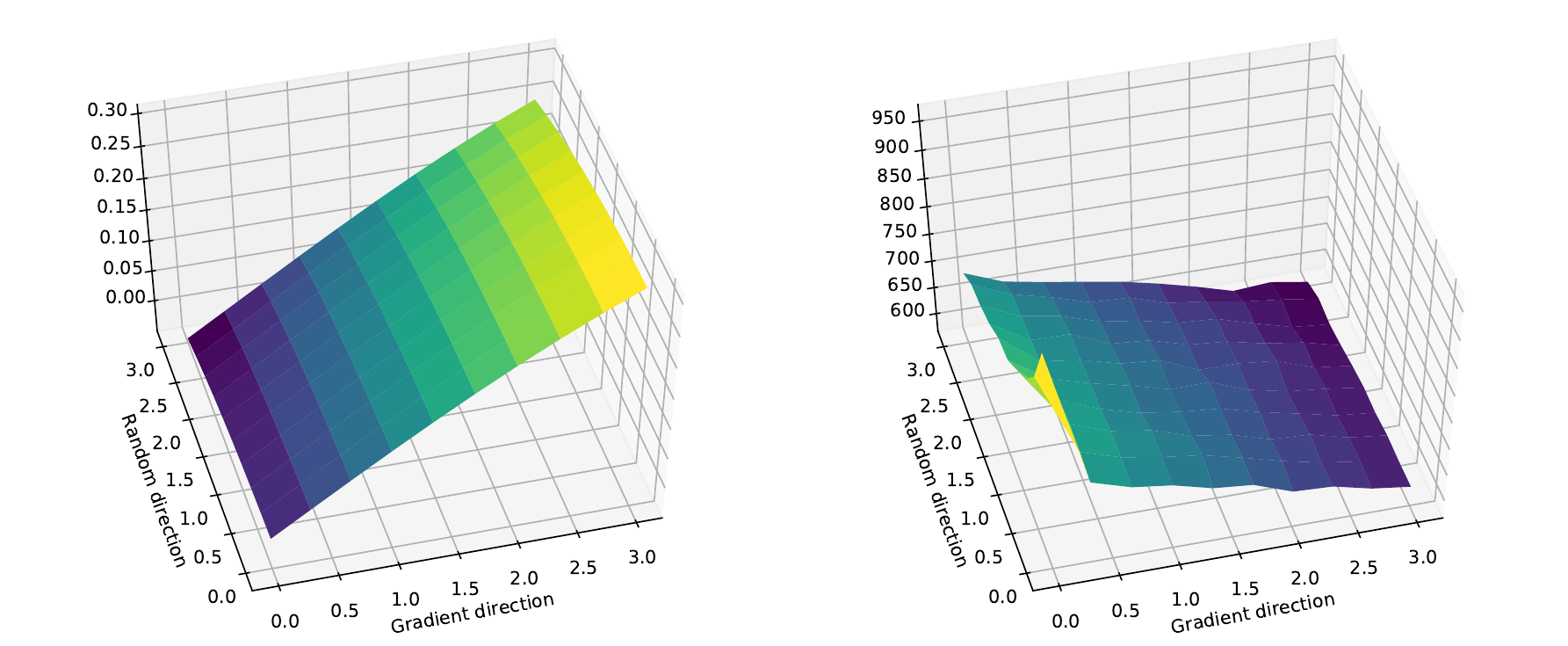} &
    \includegraphics[width=.45\textwidth]{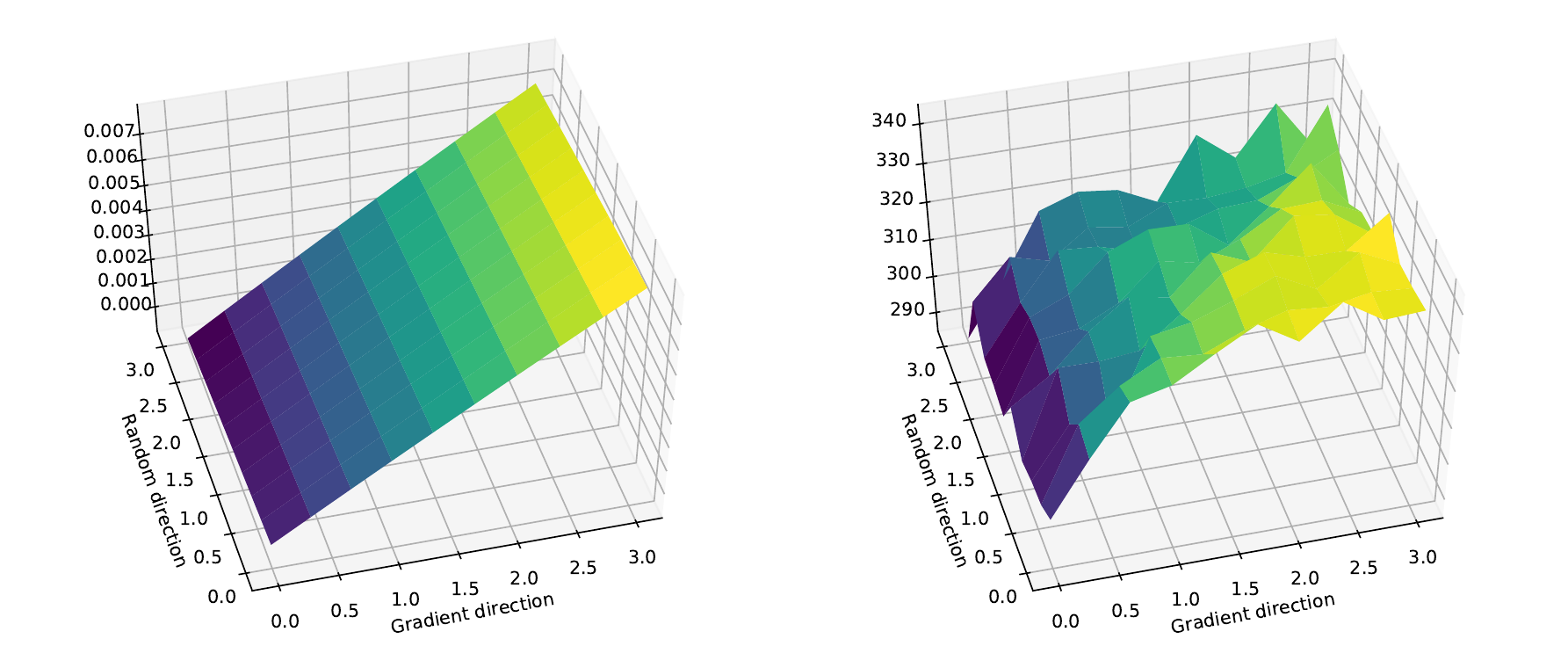} \\
\raisebox{1.5cm}{\rotatebox[origin=c]{90}{Step 300}} &
    \includegraphics[width=.45\textwidth]{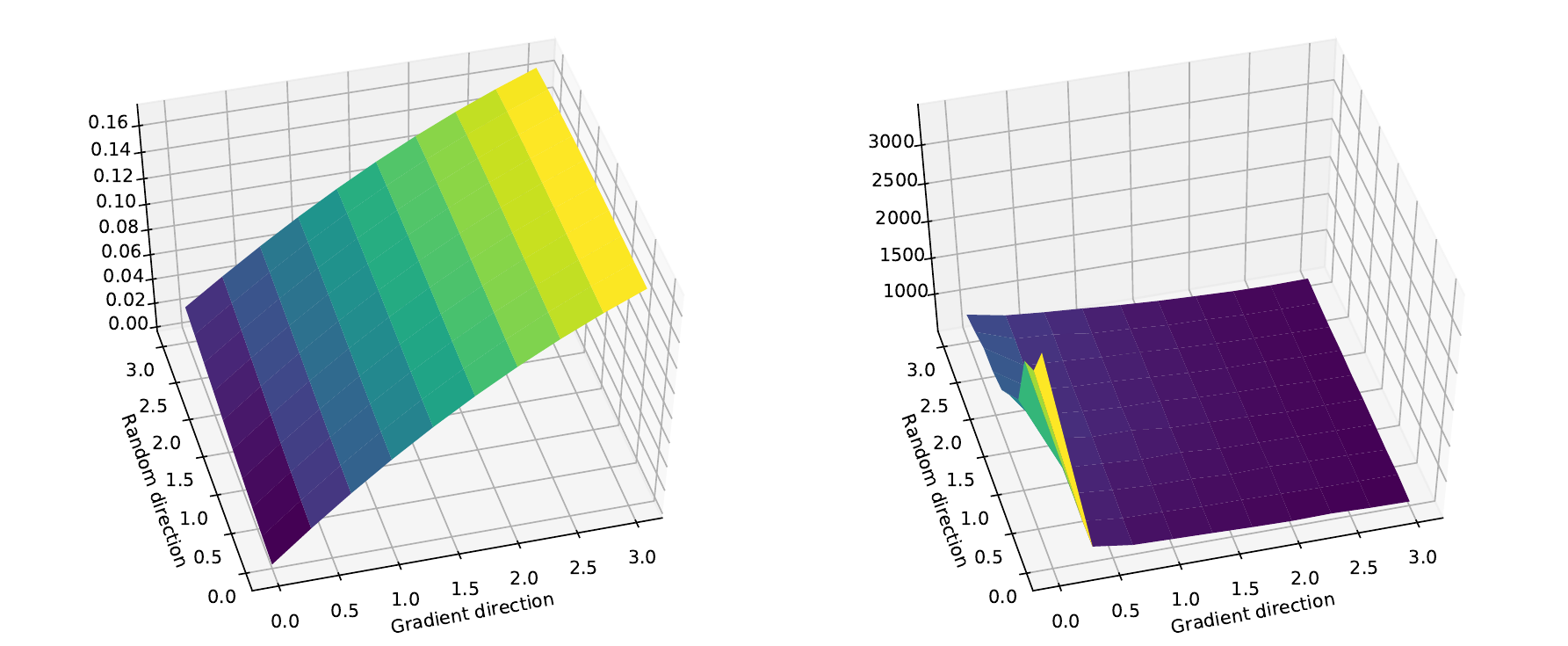} &
    \includegraphics[width=.45\textwidth]{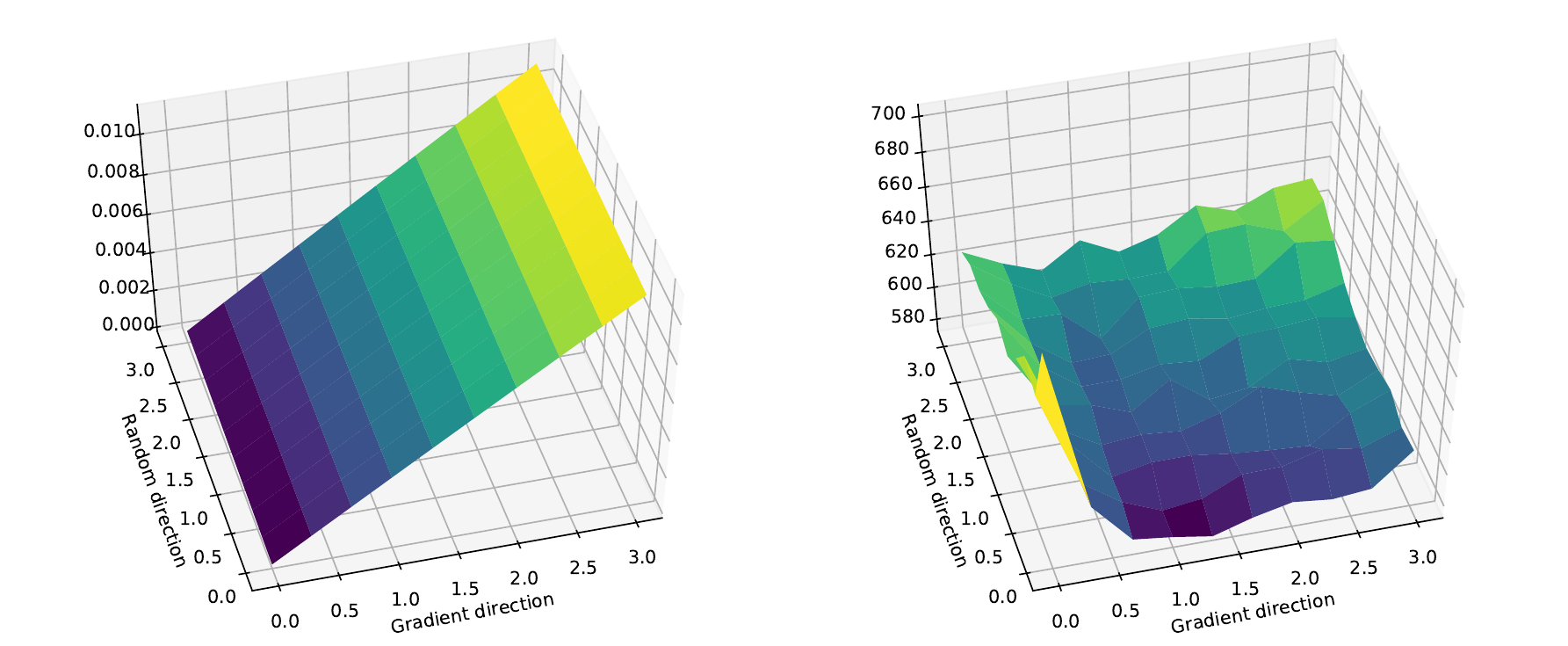} \\
\raisebox{1.5cm}{\rotatebox[origin=c]{90}{Step 450}} &
    \includegraphics[width=.45\textwidth]{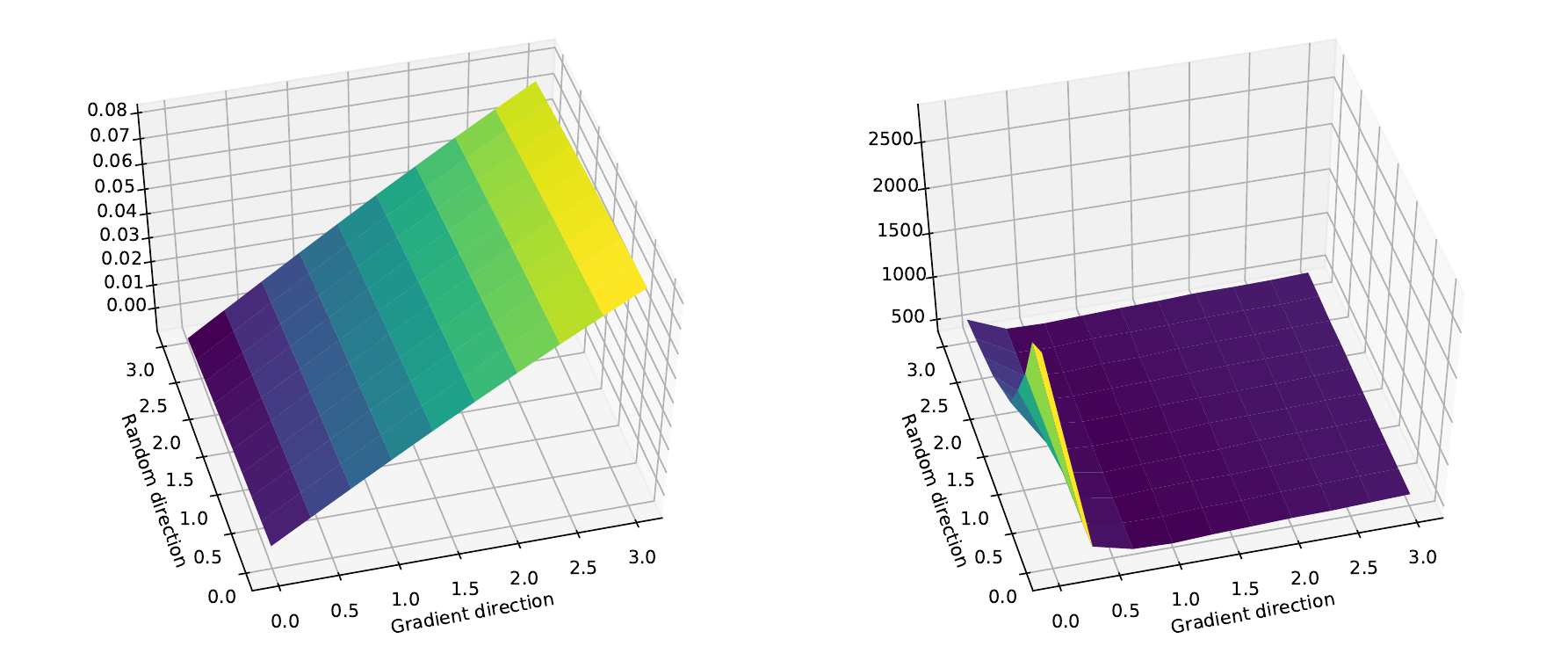} &
    \includegraphics[width=.45\textwidth]{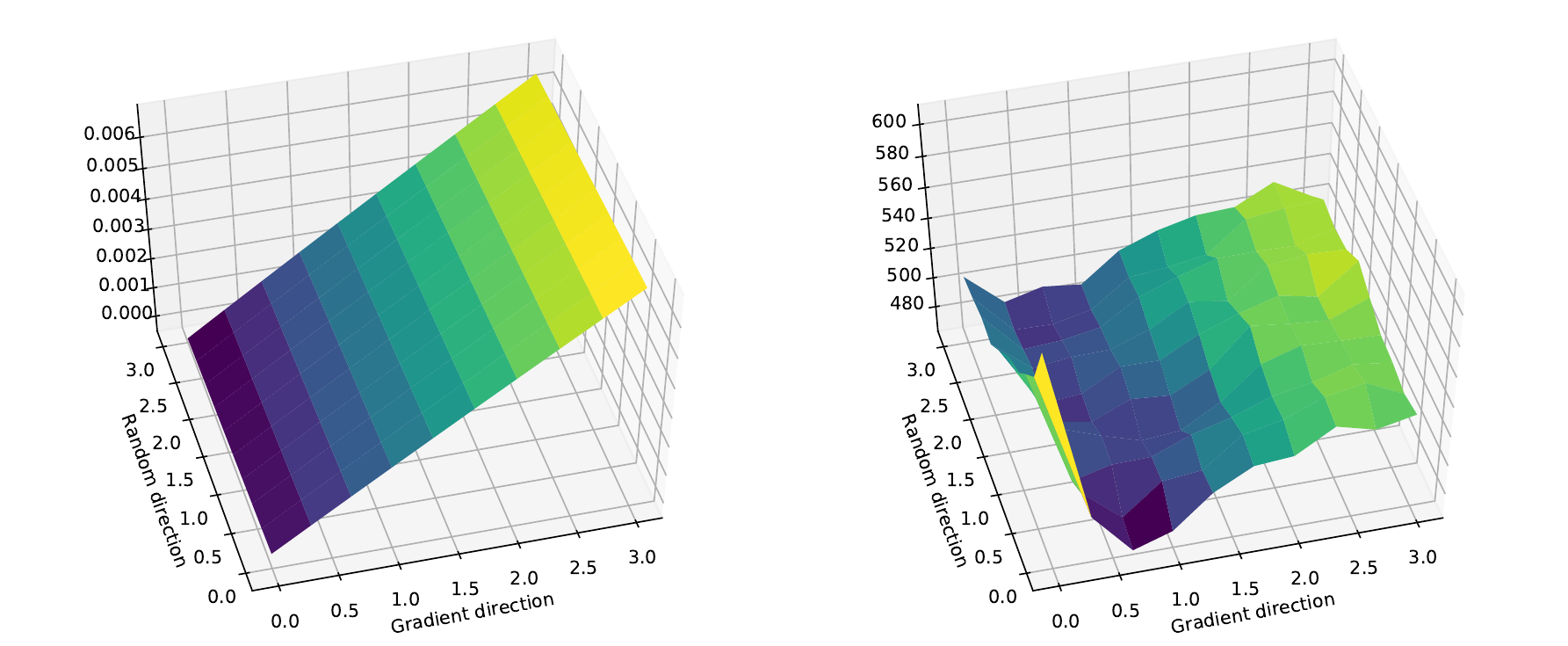}
\end{tabular}
\caption{Walker2d-v2 -- PPO reward landscapes.}
\label{fig:walker2d_landscape_ppo}
\end{figure}

\begin{figure}[htp]
\begin{tabular}{cc|c}
& Few state-action pairs (2,000) & Many state-action pairs ($10^6$) \\
& \begin{tabularx}{.45\textwidth}{CC} Surrogate & True reward \end{tabularx}
& \begin{tabularx}{.45\textwidth}{CC} Surrogate & True reward \end{tabularx} \\
\raisebox{1.5cm}{\rotatebox[origin=c]{90}{Step 0}} &
    \includegraphics[width=.45\textwidth]{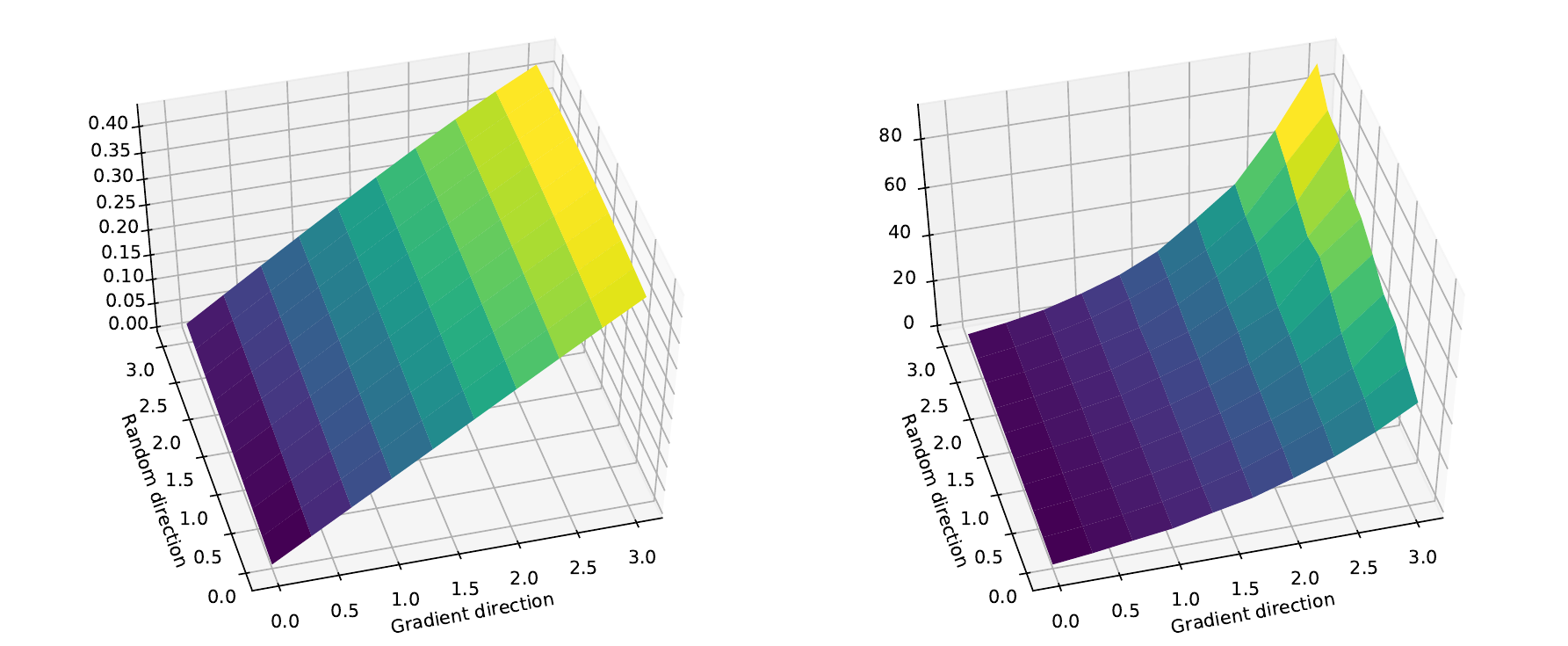} &
    \includegraphics[width=.45\textwidth]{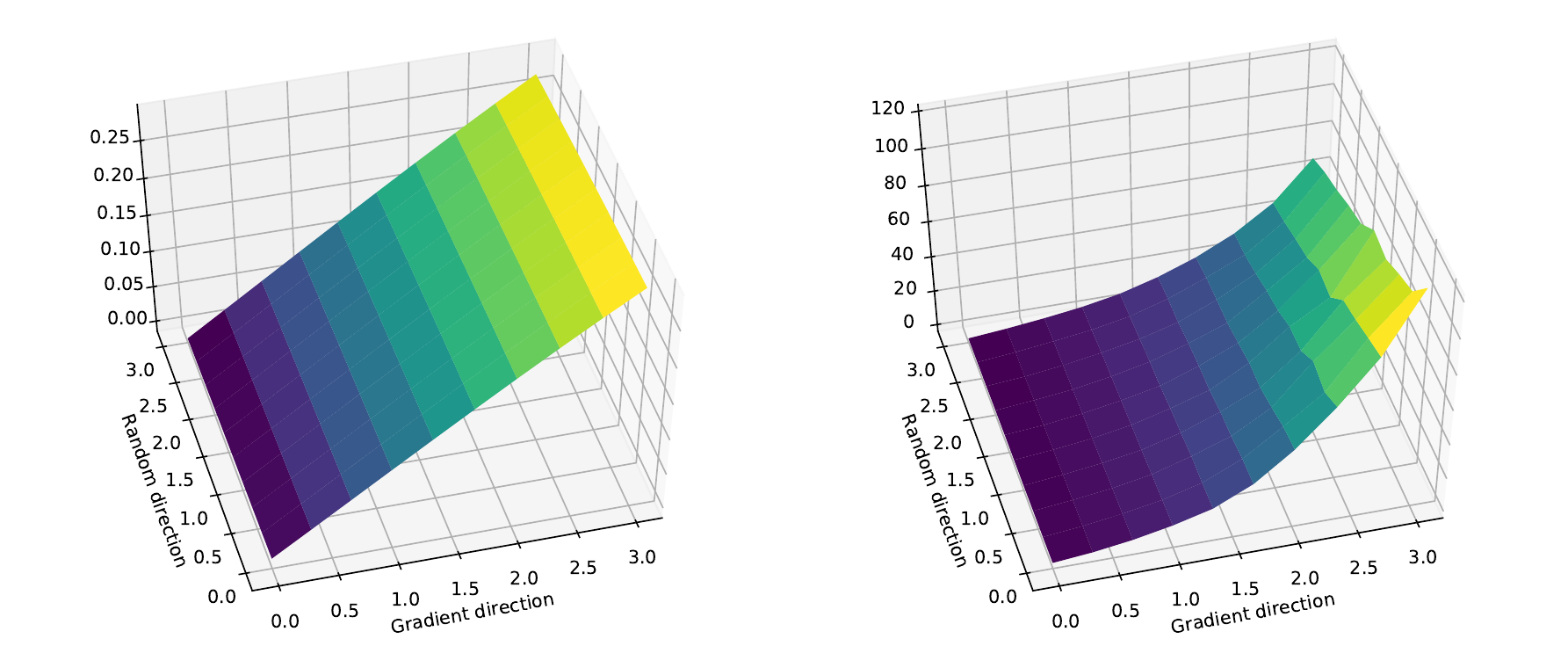} \\
\raisebox{1.5cm}{\rotatebox[origin=c]{90}{Step 150}} &
    \includegraphics[width=.45\textwidth]{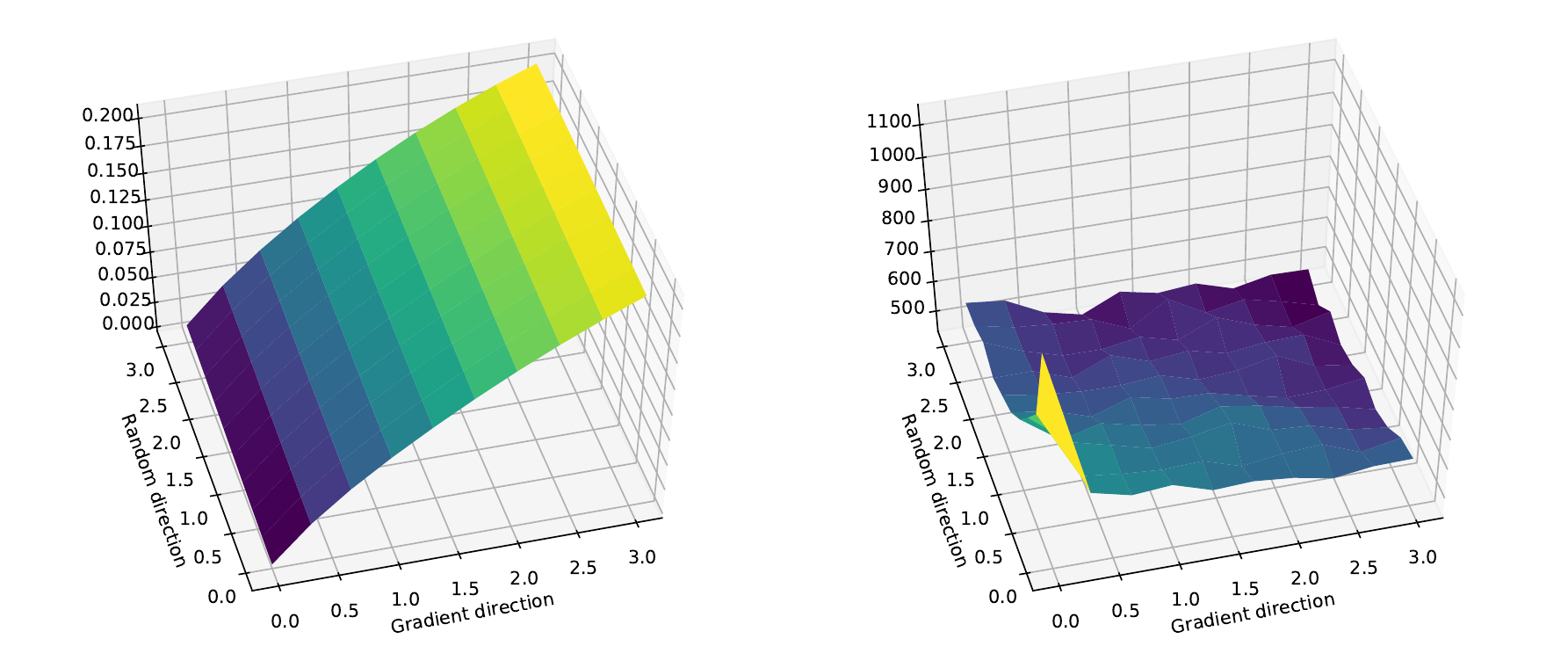} &
    \includegraphics[width=.45\textwidth]{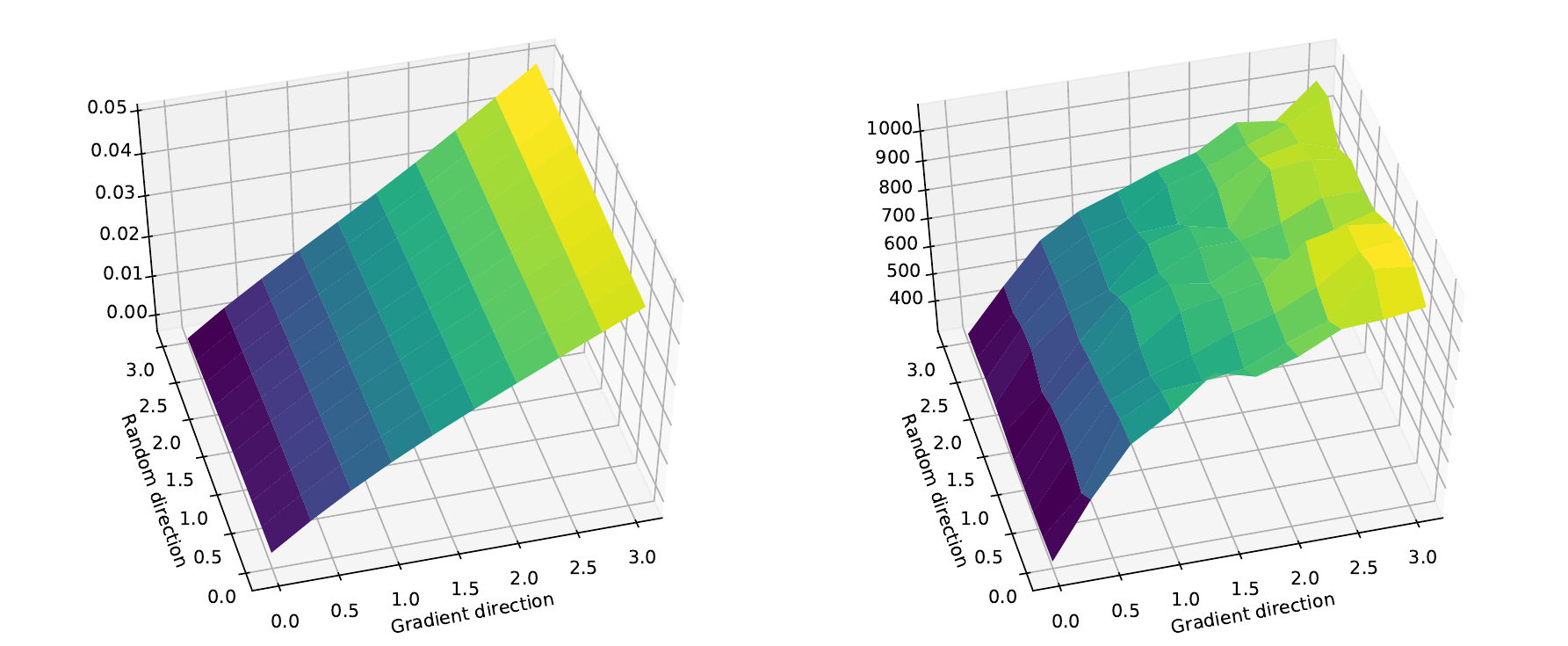} \\
\raisebox{1.5cm}{\rotatebox[origin=c]{90}{Step 300}} &
    \includegraphics[width=.45\textwidth]{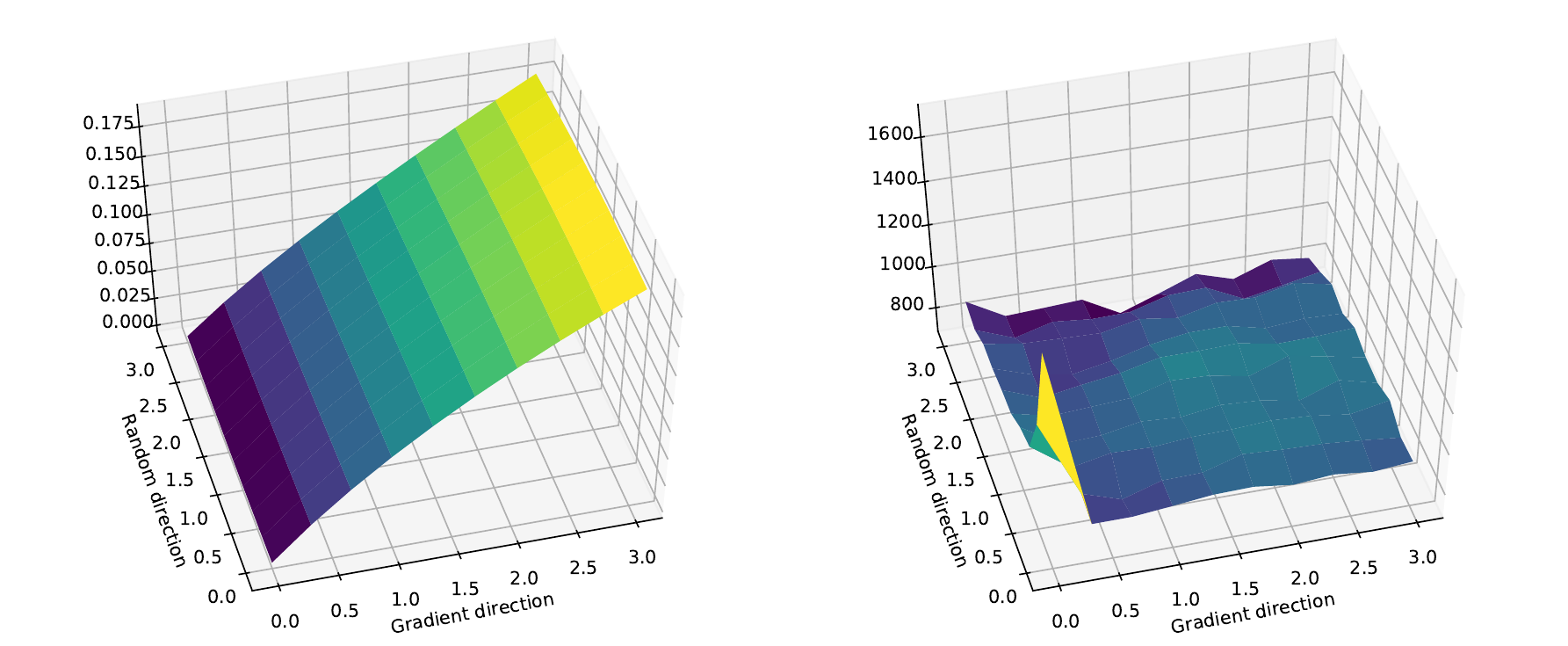} &
    \includegraphics[width=.45\textwidth]{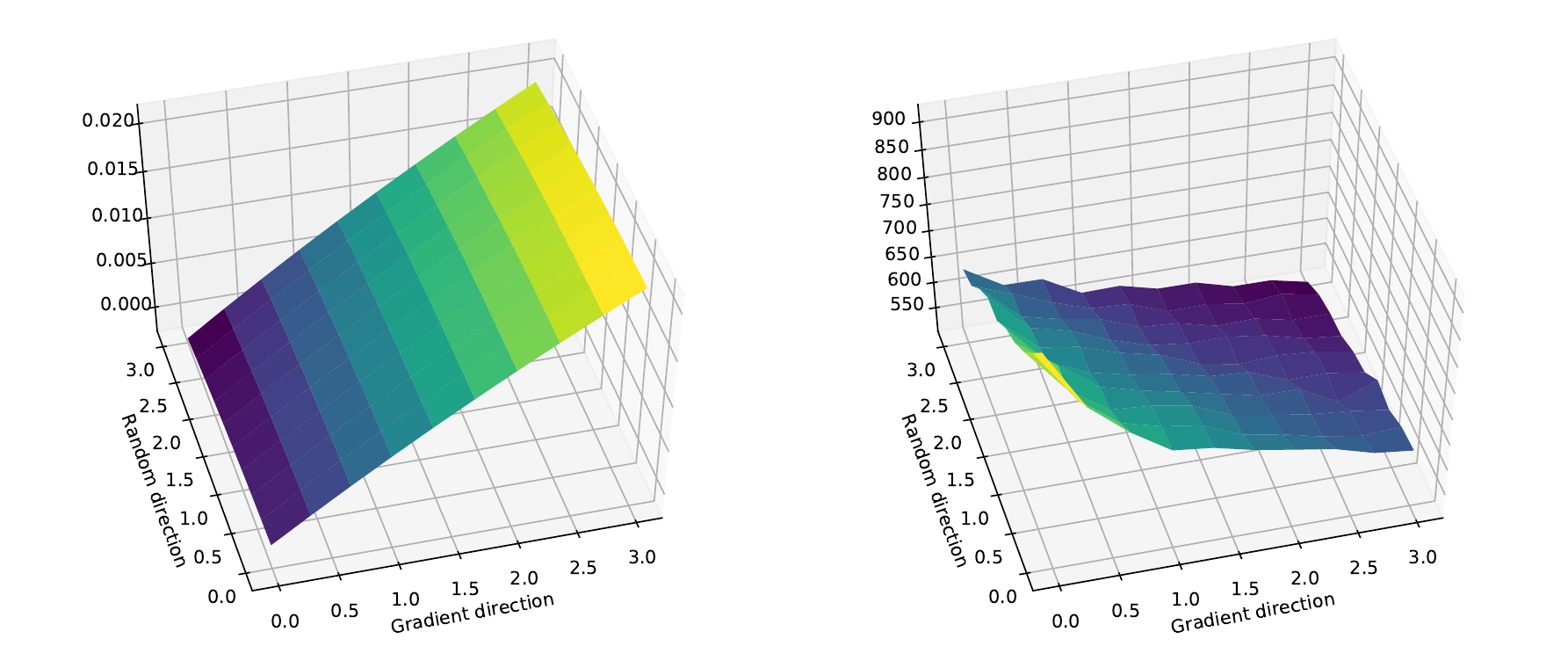} \\
\raisebox{1.5cm}{\rotatebox[origin=c]{90}{Step 450}} &
    \includegraphics[width=.45\textwidth]{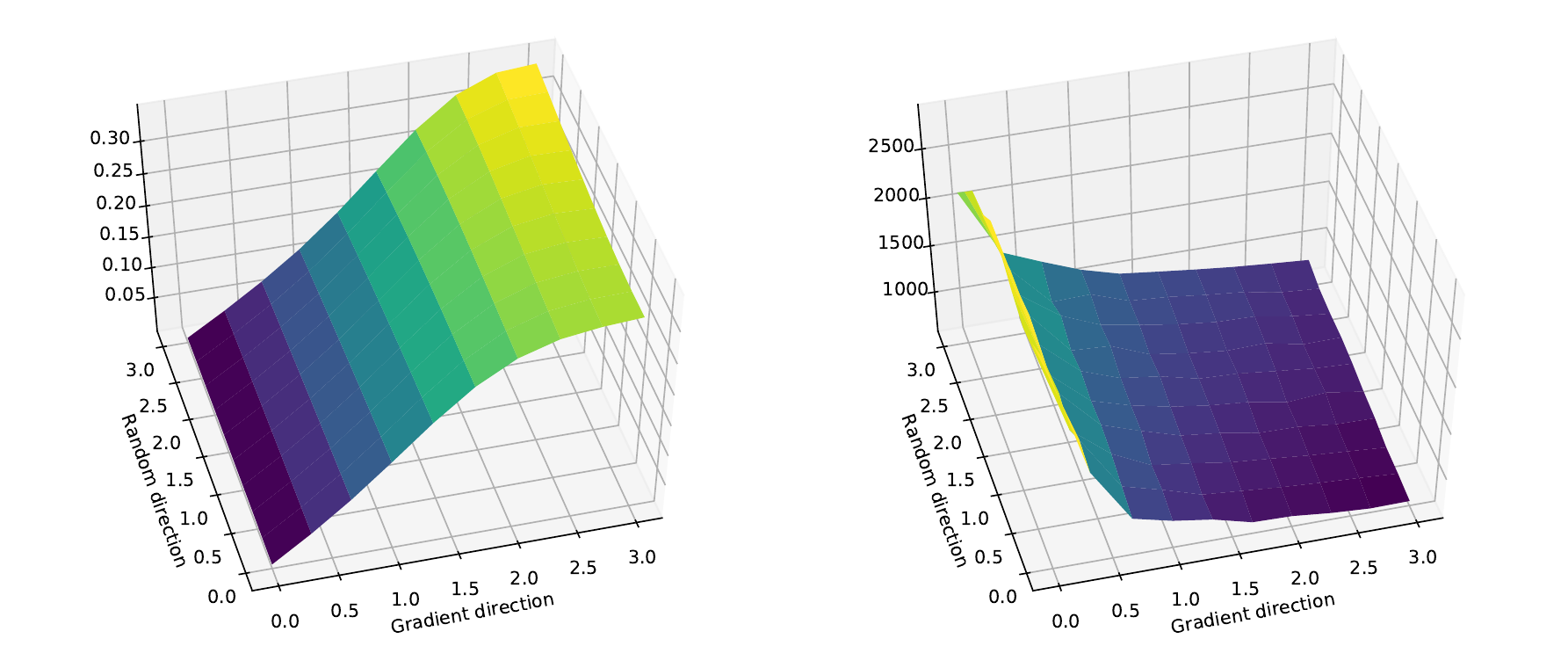} &
    \includegraphics[width=.45\textwidth]{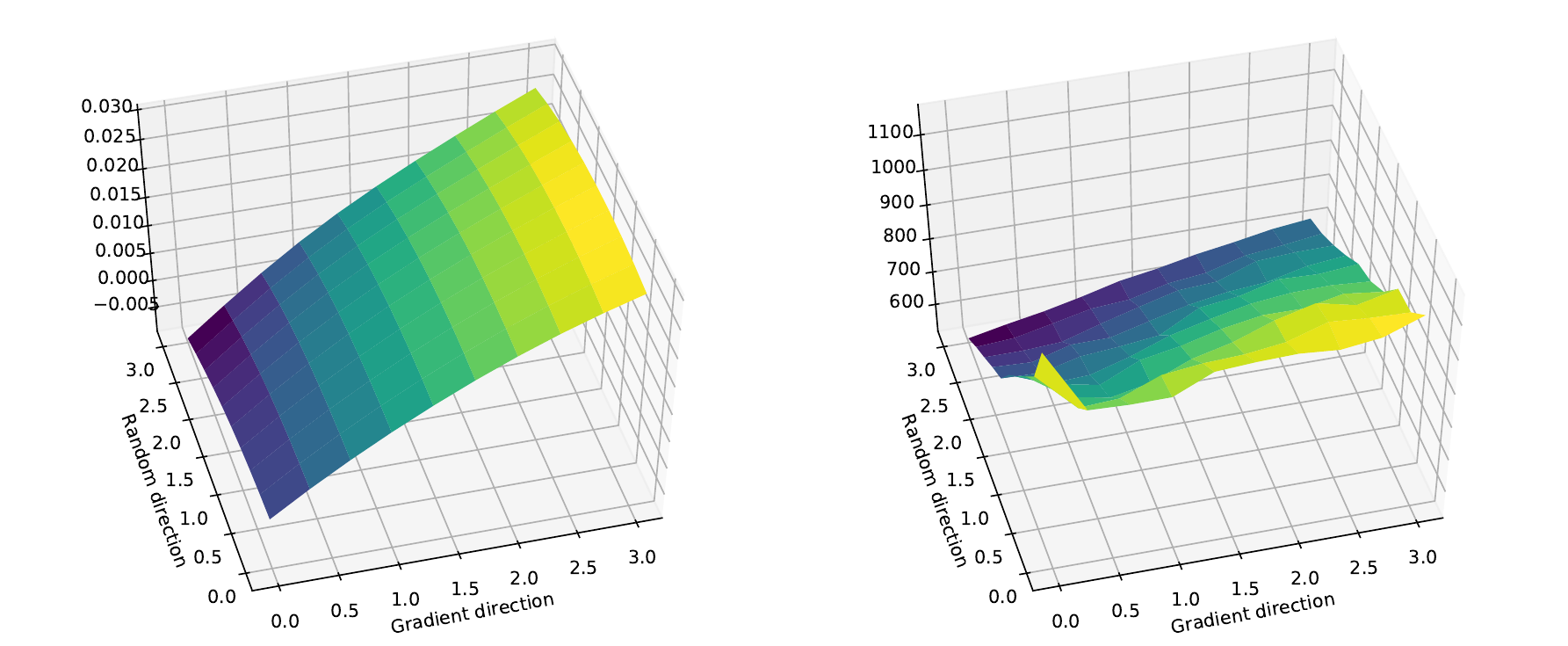}
\end{tabular}
\caption{Walker2d-v2 -- TRPO reward landscapes.}
\label{fig:walker2d_landscape_trpo}
\end{figure}

\begin{figure}[htp]
\begin{tabular}{cc|c}
& Few state-action pairs (2,000) & Many state-action pairs ($10^6$) \\
& \begin{tabularx}{.45\textwidth}{CC} Surrogate & True reward \end{tabularx}
& \begin{tabularx}{.45\textwidth}{CC} Surrogate & True reward \end{tabularx} \\
\raisebox{1.5cm}{\rotatebox[origin=c]{90}{Step 0}} &
    \includegraphics[width=.45\textwidth]{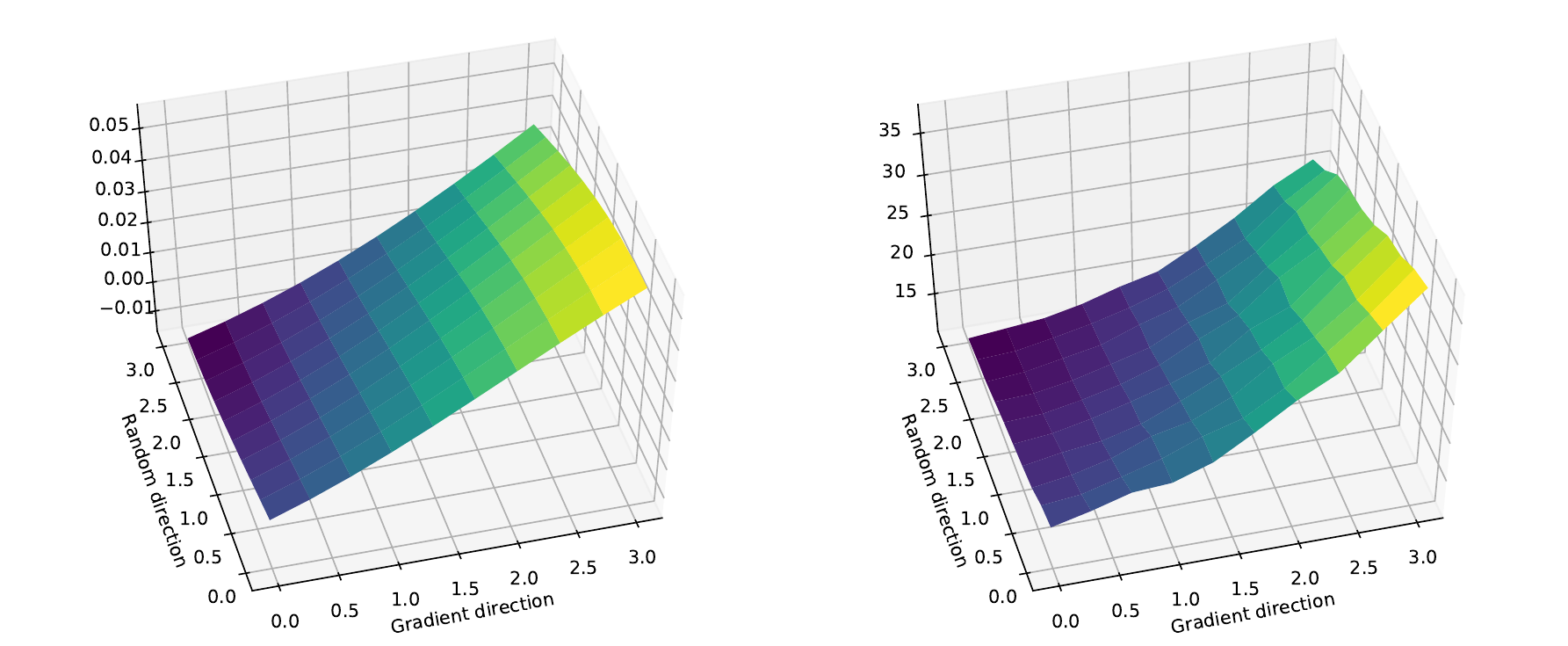} &
    \includegraphics[width=.45\textwidth]{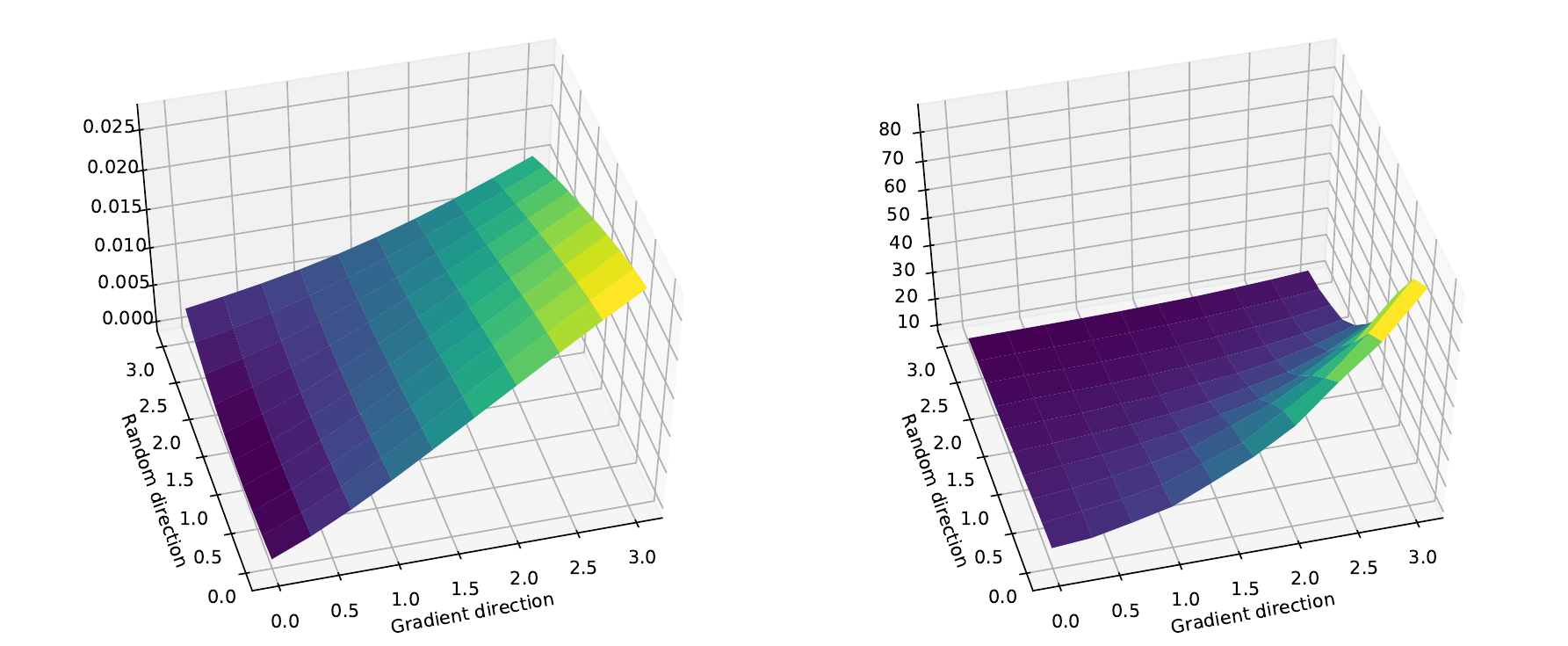} \\
\raisebox{1.5cm}{\rotatebox[origin=c]{90}{Step 150}} &
    \includegraphics[width=.45\textwidth]{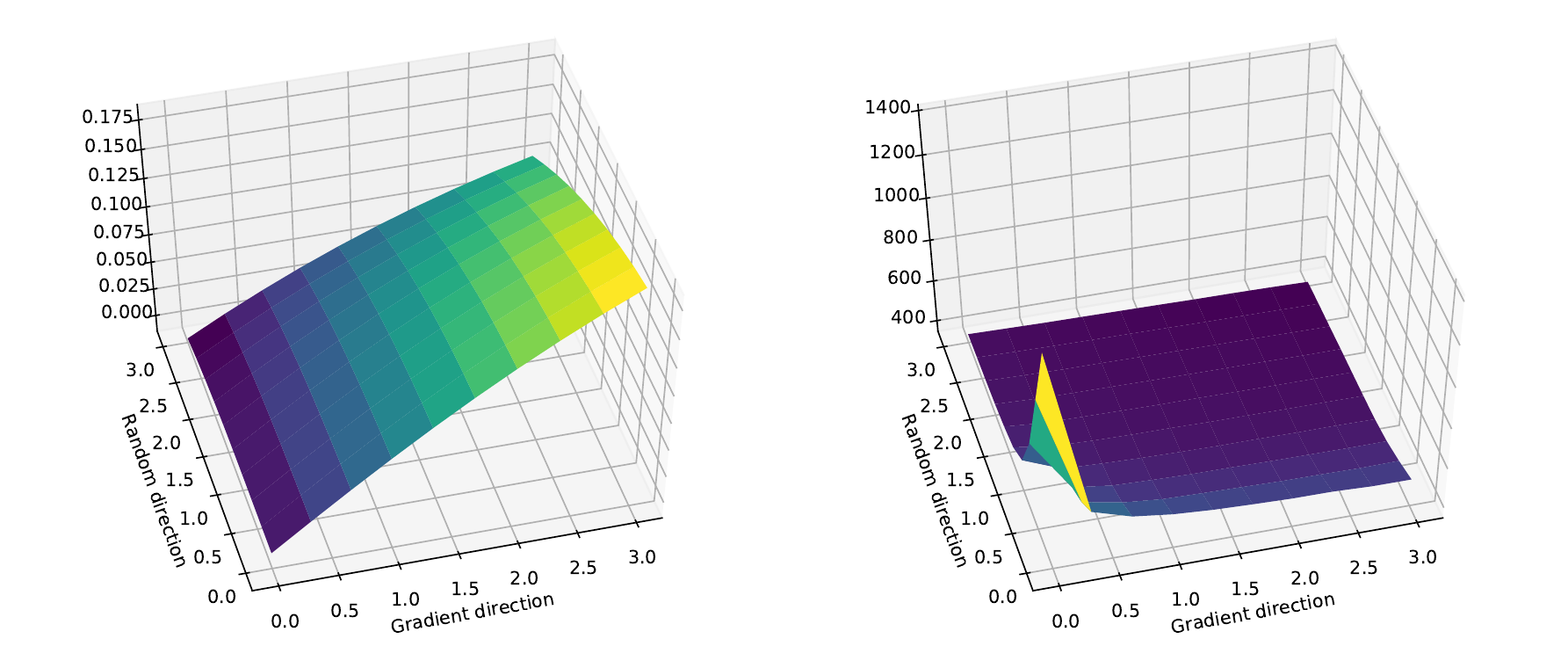} &
    \includegraphics[width=.45\textwidth]{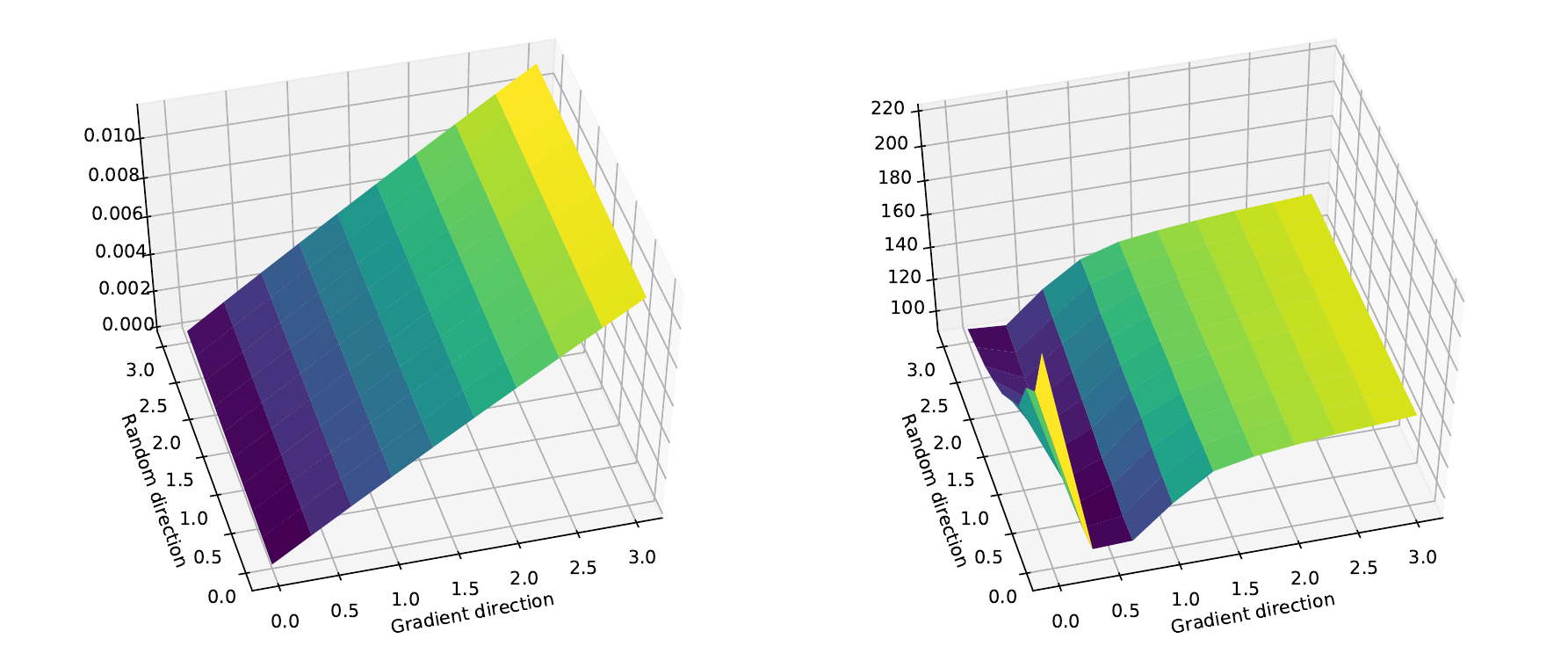} \\
\raisebox{1.5cm}{\rotatebox[origin=c]{90}{Step 300}} &
    \includegraphics[width=.45\textwidth]{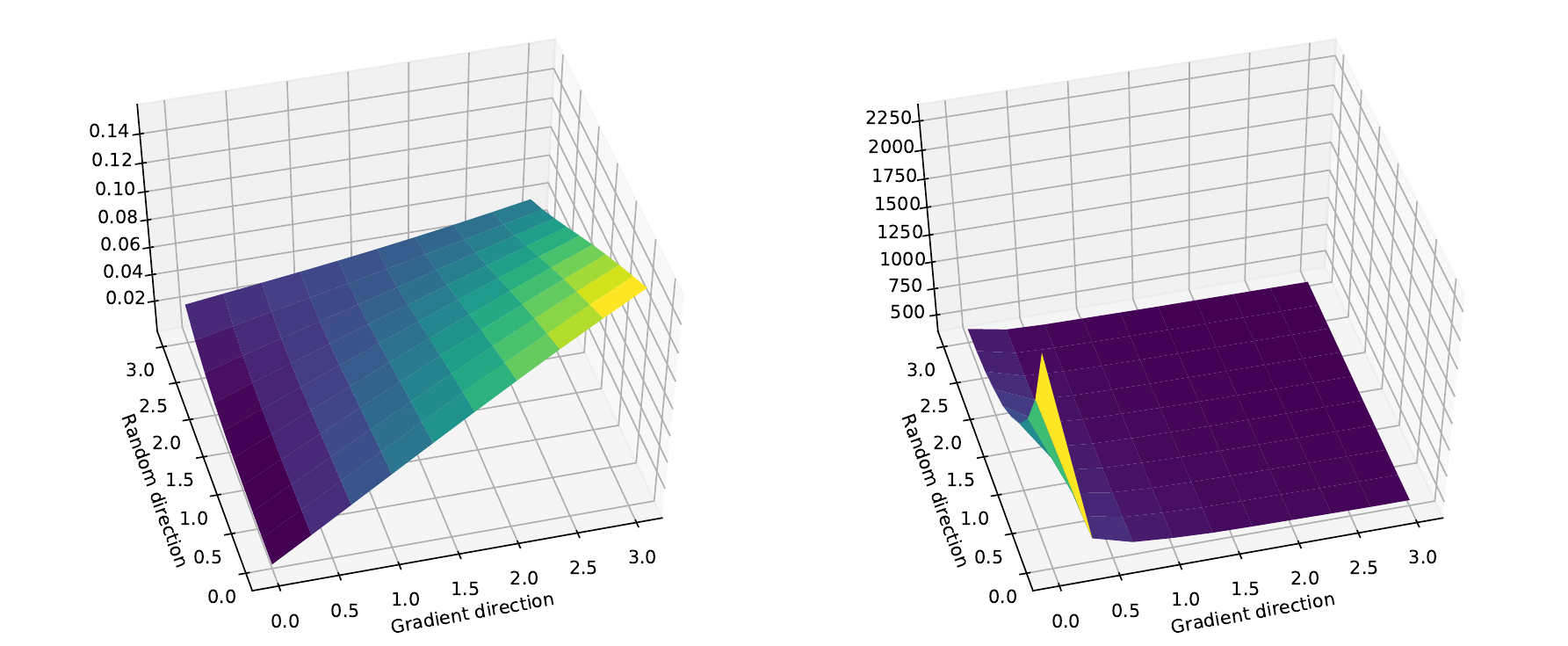} &
    \includegraphics[width=.45\textwidth]{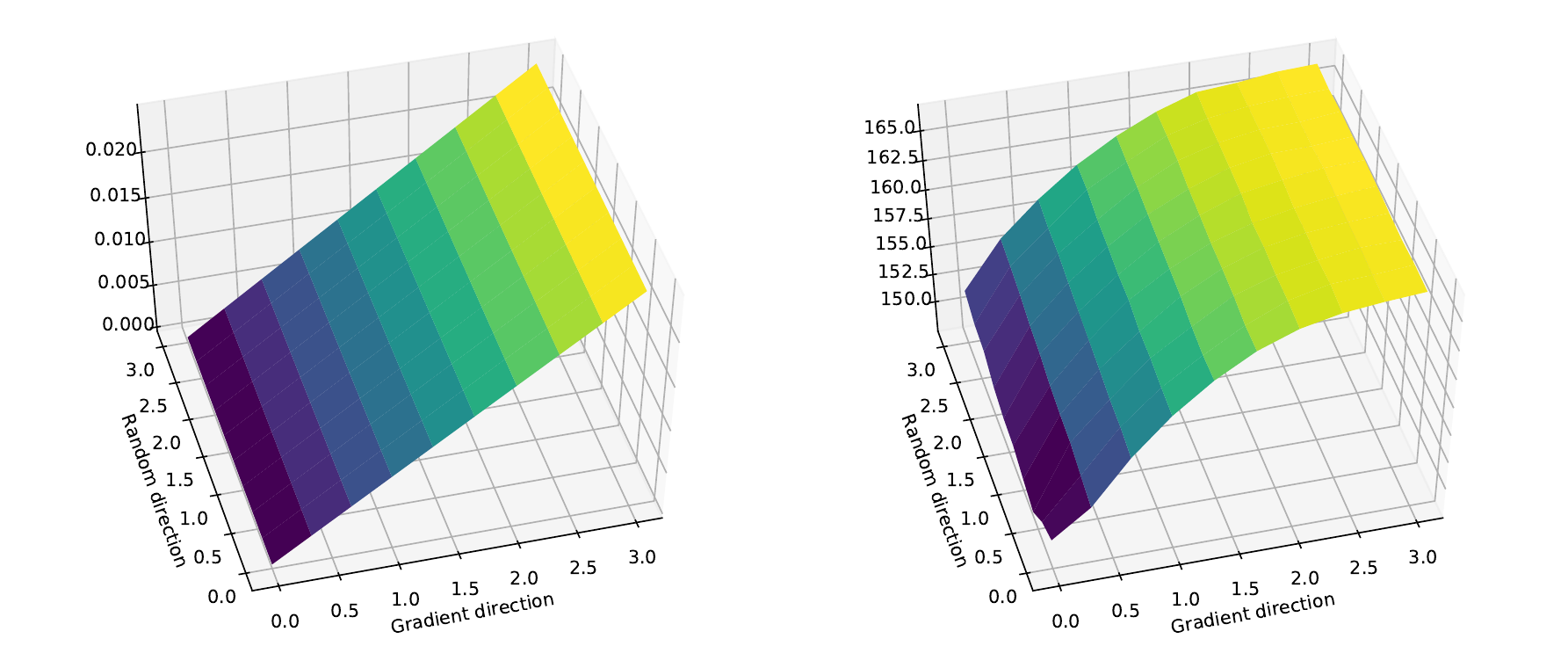} \\
\raisebox{1.5cm}{\rotatebox[origin=c]{90}{Step 450}} &
    \includegraphics[width=.45\textwidth]{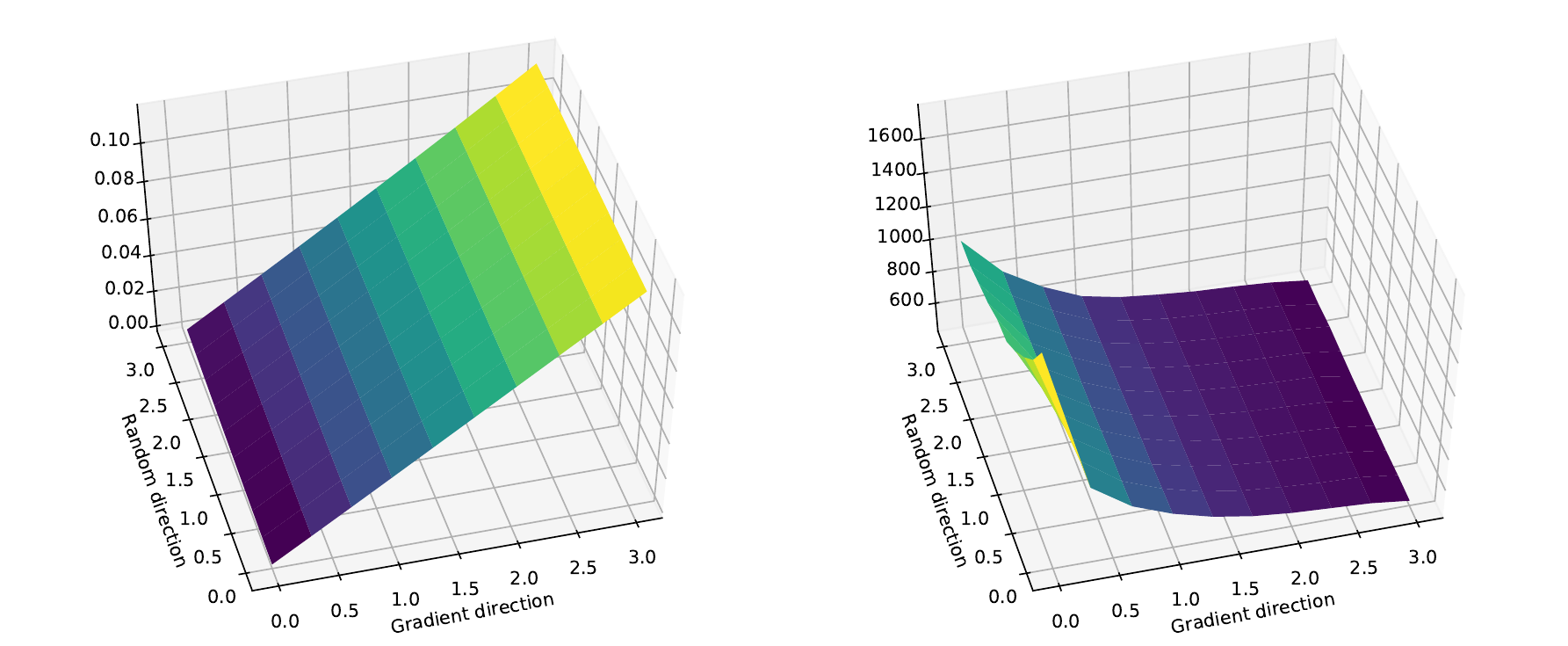} &
    \includegraphics[width=.45\textwidth]{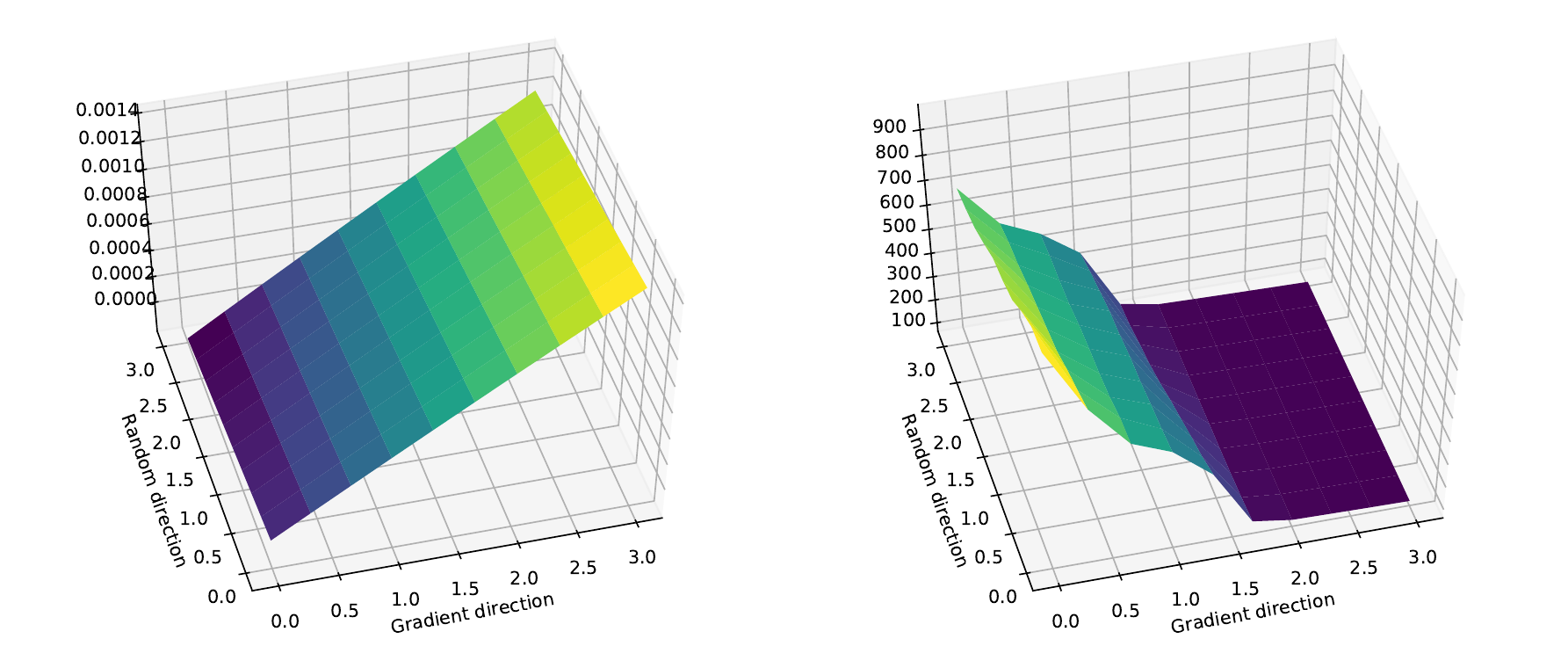}
\end{tabular}
\caption{Hopper-v2 -- PPO reward landscapes.}
\label{fig:hopper_landscape_ppo}
\end{figure}

\begin{figure}[htp]
\begin{tabular}{cc|c}
& Few state-action pairs (2,000) & Many state-action pairs ($10^6$) \\
& \begin{tabularx}{.45\textwidth}{CC} Surrogate & True reward \end{tabularx}
& \begin{tabularx}{.45\textwidth}{CC} Surrogate & True reward \end{tabularx} \\
\raisebox{1.5cm}{\rotatebox[origin=c]{90}{Step 0}} &
    \includegraphics[width=.45\textwidth]{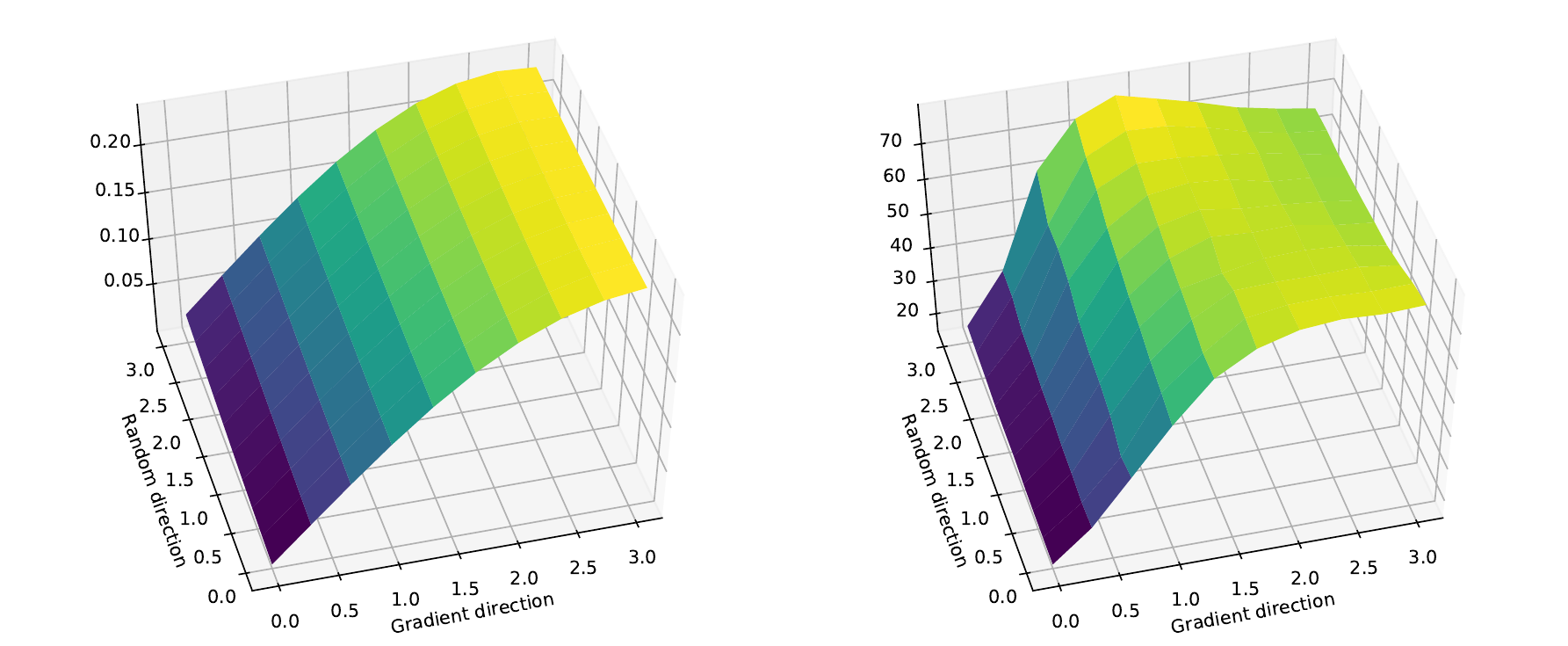} &
    \includegraphics[width=.45\textwidth]{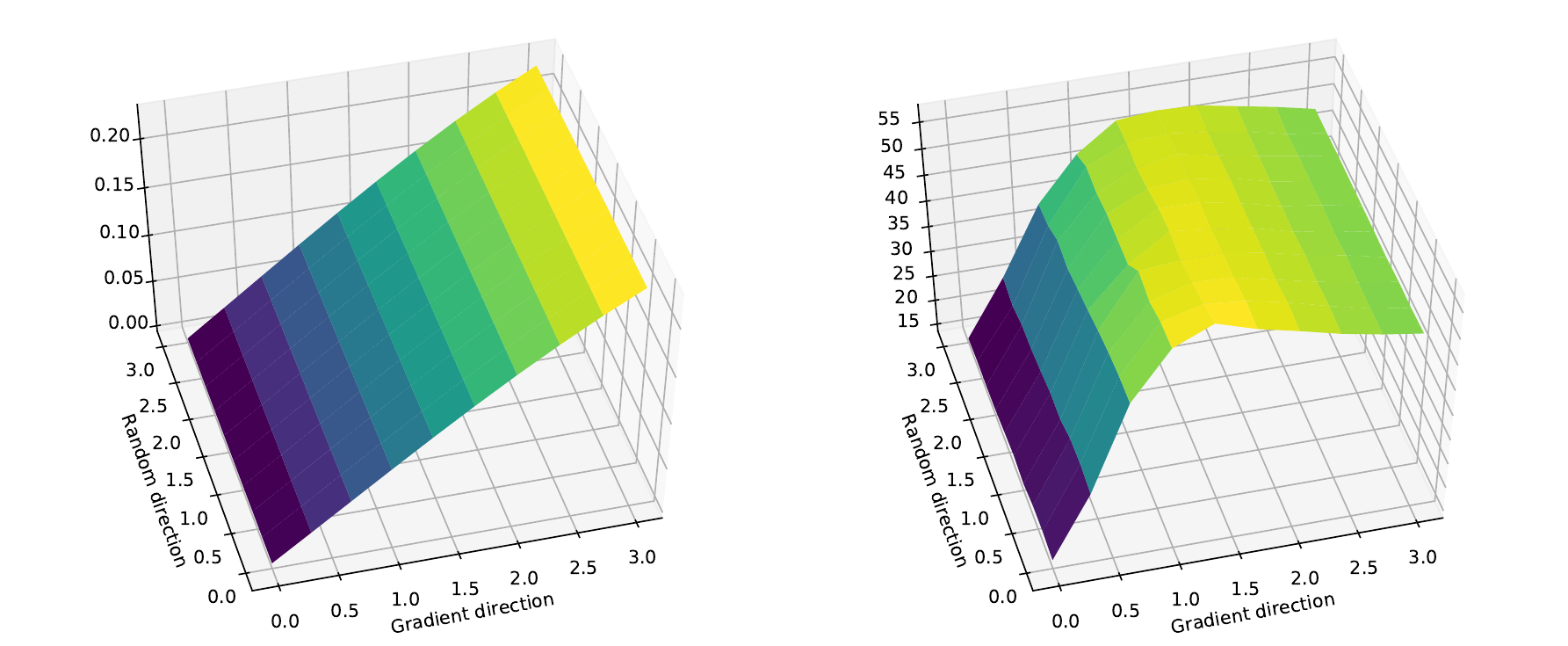} \\
\raisebox{1.5cm}{\rotatebox[origin=c]{90}{Step 150}} &
    \includegraphics[width=.45\textwidth]{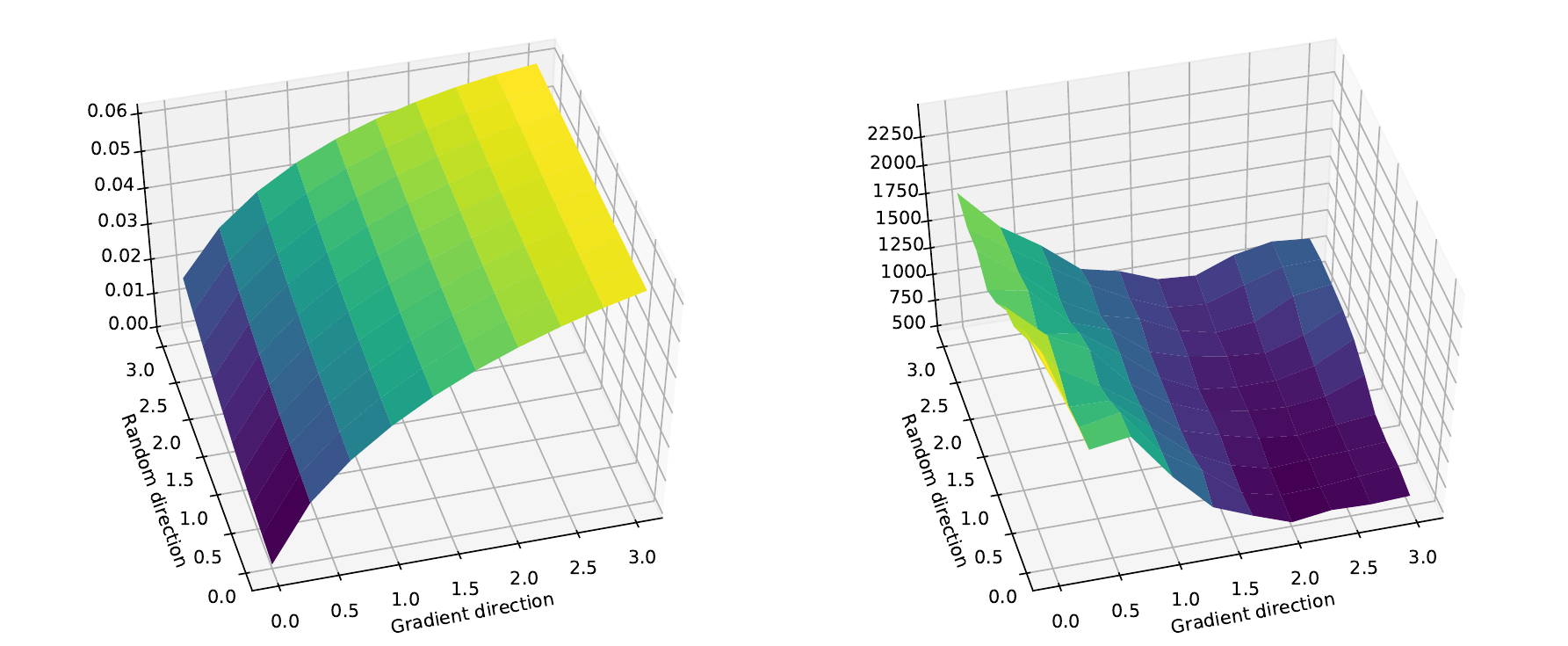} &
    \includegraphics[width=.45\textwidth]{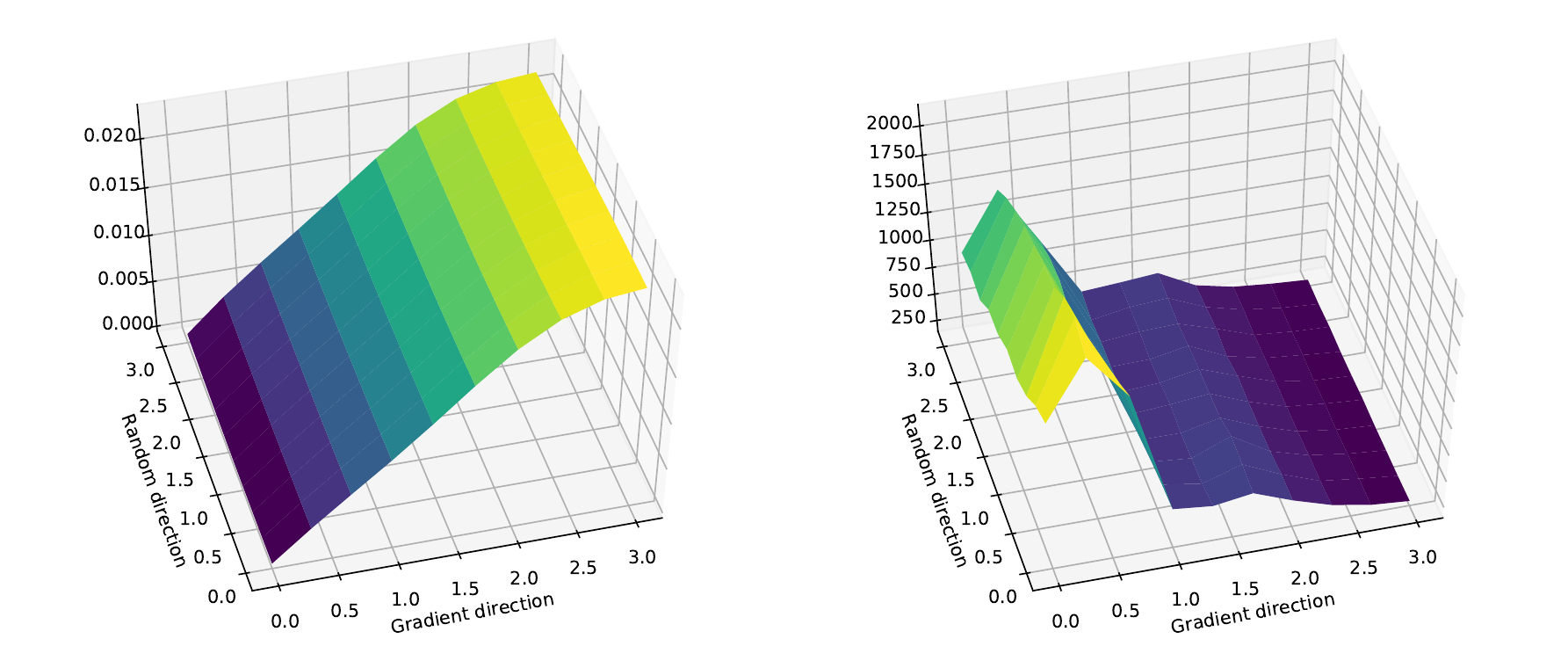} \\
\raisebox{1.5cm}{\rotatebox[origin=c]{90}{Step 300}} &
    \includegraphics[width=.45\textwidth]{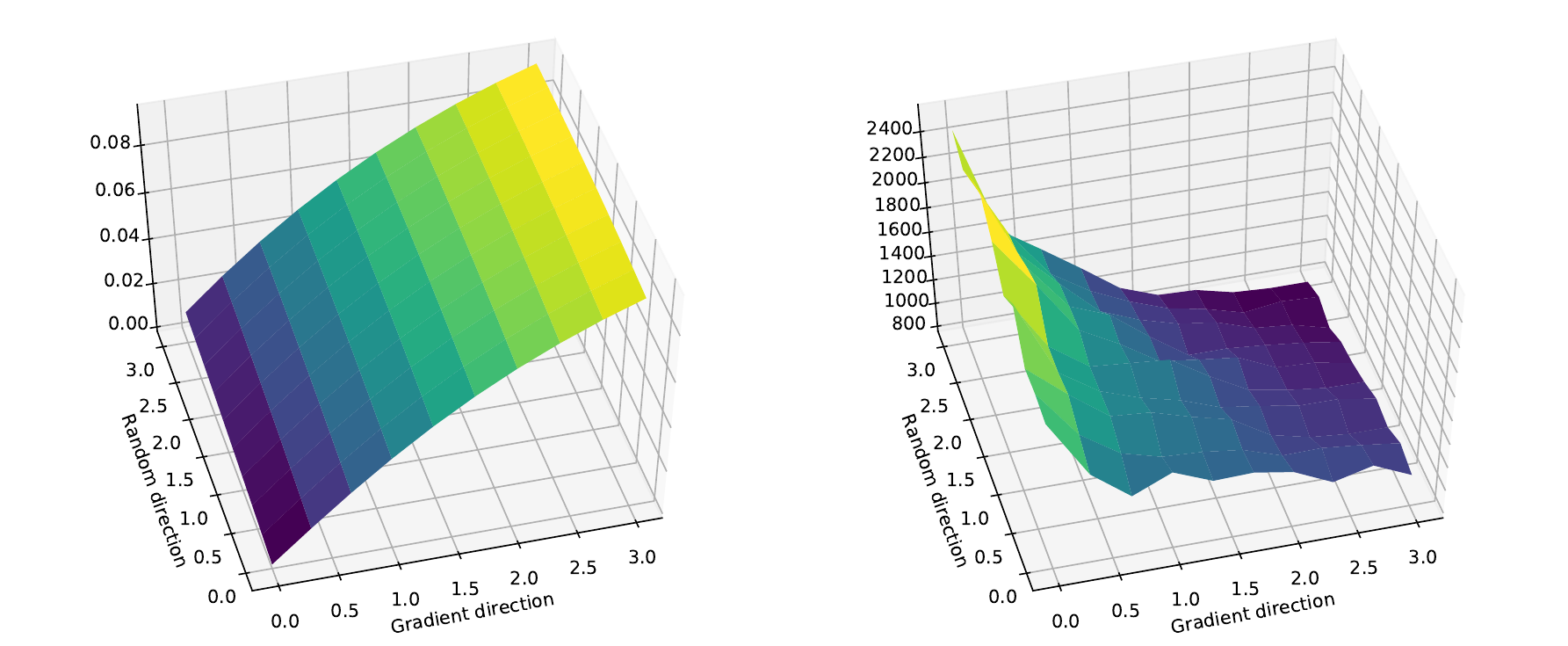} &
    \includegraphics[width=.45\textwidth]{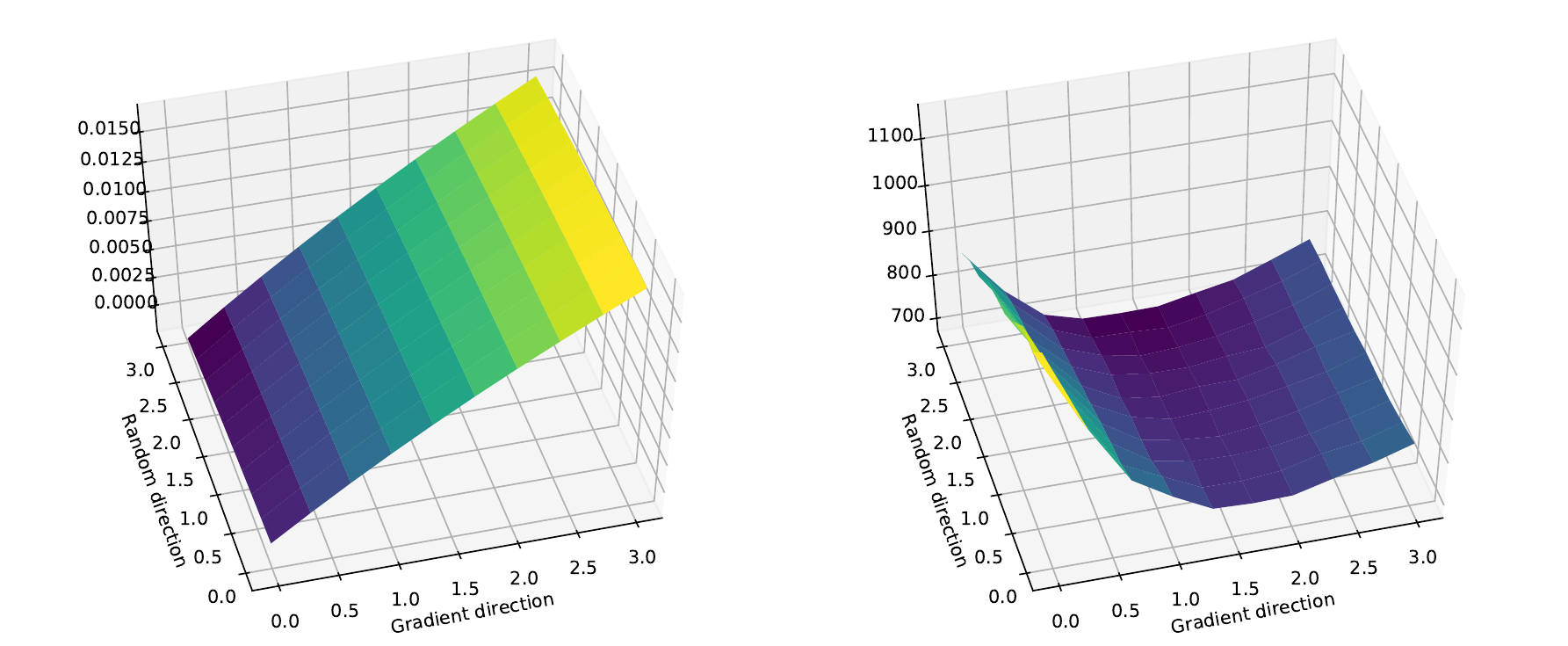} \\
\raisebox{1.5cm}{\rotatebox[origin=c]{90}{Step 450}} &
    \includegraphics[width=.45\textwidth]{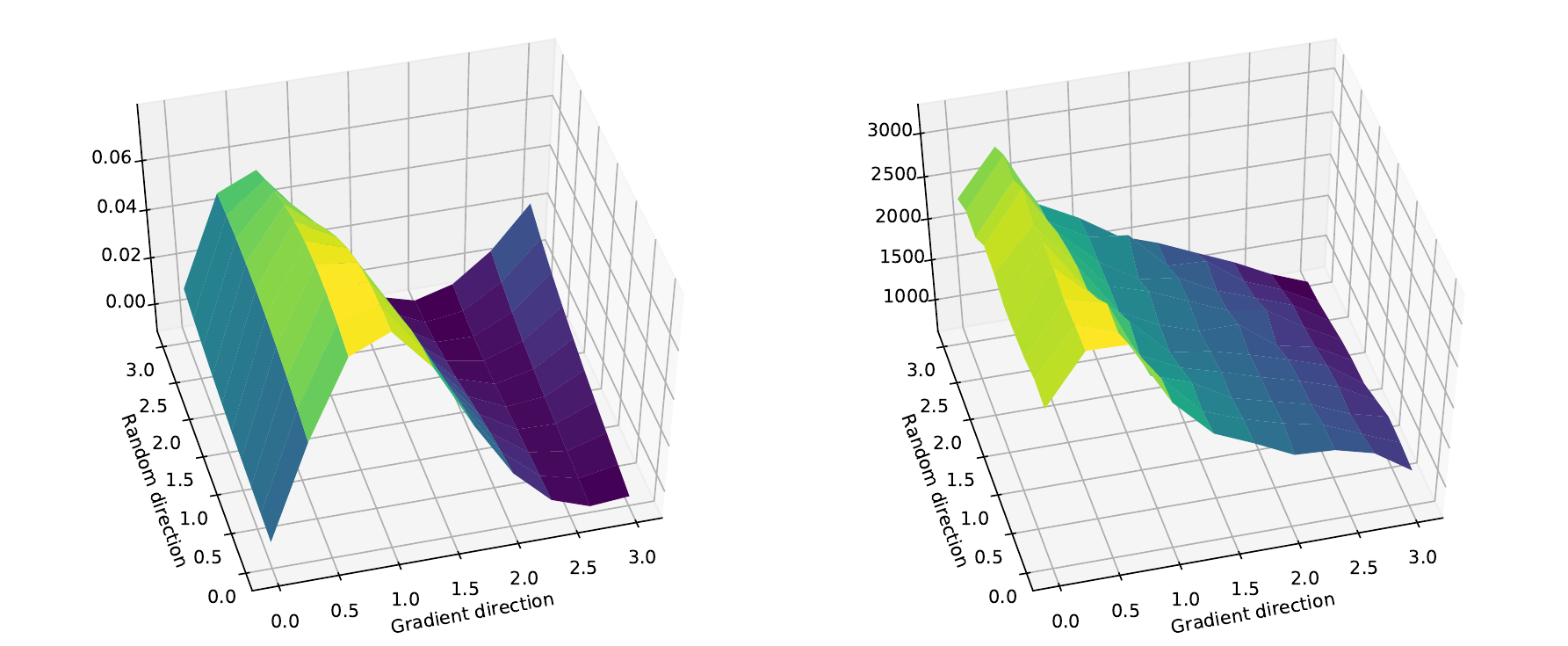} &
    \includegraphics[width=.45\textwidth]{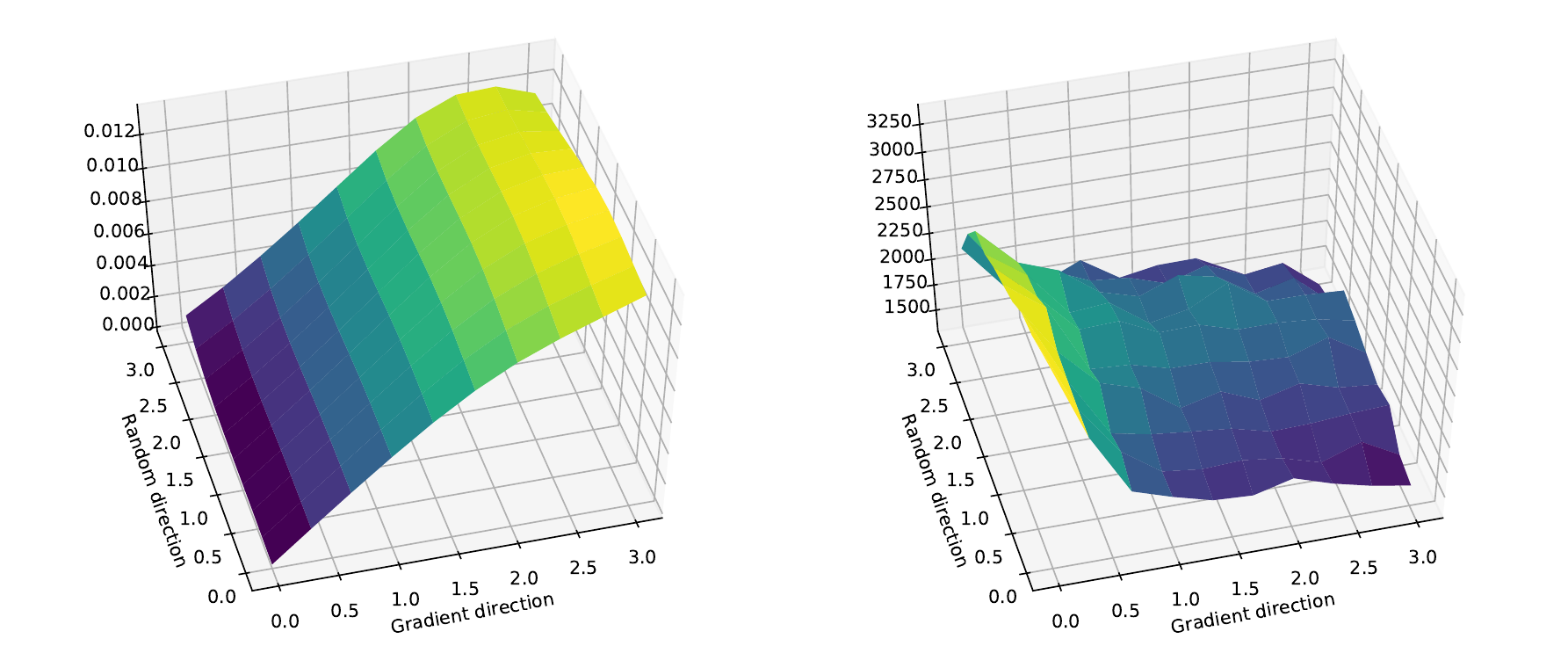}
\end{tabular}
\caption{Hopper-v2 -- TRPO reward landscapes.}
\label{fig:hopper_landscape_trpo}
\end{figure}

\begin{figure}
\begin{tabular}{cc}
& \begin{tabularx}{\textwidth}{CCC}
    2,000 state-action pairs
    & 20,000 state-action pairs
    & 100,000 state-action pairs
\end{tabularx}\\
\raisebox{2cm}{\rotatebox[origin=c]{90}{Step 0}} &
\includegraphics[width=\textwidth]{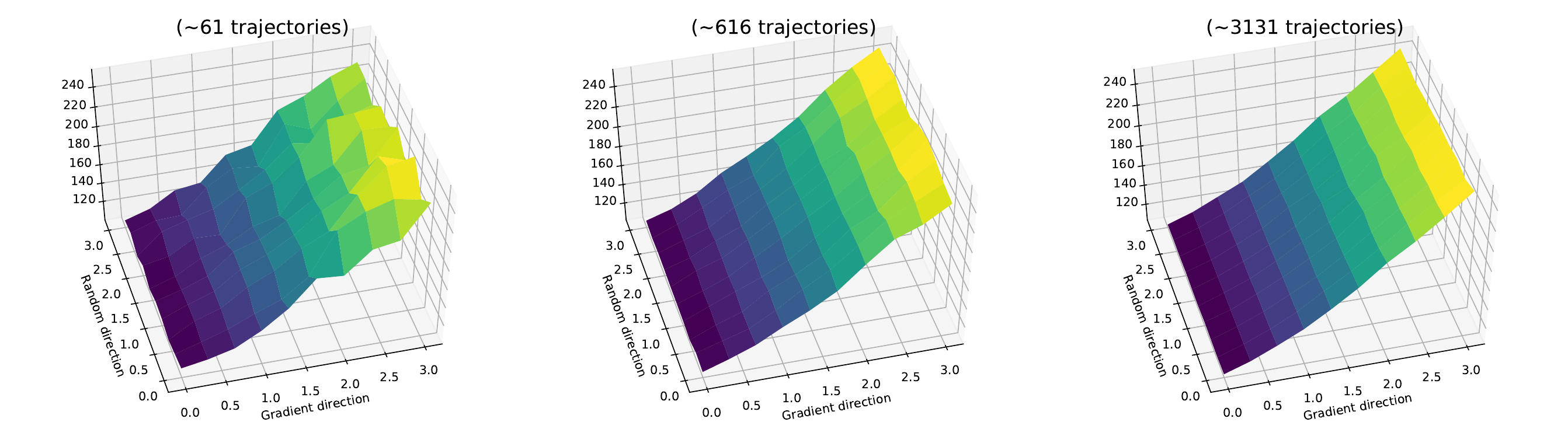} \\
\raisebox{1.8cm}{\rotatebox[origin=c]{90}{Step 150}} &
\includegraphics[width=\textwidth]{Figures/landscapes/humanoid_conc_150_trpo-no-hacks}\\
\raisebox{1.8cm}{\rotatebox[origin=c]{90}{Step 300}} &
 \includegraphics[width=\textwidth]{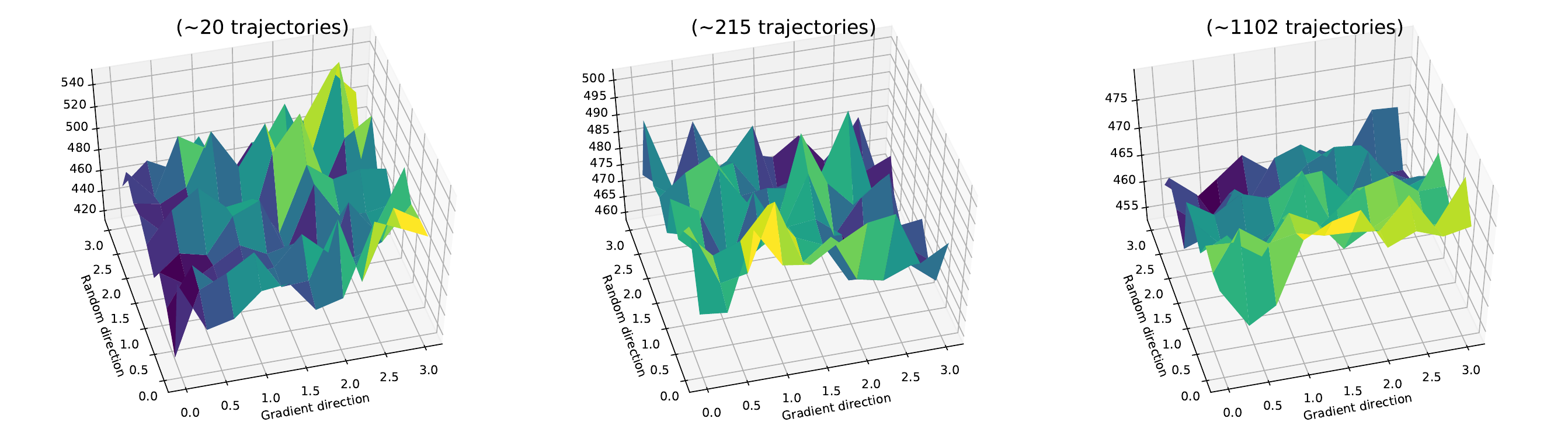}\\
\raisebox{1.8cm}{\rotatebox[origin=c]{90}{Step 450}} &
 \includegraphics[width=\textwidth]{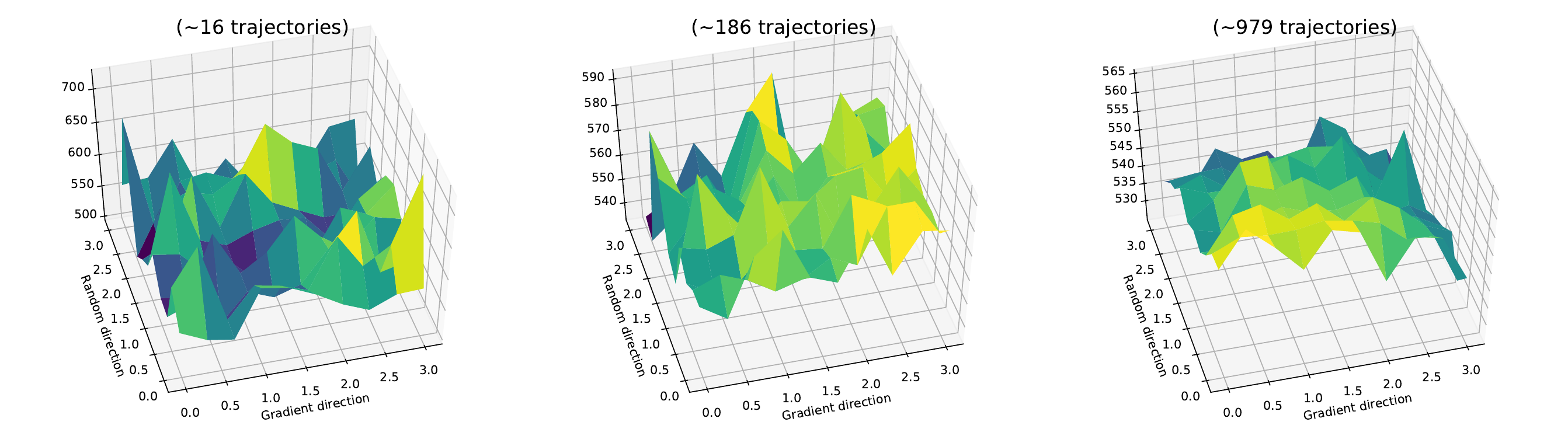}\\
\end{tabular}
\caption{Humanoid-v2 TRPO landscape concentration (see Figure~\ref{fig:trpo_landscape_concentration_main} for a description).}
\label{fig:trpo_landscape_concentration}
\end{figure}

\begin{figure}
\begin{tabular}{cc}
& \begin{tabularx}{\textwidth}{CCC}
    2,000 state-action pairs
    & 20,000 state-action pairs
    & 100,000 state-action pairs
\end{tabularx} \\
\raisebox{2cm}{\rotatebox[origin=c]{90}{Step 0}} &
\includegraphics[width=\textwidth]{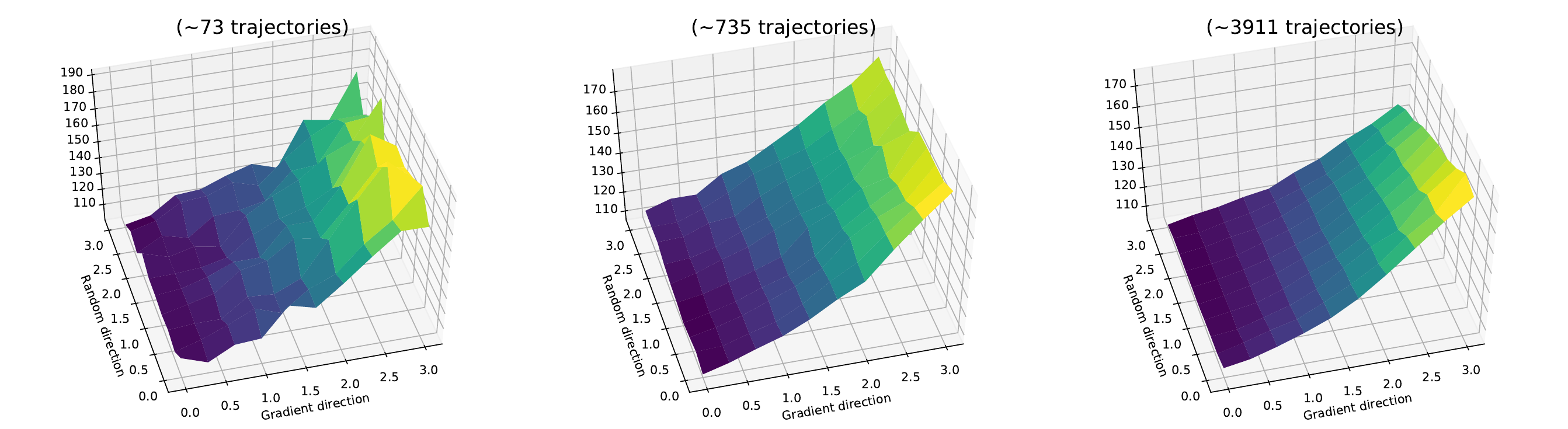}\\
\raisebox{1.8cm}{\rotatebox[origin=c]{90}{Step 150}} &
\includegraphics[width=\textwidth]{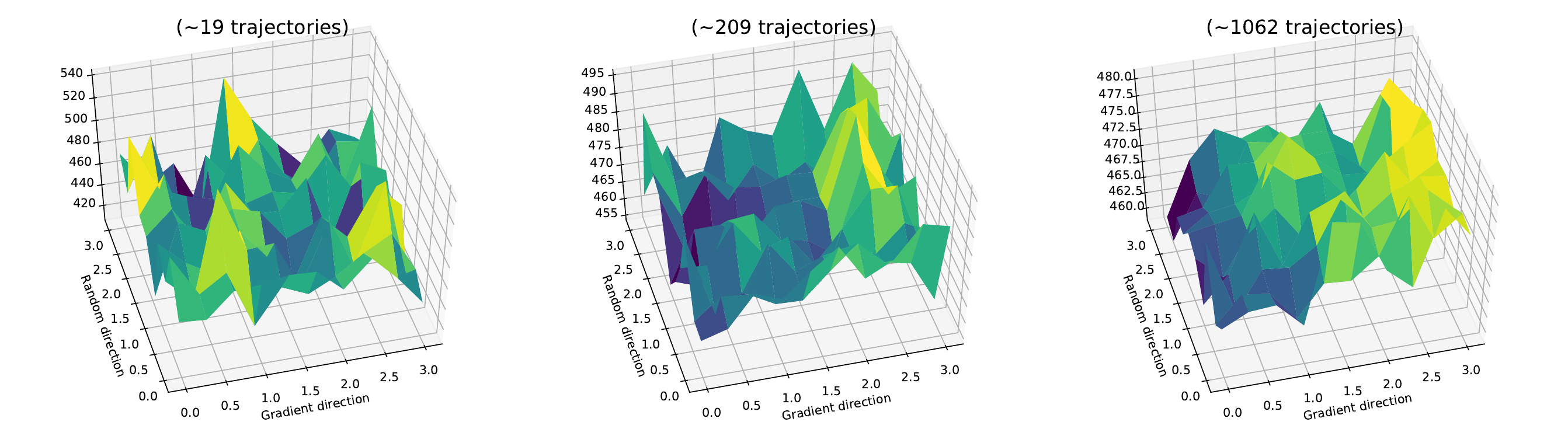}\\
\raisebox{1.8cm}{\rotatebox[origin=c]{90}{Step 300}} &
\includegraphics[width=\textwidth]{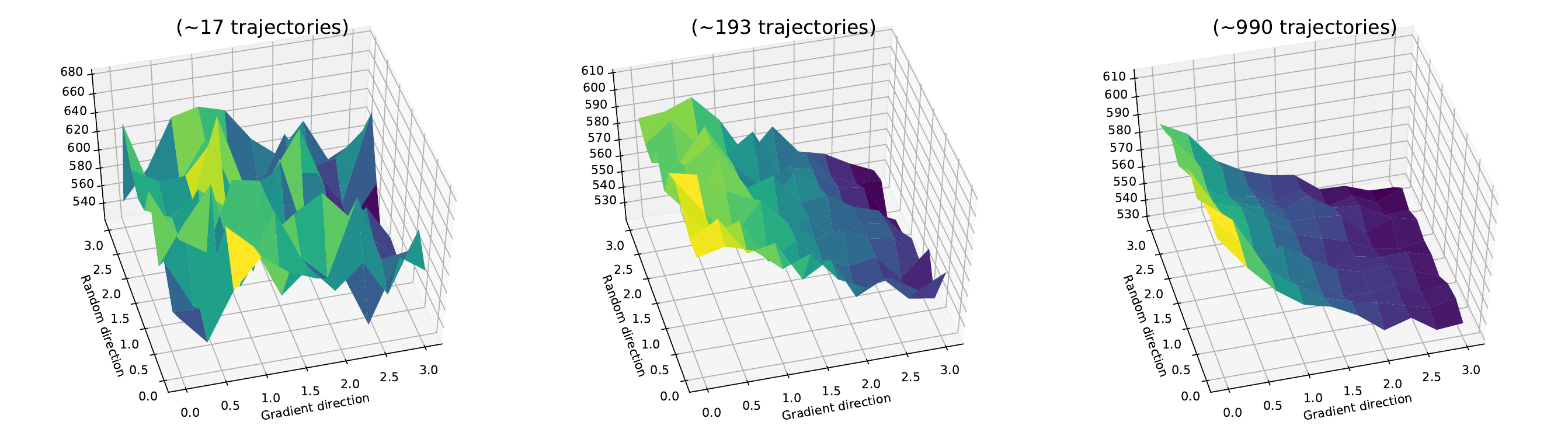}\\
\raisebox{1.8cm}{\rotatebox[origin=c]{90}{Step 450}} &
\includegraphics[width=\textwidth]{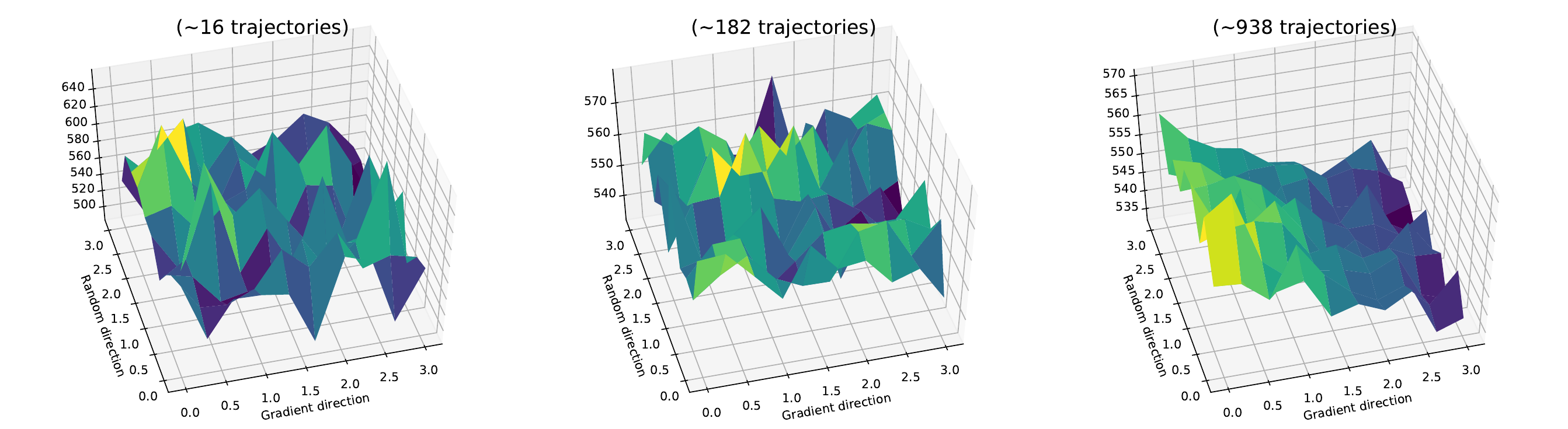}\\
\end{tabular}
\caption{Humanoid-v2 PPO landscape concentration (see Figure~\ref{fig:trpo_landscape_concentration_main} for a description).}
\label{fig:ppo_landscape_concentration}
\end{figure}

\end{document}